\pdfoutput=1
\documentclass[10pt]{article}

\usepackage[OT1]{fontenc}
\usepackage{amsmath,amssymb,amsthm}
\usepackage{natbib}
\usepackage[dvipsnames]{xcolor}
\usepackage[margin=1in]{geometry}
\usepackage{graphicx}
\usepackage{needspace}
\usepackage{mathtools,booktabs,enumitem,bbm}
\usepackage[protrusion=false,expansion=false]{microtype}
\theoremstyle{plain}
\newtheorem{theorem}{Theorem}
\newtheorem{lemma}[theorem]{Lemma}
\newtheorem{proposition}[theorem]{Proposition}
\newtheorem{corollary}[theorem]{Corollary}
\newtheorem{definition}[theorem]{Definition}
\newtheorem{remark}[theorem]{Remark}
\let\prop\proposition
\usepackage{smile_new}
\counterwithout{theorem}{section}
\theoremstyle{plain}
\usepackage[algo2e,ruled]{algorithm2e}
\usepackage{xurl}
\usepackage{hyperref}
\hypersetup{colorlinks,linkcolor=blue,citecolor=blue,urlcolor=magenta}
\usepackage{fullpage}

\bibpunct{(}{)}{;}{a}{,}{,}
\newenvironment{keywords}
  {\begin{quote}\small\noindent\textbf{Keywords:} }
  {\end{quote}}
\NewCommandCopy{\articleproof}{\proof}
\NewCommandCopy{\endarticleproof}{\endproof}
\renewenvironment{proof}[1][\proofname]
  {\Needspace{4\baselineskip}\articleproof[#1]\leavevmode}
  {\endarticleproof}
\mathtoolsset{showonlyrefs}
\newtheorem{independenttheorem}{Theorem}
\let\theorem\independenttheorem

\newcommand{\SCalg}{\mathrm{S\&C}}
\renewcommand{\E}{\mathbb E}
\newcommand{\Pp}{\mathbb P}
\newcommand{\R}{\mathbb R}
\renewcommand{\ind}{\mathbbm 1}
\newcommand{\kap}{\kappa_\nu}
\newcommand{\Tstar}{\mathcal T_\nu^\star}
\newcommand{\dd}{\mathrm{d}}
\let\argmax\relax

\DeclareMathOperator*{\argmax}{argmax}

\numberwithin{equation}{section}
\setlist[enumerate]{itemsep=3pt,topsep=4pt}
\setlist[itemize]{itemsep=3pt,topsep=4pt}
\hypersetup{pdftitle={Bayesian Fixed-Confidence Best-Arm Identification with Abstention},pdfauthor={Yuqi Huang, Yunlong Hou, Vincent Y. F. Tan}}
\title{Bayesian Fixed-Confidence Best-Arm Identification with Abstention}
\author{Yuqi Huang\textsuperscript{1}\quad
Yunlong Hou\textsuperscript{1}\quad
Vincent Y.~F.~Tan\textsuperscript{1,2} \\ \\
\textsuperscript{1}Department of Mathematics, National University of Singapore\\  \textsuperscript{2}Department of Electrical and Computer Engineering, National University of Singapore
}
\date{}
\begin{document}
\maketitle
\begin{abstract}
We study the Bayesian fixed-confidence best-arm identification problem in which a learner can abstain from making a terminal recommendation. Subject to a target error probability $\delta$ and an abstention budget $\alpha$, we 
 propose \textup{\textsc{Screen-and-Confirm}} ($\SCalg$) and analyze its expected stopping time. We further derive
the minimum expected stopping time required to satisfy both constraints for any algorithm that satisfies these constraints. 
For Gaussian priors and rewards, as $(\delta,\alpha)\downarrow(0,0)$, we establish matching upper and lower bounds up to multiplicative constants, with rate $\kappa_\nu^2[1+\log((\alpha+\delta)/\delta)]/(\alpha+\delta)$, where $\kappa_\nu$ is the prior density of the top-two gap at zero, highlighting that nearly tied instances govern the sampling cost. When abstention is disallowed ($\alpha=0$), $\SCalg$ yields
the optimal $\Theta(1/\delta)$ dependence for fixed model
parameters, removing the additional $\log(1/\delta)$ factor
in the upper bound of \citet{JKY24} under our Gaussian assumptions.
 We further characterize the exact multiplicative constant of the minimum expected stopping time for  practical regimes including (i) fixed $\alpha \in (0,1)$ and (ii) $\alpha/\delta\to\infty$, which are also achieved by $\SCalg$.
\end{abstract}

\section{Introduction}\label{sec:introduction}

Fixed-confidence best-arm identification (FC-BAI) \citep{evendar06afcbai} is a canonical pure exploration problem in sequential decision making. Given a stochastic bandit environment, a learner aims to identify
the arm with the largest mean reward using as few observations as
possible, subject to a prescribed error probability
\citep{KCG16,GK16}. This formulation describes experiments in which
observations are costly and the objective is a reliable final choice.
When the best and second-best arms have nearly equal means, however, identifying the better arm with high confidence can require too many observations \citep{GK16}. An inconclusive outcome may then be preferable:
the experiment can be deferred, referred for further investigation, or
left without a recommendation. For example, an A/B test may end without recommending a change when neither variant is clearly better. In drug screening, two closely performing compounds may be sent for further testing before selecting one.

In the classical frequentist bandit model, the means are unknown but fixed
\citep{evendar06afcbai,GK16}. We study a Bayesian model
\citep{atsidakou2023bayesbai,JKY24}: the means are drawn from a known prior,
and performance is averaged over that prior. The learner adaptively samples
arms and then recommends an arm or abstains. Prescribed budgets $\delta\in(0,1)$ and
$\alpha\in(0,1)$ constrain the probabilities of an incorrect recommendation and
abstention, respectively. The objective is to minimize the expected number of observations.

A Gaussian prior assigns positive density to instances where the
best and second-best arms have nearly equal means. Requiring
a small error probability at every instance can therefore make the
prior-averaged cost infinite~\citep{JKY24}. Bayesian constraints instead
allow the learner to distribute error and abstention budgets across instances.
In this work, we address the key question concerning both sampling and selection: \emph{how much evidence is
needed to make a decision, and which instances should be left unresolved?}

\paragraph{Contributions.}
For independent Gaussian priors and  rewards, our contributions are as follows. 
\begin{enumerate}[leftmargin=*]
\item \textbf{A Novel Algorithm.}
We introduce \textup{\textsc{Screen-and-Confirm}} ($\SCalg$) for Bayesian
FC-BAI with abstention. The algorithm combines
successive uniform screening with a confirmation
batch allocated according to conservative empirical gap estimates.

\item \textbf{Matching Rates Along Every Path.}
We derive an upper bound for the sample complexity of  $\SCalg$ and a matching information-theoretic lower bound.
This establishes the optimal expected sample size up to multiplicative
constants of $\SCalg$ along every path with $\alpha+\delta\downarrow0$:
\[
 \mathfrak{R}_\nu(\delta,\alpha):=\frac{\kap^2}{\alpha+\delta}
 \left[1+\log\frac{\alpha+\delta}{\delta}\right],
\]
where $\kap$ is the prior density of the top-two gap at zero.
Furthermore, setting $\alpha=0$ also yields matching upper and lower bounds for
the \emph{non-abstaining} Bayesian FC-BAI problem, sharpening the
sample-complexity characterization in the recent work of~\citet{JKY24}.

\item \textbf{Exact Leading Constants.}
We determine the \emph{exact} leading constants in two practical regimes:
(1) shrinking $\alpha\downarrow0$ with $\alpha/\delta\to\infty$ which we call the \textbf{vanishing abstention-dominant} regime and (2)  fixed $\alpha\in(0,1)$ with $\delta\downarrow0$ which we call the \textbf{fixed positive abstention} regime.
In each regime, the upper bound for
\textup{\textsc{Screen-and-Confirm}} matches an information-theoretic lower bound.
Subsequently, these results also characterize the case of iterated limit:
$\delta\downarrow0$ followed by $\alpha\downarrow0$,
and the power path: $\alpha=\delta^p$ for $0<p<1$.
\end{enumerate}

\subsection{Related Work}

\paragraph{Frequentist Fixed-Confidence BAI.}
Best-arm identification is a central pure-exploration problem in multi-armed
bandits. In the fixed-confidence setting, the objective is to identify the best
arm with prescribed confidence using as few samples as possible \citep{banditAlg,evendar06afcbai}.
 \citet{KCG16} provides a change-of-measure lower bound for adaptive sampling.
\citet{GK16} then identifies the characteristic time and proves asymptotic
optimality of their Track-and-Stop algorithm. We build on these ideas by developing comparisons with prior mixtures
to derive lower bounds under prior-averaged error and abstention
constraints. Bayesian sampling rules
have also been studied: \citet{Russo16} analyzes
posterior convergence for top-two rules, and \citet{Shang20} establishes
fixed-confidence guarantees for Top-Two Thompson Sampling and a simpler
variant. Their setups differ from optimizing prior-averaged cost.

\paragraph{Bayesian BAI.}
For the sample complexity of Bayesian FC-BAI without abstention, \citet{JKY24} derives a fixed-confidence lower bound
of order $\kap^2/\delta$ and an upper bound with an additional $\log(1/\delta)$ factor. In this work, we determine the joint
dependence of sample complexity on error and abstention budgets. At $\alpha=0$, our algorithm and upper bound also removes the logarithmic factor. In the fixed-budget setting, \citet{atsidakou2023bayesbai} studies
Bayesian misidentification probability, while \citet{KAKQ24} characterizes
Bayesian simple regret and the role of small gaps.

\paragraph{Abstention.}
Reject-option decision theory studies the tradeoff between errors and
inconclusive decisions \citep{Chow70,BW08}. Here the learner also controls
sampling and stopping. The Bayesian fixed-budget analysis of
\citet{HHT26} identifies a polynomial-to-exponential transition in error
probability and identifies the near-tie density $\kap$ as a fundamental limit. A deterministic
budget bound does not directly characterize the expected cost of adaptive
stopping. Our fixed-abstention result requires trimming by the full
characteristic time, and the joint limit depends on both risk budgets.

\paragraph{Connection to Erasure Decoding.} Our abstention formulation also has a natural analogue in erasure decoding in information theory. See Appendix~\ref{app:erasure}.

\section{Problem Formulation and Preliminaries}\label{sec:model}

\subsection{Model and Bayesian Risk Constraints}

We now introduce the formal model. For an integer $K\ge 2$, let $[K]:=\{1,\ldots,K\}$. The learner is given a target error probability $\delta\in(0,1)$, an abstention budget $\alpha\in[0,1)$, and a collection of $K$ arms.
Arm $i\in[K]$ has an unknown mean reward $\mu_i$. We study a Bayesian model in which the mean vector $\mu=(\mu_1,\ldots,\mu_K)$ is drawn from a known product Gaussian prior
\begin{equation}\label{eq:prior}
    P_\nu(\dd\mu)
    :=
    \bigotimes_{i=1}^K
    \mathcal N(\nu_i,\sigma_0^2)(\dd\mu_i),
\end{equation}
where the prior means $\nu_1,\ldots,\nu_K\in\mathbb R$ and prior variance (which is common to all arms)
$\sigma_0^2>0$ are fixed and known. 
We define the best arm by $a^\star=a^\star(\mu):=\argmax_{i\in[K]}\mu_i$,
which is unique $P_\nu$-almost surely.
Conditional on $\mu$, the rewards from the arms are independent (across arms and time) and normally distributed:
\[
    Y_{i,s}\mid \mu_i \sim \mathcal N(\mu_i,1),
    \qquad \mbox{for}\;s \in \mathbb{N},~i \in [K].\] 

A {\em policy} $\pi$ consists of a sampling rule, stopping rule,  and a decision rule.
At round $t$, the sampling rule of the policy selects an arm $A_t^\pi\in[K]$ and observes $X_t:=Y_{A_t^\pi,N_{A_t^\pi}(t)}$, where $N_i(t):=\sum_{u=1}^t\ind\{A_u^\pi=i\}$ is the number of pulls of arm $i$ through time $t$.
With $U$ an independent random seed, let
$\mathcal F_t:=\sigma(U,A_1^\pi,X_1,\ldots,A_t^\pi,X_t)$ and $\mathcal F_0:=\sigma(U)$.
Each $A_t^\pi$ is $\mathcal F_{t-1}$-measurable. 
The policy's stopping rule then determines if to continue sampling, or to terminate. If it terminates, the decision rule makes a terminal decision.
The \emph{sample size} $\tau^\pi$ is thus an $(\mathcal F_t)$-stopping time, and the terminal decision $D^\pi\in[K]\cup\{?\}$ is $\mathcal F_\tau$-measurable, where $?$ denotes abstention.

For a fixed instance $\mu$, $\Pp_\mu$ and $\E_\mu$ are taken over the randomness of the observations and the internal randomization of the algorithm. The \emph{Bayesian laws} average over the prior on the instance $\mu$, i.e.,  the Bayesian probability for any event $B$ is
\[
 \Pp_\nu(B)=\int\Pp_\mu(B)P_\nu(\dd\mu),\qquad
 \E_\nu[\tau]=\int\E_\mu[\tau]P_\nu(\dd\mu).
\]
Given $\delta>0$ and $\alpha\ge0$, a policy $\pi$ is \emph{$(\delta,\alpha)$-feasible}
if $\Pp_\nu(\tau^\pi<\infty)=1$ and
\begin{equation}\label{eq:budgets}
 \Pp_\nu(D^\pi\in[K],D^\pi\ne a^\star)\le\delta,
 \qquad \Pp_\nu(D^\pi=?)\le\alpha.
\end{equation}
The first event is known as \emph{undetected error event}: the learner recommends an
incorrect arm without abstaining.
Define the \emph{optimal expected sample size} as
\begin{equation}\label{eq:optimal-value}
 \Tstar(\delta,\alpha)
 :=\inf_{\text{$(\delta,\alpha)$-feasible policies } \pi}\E_\nu[\tau^\pi].
\end{equation}
In what follows, when the policy is clear from the context, we suppress the superscript $\pi$.

We consider $\alpha+\delta\le1$. A \emph{budget path} is a sequence
$((\alpha_n,\delta_n))_{n\ge1}$ satisfying
$\delta_n>0$, $\alpha_n\ge0$, and $\alpha_n+\delta_n\le1$.
Limits along a path are taken as $n\to\infty$ under the stated conditions.

\subsection{Prior Mass near Ties and the Characteristic Time}

Let $\mu_{(1)}>\cdots>\mu_{(K)}$ denote the ordered means. Define the
\emph{top-two gap} and its cumulative distribution function by
\[
 \Gamma:=\mu_{(1)}-\mu_{(2)},\qquad F_\Gamma(x):=P_\nu(\Gamma\le x).
\]
For the standard normal density and distribution function
$\phi(x):=(2\pi)^{-1/2}e^{-x^2/2}$ and
$\Phi(x):=\int_{-\infty}^x\phi(u)\,\dd u$, write the prior density and distribution function of arm $i$
\[
 f_i(x):=\frac1{\sigma_0}\phi\!\left(\frac{x-\nu_i}{\sigma_0}\right),
 \qquad F_i(x):=\Phi\!\left(\frac{x-\nu_i}{\sigma_0}\right).
\]
Define the \emph{near-tie density}
\begin{equation}\label{eq:kappa}
 \kap:=2\sum_{i<j}\int_\R
 f_i(x)f_j(x)\prod_{k\notin\{i,j\}}F_k(x)\,\dd x.
\end{equation}
It measures the prior probability mass lies near a tie between the two best arms; Lemma~\ref{geom:gap} shows that it is finite and strictly positive, and also identifies it as the density of $\Gamma$ at zero, i.e., $\lim_{x\downarrow0}\frac{P_\nu(\Gamma\le x)}{x}=\kap$. We remark that $\kappa_\nu=L(\mathbf{H})$, where $L(\mathbf{H})$ is identified as the key hardness parameter for Bayesian FC-BAI problem in \citet{JKY24}.

Given an instance $\mu$ with unique best arm and sampling proportions
$w\in\Delta_K:=\{w\in[0,1]^K:\sum_iw_i=1\}$, define the \emph{characteristic time} $ H(\mu)$ as 
\begin{align}
 H(\mu)&:=\Big[\max_{w\in\Delta_K}I(\mu,w)\Big]^{-1}  \quad\mbox{where}\quad
 I(\mu,w) :=\min_{i\ne a^\star}
 \frac{w_{a^\star}w_i}{w_{a^\star}+w_i}
 \frac{(\mu_{a^\star}-\mu_i)^2}{2},\label{eq:info}
\end{align}
A fraction with both weights zero is defined as zero. 
The characteristic time $H(\mu)$ can be understood as the reciprocal of \emph{the largest achievable
information per observation against an alternative best arm}. Under our unit-variance Gaussian reward model, it is  exactly $T^*(\mathbf{\mu})$ in \citet{GK16}.
For $a\in(0,1)$, define its $a$-upper quantile $h_a$ and $a$-trimmed expectation $V_\nu(a)$ by
\begin{equation}\label{eq:quantile}
 P_\nu(H>h_a)=a,\qquad V_\nu(a):=\E_\nu[H\ind\{H\le h_a\}].
\end{equation}
Lemma~\ref{geom:regularity} establishes the existence and uniqueness of $h_a$.
The following lemma explains the connection between the characteristic time $H(\mu)$ and the $a$-trimmed expectation $V_\nu(a)$ and shows that $\alpha V_\nu(\alpha)$ converges to $8\kap^2$ as $\alpha \downarrow0$. We note that   $V_\nu(a)$  after normalization by $1-a$  equals the lower-tail conditional mean of $H$, a lower-tail analogue of the conditional value-at-risk (CVaR) \citep{rockafellar2000optimization}.

\begin{lemma}[Prior Geometry and Optimal Trimming]\label{main:prior-geometry}
For every instance with a unique best arm,
\begin{equation}\label{eq:hardness-sandwich-intro}
 \frac8{\Gamma^2}\le H(\mu)\le\frac{4K}{\Gamma^2}.
\end{equation}
Thus, for fixed $K$, $H(\mu)$ is proportional to $\Gamma^{-2}$
up to multiplicative constants independent of $\mu$.
Furthermore,
\begin{equation}\label{eq:tail-intro}
 \lim_{t\to\infty}\sqrt t\,P_\nu(H>t)=\sqrt8\,\kap,
 \qquad \lim_{a\downarrow0}a^2h_a
 =\lim_{a\downarrow0}aV_\nu(a)=8\kap^2.
\end{equation}
\end{lemma}
\begin{proof}
These follow from Lemmas~\ref{geom:gap}, \ref{geom:hardness}, and
\ref{geom:trimming}, Proposition~\ref{geom:tail}, and
Corollary~\ref{geom:quantile} in Appendix~\ref{sec:geometry}.
\end{proof}


\section{Algorithm and Upper Bounds}\label{sec:algorithm}

\subsection{\textsc{Screen-and-Confirm} (\textsc{S\&C})}

\begin{figure}[t]
 \centering
 \includegraphics[width=\linewidth]{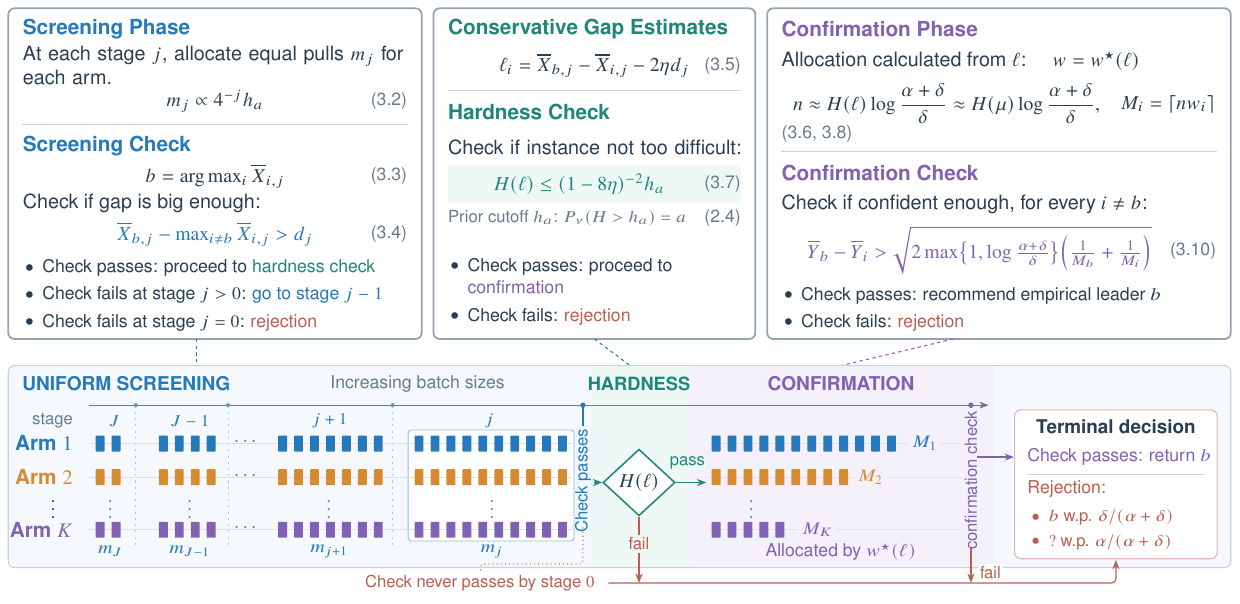}
 \caption{Overview of \textup{\textsc{Screen-and-Confirm}}.
 }
 \label{fig:screen-and-confirm}
\end{figure}

 $\SCalg$ (Algorithm~\ref{alg:Screen-and-Confirm})  first screens arms uniformly, using a geometrically increasing number of pulls over stages.
If the empirical leader is sufficiently far  from other arms, we compute the optimal allocation from
  empirical gap estimates. If the estimated characteristic time is below a   quantile, it collects one independent confirmation batch to confirm its decision. Otherwise,
or if confirmation fails, it applies a terminal randomized rule.
Figure~\ref{fig:screen-and-confirm} illustrates these steps. 

\paragraph{Calibration and Quantity Definitions.}
We first introduce the quantities used to define the number of pulls
per arm $m_j$ at each screening stage.

Define the \emph{screening accuracy parameter} $\eta$ and \emph{prior tail probability} $a$ by
\begin{equation}\label{alg:slack}
 \eta:=\frac1{10}\left[1+\log\frac{\alpha+\delta}{\delta}\right]^{-1/8},
 \qquad a:=(1-4\eta)(\alpha+\delta).
\end{equation}
We target an estimation accuracy of order $(\alpha+\delta)\eta$.
As $\alpha/\delta\to\infty$, $\eta\to0$ slowly: the mean estimates become increasingly accurate,
while the number of screening pulls remains of smaller order than the
confirmation pulls (Lemma~\ref{asym:cost-lemma}).
We will choose to abstain based on the characteristic time $H$ of the underlying instance, as larger $H$ implies larger sample size $\tau$~\citep{GK16}.
The quantity $a$ will determine the quantile $h_a$ through $P_\nu(H>h_a)=a$ in~\eqref{eq:quantile}.
It specifies the prior mass of the hardest instances that we should abstain or make error on, and leaves a slack of
$4\eta(\alpha+\delta)$ for
screening inaccuracies and confirmation failures.

Define the \emph{finest screening threshold} $d_0$ by
\begin{equation}\label{alg:cutoff}
 d_0:=\frac{\sqrt{8/h_a}}{1+2\eta}
\end{equation}
The threshold $d_0$ sets the smallest gap scale examined by screening.
It ensures that, if every reached stage is accurate,
failure at stage zero implies $H(\mu) \ge h_a$
(Lemma~\ref{screen:lemma}).

Screening uses geometric thresholds
\begin{equation}\label{alg:scales}
 J:=\max\{0,\lceil\log_2(1/d_0)\rceil\},\qquad d_j:=2^jd_0.
\end{equation}
The stages run from $J$ down to $0$. After a failed check at stage $j>0$,
the next threshold is $d_{j-1}=d_j/2$, allowing progressively smaller
gaps to be resolved. At stage $j$, the target estimation accuracy is of order $d_j\eta \asymp2^j(\alpha+\delta)\eta$.
Figure~\ref{fig:screen-and-confirm} illustrates this progression in its lower timeline.

Denote the normalization $Z$ and \emph{stage error allowance} $q_j$ as
\begin{equation}\label{alg:D}
 q_j:=\min\left\{\frac14,\frac{\eta(\alpha+\delta)}{Z4^j}\right\},
 \qquad \mbox{where}\quad
 Z:=\min\left\{\frac43,\,4^{-J}
       +\frac{3K(K-1)d_0}{\sigma_0\sqrt\pi}\right\}.
\end{equation}
The allowance $q_j$ controls the  probability of inaccurate estimation at stage $j$; and is
also used to choose the confirmation batch size $n$, so that the confirmation check (after an accurate screening) fails with probability at most $q_j$.
The normalization $Z$ accounts for the probabilities of reaching the
different stages. The choices of $q_j$ and $Z$ bound the probability of inaccurate screening and failed confirmation by $2\eta(\alpha+\delta)$
(Lemmas~\ref{risk:screening} and~\ref{risk:confirmation}).

Finally, the number of fresh observations per arm at stage $j$ is
\begin{equation}\label{alg:blocks}
 m_j:=\bigg\lceil\frac{2}{\eta^2d_j^2}
    \log\frac{2K}{q_j}\bigg\rceil.
\end{equation}
Its inverse-square dependence on $d_j$ means that each halving of the
threshold quadruples the batch size, up to lower order terms.
Crucially, if all stages are accurate, reaching a finer
stage $j<J$ requires $\Gamma\le3d_j$ (Lemma~\ref{screen:lemma}).
The prior probability of this event is at most
$3K(K-1)d_j/(2\sigma_0\sqrt\pi)$ (Lemma~\ref{geom:gap}).
Thus the more expensive screening stages are reached mainly on nearly
tied instances, whose prior probability is small.

\paragraph{Screening Phase.} (Figure~\ref{fig:screen-and-confirm}, left panel)
At each stage $j \in \{J,J-1,\ldots,0\}$, sample every arm $m_j$ times. Let \(Y_{i,j,s}\) denote the \(s\)-th fresh observation from arm \(i\) at screening stage \(j\). Define each arm's
\emph{stage mean} and empirical leader of the stage by
\begin{equation}\label{alg:screen-mean}
 \overline X_{i,j}:=\frac1{m_j}\sum_{s=1}^{m_j}Y_{i,j,s},\qquad
 b_j:=\argmax_{i\in[K]}\overline X_{i,j}.
\end{equation}
We abbreviate $b_j$ as $b$ in stage $j$.
The stage \emph{succeeds} when
\begin{equation}\label{alg:pass}
 \overline X_{b,j}-\max_{i\ne b}\overline X_{i,j}>d_j.
\end{equation}

\paragraph{Hardness Check and Allocation.} (Figure~\ref{fig:screen-and-confirm}, middle panel)
At the first successful stage, define the \emph{conservative gap estimates}:
\begin{equation}\label{alg:lowergaps}
 \ell_i:=\overline X_{b,j}-\overline X_{i,j}-2\eta d_j,
 \qquad i\ne b.
\end{equation}
For these estimated gaps, define the \emph{artificial} instance and allocation by
\begin{equation}\label{alg:gap-instance}
 \mu_b^\ell:=0,\quad \mu_i^\ell:=-\ell_i\;\; (i\ne b),\qquad
 H(\ell):=H(\mu^\ell),\quad
 w^\star(\ell)\in\argmax_{w\in\Delta_K}I(\mu^\ell,w).
\end{equation}
Set $w=w^\star(\ell)$. We proceed for confirmation if the \emph{hardness check} passes: 
\begin{equation} \label{hardness_check_formula}
    H(\ell)\le(1-8\eta)^{-2}h_a.
\end{equation}

\paragraph{Confirmation Phase.} (Figure~\ref{fig:screen-and-confirm}, right panel)
If screening succeeds and hardness check passes, pull each arm $M_i$ times, where
\begin{equation}\label{alg:allocation}
 M_i:=\lceil nw_i\rceil\quad\mbox{where}\quad n:=\Big\lceil H(\ell)
       \big(\sqrt\beta+\sqrt{\log((K-1)/q_j)}\big)^2\Big\rceil.
\end{equation}
The confirmation sample counts are determined by \(w^\star(\ell)\), as illustrated in the lower timeline of Figure~\ref{fig:screen-and-confirm}. Here we define the \emph{confirmation confidence
parameter} $
 \beta:=\max\left\{1,\log\frac{\alpha+\delta}{\delta}\right\}$, so that the confirmation make a mistake with probability at most
$e^{-\beta}\le\delta/(\alpha+\delta)$ (Lemma~\ref{risk:wrong}).

Let $Y_{i,s}$ denote the $s$-th fresh confirmation
observation from arm $i$. Define
\begin{equation}\label{alg:confirmation-mean}
 \overline Y_i:=\frac1{M_i}\sum_{s=1}^{M_i}Y_{i,s}.
\end{equation}
Conditional on $\mu$ and the screening history, these observations
are independent, with $Y_{i,s}\sim\mathcal N(\mu_i,1)$.
The \emph{confirmation check} (as in the upper-right panel of
Figure~\ref{fig:screen-and-confirm}) \emph{succeeds} if
\begin{equation}\label{alg:certificate}
 \overline Y_b-\overline Y_i>
 \sqrt{2\beta(1/M_b+1/M_i)}\qquad\text{for every }i\ne b.
\end{equation}
We recommend the empirical best arm $b$ if confirmation check succeeds.

\paragraph{Terminal Decision.}
If screening never succeeds, or the hardness check \eqref{hardness_check_formula} fails , or confirmation
fails, apply the following terminal rule to the last screening candidate:
\begin{equation}\label{alg:terminal}
 D:=\begin{cases}
 b,&U\le\delta/(\alpha+\delta),\\
 ?,&U>\delta/(\alpha+\delta),
 \end{cases}\qquad U\sim\operatorname{Unif}(0,1),
\end{equation}
where $U$ is independent of the means and all sampling data and is used
only by the terminal rule.
Entry into this rule is called a \emph{rejection}; the final output may
be either a recommendation or abstention. The red arrows in Figure~\ref{fig:screen-and-confirm} shows these routes in red and the terminal rule at the lower right.

\begin{algorithm2e}[t]
\caption{\textup{\textsc{Screen-and-Confirm}}}\label{alg:Screen-and-Confirm}
\DontPrintSemicolon
\KwIn{Known prior; $\delta>0$, $\alpha\ge0$, $\alpha+\delta\le1$.}
Compute $\eta$, $a$, $d_j$, $\beta$, $Z$, $q_j$, $m_j$ and $h_a$ using \eqref{alg:slack}--\eqref{alg:blocks}\;
\For(\tcp*[f]{Screening Stages}){$j=J,J-1,\ldots,0$}{
 Collect $m_j$ new observations per arm; compute the empirical mean $\overline X_{i,j}$ from this stage \eqref{alg:screen-mean}\;
 Set this stage's empirical leader $b\gets\argmax_i\overline X_{i,j}$, breaking ties by index\;
 \If(\tcp*[f]{Screening check}){ $\overline X_{b,j}-\max_{i\ne b}\overline X_{i,j}>d_j$}{
  Compute the conservative gap estimates $\ell$ using \eqref{alg:lowergaps}\;
  Compute $H(\ell)$ and $w^\star(\ell)$   per \eqref{alg:gap-instance}\;
  \If(\tcp*[f]{Hardness check}){ $H(\ell)>(1-8\eta)^{-2}h_a$}{Stop and apply the terminal rule\;} 
  Compute $M_i$ using \eqref{alg:allocation}\;
  Collect $M_i$ new observations per arm; compute the empirical means $\overline Y_i$ from these confirmation observations \eqref{alg:confirmation-mean}\;
  \eIf(\tcp*[f]{Confirmation}){ $\overline Y_b-\overline Y_i>
 \sqrt{2\beta(1/M_b+1/M_i)}$ for every $i\ne b$}{Return $b$\;}{Stop and apply the terminal rule\;}
 }
}
Apply the terminal rule \eqref{alg:terminal} with the last candidate $b$\;
\smallskip
\textbf{Return} $b$ with probability $\delta/(\alpha+\delta)$;
otherwise return $?$\tcp*[f]{Terminal rule}
\end{algorithm2e}

\paragraph{Computation.} We compute $w$ and $H(\ell)$ by solving monotone scalar equations via the bisection  method (Appendix~\ref{comp:root}).
Appendix~\ref{comp:quantile} describes Monte Carlo estimation
and a small-$a$ approximation for the prior quantile $h_a$.


\subsection{Feasibility and Upper Bounds Analysis}

\begin{theorem}[Guarantees for \textsc{Screen-and-Confirm}]\label{thm:upper-rate}
For every $\delta>0$, $\alpha\ge0$ with $\alpha+\delta\le1$, \textsc{Screen-and-Confirm} is $(\delta,\alpha)$-feasible and has a deterministically bounded stopping time. Let $\tau_{\delta,\alpha}^{\SCalg}$ be \textsc{Screen-and-Confirm}'s stopping time.
\begin{enumerate}[label=(\roman*)]
    \item (\textbf{ Uniform upper bound}) There exists $C=C(K,\nu,\sigma_0)<\infty$ such that 
\begin{equation}\label{main:uniform-upper}
\E_\nu[\tau_{\delta,\alpha}^{\SCalg}]
 \le C \cdot \mathfrak{R}_\nu(\delta,\alpha).
\end{equation}

\item (\textbf{Fixed positive abstention}) For every fixed $\alpha\in(0,1)$,
\begin{equation}\label{eq:main-fixed}
 \limsup_{\delta\downarrow0}
 \frac{\E_\nu[\tau_{\delta,\alpha}^{\SCalg}]}{\log(1/\delta)}\le V_\nu(\alpha).
\end{equation}

\item (\textbf{Vanishing abstention-dominant regime}) Along every path with $\alpha\downarrow0$ and $\alpha/\delta\to\infty$,
\begin{equation}\label{eq:main-separated}
 \limsup_{ \alpha,\delta\downarrow0, ~\alpha/\delta\to\infty}
 \frac{\alpha\E_\nu[\tau_{\delta,\alpha}^{\SCalg}]}{\log(\alpha/\delta)}\le8\kap^2.
\end{equation}
In particular, for every fixed $p\in(0,1)$, when $\alpha=\delta^p$, 
\begin{equation}\label{eq:upper-power-value}
 \limsup_{\delta\downarrow0}
 \frac{\delta^p\E_\nu[\tau_{\delta,\delta^p}^{\SCalg}]}{\log(1/\delta)}
 \leq 8(1-p)\kap^2.
\end{equation}
\end{enumerate}
\end{theorem}
The following are consequences of Theorem \ref{thm:upper-rate}.

\noindent
$\bullet$
\textbf{No abstention: closing the logarithmic gap.}
Setting $\alpha=0$ in~\eqref{main:uniform-upper} yields
$\E_\nu[\tau_{\delta,0}^{\SCalg}]
 \le C \cdot\mathfrak{R}(\delta,0)=O(\kap^2/\delta)$  
for fixed model parameters. Under our Gaussian assumptions, our upper bound removes the additional $\log(1/\delta)$ factor in the upper bound of
\citet[Theorem~6]{JKY24}, matching the order of their lower bound
and closing the logarithmic gap for Bayesian FC-BAI without abstention. The rate expression involves the squared near-tie density
$\kappa_\nu^2$, 
reflecting that prior mass on the instances whose two largest arm means are nearly equal \emph{dominate} the expected sample size. These instances feature large characteristic times $H(\mu)$ and  consume large sample size if the best arm needs to be identified. Thus, the abstention budget $\alpha$ is  used on these instances.

\noindent
$\bullet$
\textbf{Fixed positive abstention: logarithmic sample complexity.} For every   fixed $\alpha\in(0,1)$, \eqref{eq:main-fixed} guarantees an
expected sample size of $O(\log(1/\delta))$ as $\delta\downarrow0$,
with leading upper bound coefficient $V_\nu(\alpha)$.
This coefficient is the contribution to the prior-averaged
characteristic time from easy instances, which together
have prior probability $1-\alpha$.
Here, difficulty is measured by $H(\mu)$. It captures the benefit of allowing the learner to stop and abstain on difficult instances without collecting additional
observations needed to identify the best arm reliably. Without abstention, the lower bound of \citet{JKY24} requires an expected sample size of $\Omega(1/\delta)$ for fixed model parameters. Thus, any fixed positive abstention budget makes logarithmic sampling cost achievable, whereas the no-abstention problem requires at least order $1/\delta$ observations in expectation. This mirrors the polynomial-to-exponential transition in error decay shown by \citet{HHT26} in the fixed-budget setting.


\noindent
$\bullet$
\textbf{Vanishing abstention: intermediate rates.} For $\alpha=\delta^p$ with fixed $p\in(0,1)$, \eqref{eq:upper-power-value} gives an expected sample size of $O(\delta^{-p}\log(1/\delta))$  as $\delta\downarrow0$. Thus, substantial savings remain possible even when both budgets vanish.

These upper bounds describe the sampling costs achievable by
$\SCalg$. The matching lower bounds in
Section~\ref{sec:lower-main} establish rate optimality and
show that the leading constants are sharp in the fixed positive
abstention and vanishing abstention-dominant regimes;
see Corollary~\ref{thm:rate}.

\emph{Proof Sketch of Theorem~\ref{thm:upper-rate}.} For $(\delta,\alpha)$-feasibility: If a successful screening stage $j$ selects an incorrect candidate
$b\ne a^\star$, the reached stage must have inaccurate estimates, in the sense that $\max_i|\overline X_{i,j}-\mu_i|>\eta d_j$.
This event has probability at most $\eta(\alpha+\delta)$
(Lemmas~\ref{risk:screening} and~\ref{screen:lemma}).
Conditional on the instance and screening history, an incorrect
candidate accepted for confirmation passes with probability at most
$e^{-\beta}\le\delta/(\alpha+\delta)$.
Integrating this conditional bound gives probability at most
$\eta\delta$ of an incorrect certified recommendation
(Lemma~\ref{risk:wrong}).
On an accurate screen, rejection before
confirmation requires $H\ge h_a$, whose prior probability is $a$;
confirmation failure contributes at most another $\eta(\alpha+\delta)$
(Lemmas~\ref{screen:lemma} and~\ref{risk:confirmation}). Also recall the probability of inaccurate screening is at most $\eta(\alpha+\delta)$, hence the terminal
rule is entered with probability at most
$a+2\eta(\alpha+\delta)=(1-2\eta)(\alpha+\delta)$.
The terminal randomization gives total error at most
$\eta\delta+(1-2\eta)\delta\le\delta$ and abstention at most
$(1-2\eta)\alpha\le\alpha$. 
Appendix~\ref{sec:risk} gives the complete proof.

For \eqref{main:uniform-upper},  we note that 
a finer stage $j$ can be reached only if $\Gamma\le3d_j$ or an
earlier reached stage was inaccurate (Lemma~\ref{screen:lemma}).
The small-gap bound $P_\nu(\Gamma\le x)\le D_\Gamma x$
(Lemma~\ref{geom:gap}) and the screening-error bound therefore
control the probability of reaching each stage. Weighting the
stage costs (pulls) by these probabilities gives the expected screening
cost (Lemma~\ref{cost:geometric-lemma}).
After a successful stage $j$, the \emph{estimated characteristic time}
can be bounded by $H(\ell)\le C_8/d_j^2$. The same stage probabilities
therefore also control the expected confirmation cost
(Lemma~\ref{cost:hardness-lemma}). Combining the two costs gives
$\E_\nu[\tau_{\delta,\alpha}^{\SCalg}]
 \le \frac{C_{13}}{\alpha+\delta}
 \left[
 \eta^{-2}\bigl(1+\log(1/\eta)\bigr)+\beta
 \right]$
(Proposition~\ref{cost:master-proposition}), where $C_{13}:=C_9+3C_{10}+K+1$.
The  choices of $\eta$ and $\beta$ bound the bracket
by a multiple of
$1+\log((\alpha+\delta)/\delta)$.
See Appendix~\ref{sec:cost}.

For \eqref{eq:main-fixed} and \eqref{eq:main-separated},
as $\alpha/\delta\to\infty$, the expected screening cost,
rounding, and contribution from
inaccurate screening are all of lower order than
$\beta/(\alpha+\delta)$. Lemma~\ref{asym:cost-lemma} gives
\[
 \E_\nu[\tau_{\delta,\alpha}^{\SCalg}]
 \le
 \frac{\beta}{(1-8\eta)^2}
 \E_\nu\!\left[
 H\cdot\ind\!\left\{H\le\frac{h_a}{(1-8\eta)^2}\right\}
 \right]
 +o\!\left(\frac{\beta}{\alpha+\delta}\right).
\] 
(ii) For fixed $\alpha$, $\frac{h_a}{(1-8\eta)^2}$ tends to $h_\alpha$, so the
expectation tends to $V_\nu(\alpha)$, while
$\beta/\log(1/\delta)\to1$. (iii) When $\alpha\downarrow0$ and $\alpha/\delta\to\infty$,
we have $a/\alpha\to1$, $\eta\to 0$ and
$\beta/\log(\alpha/\delta)\to1$. 
Appendix~\ref{sec:asymptotics} proves each limit in detail.

\section{Lower Bounds}\label{sec:lower-main}


The following information-theoretic lower bounds apply to any adaptive randomized policy satisfying
the Bayesian risk constraints. We first establish the rate for arbitrary budget ratios and then obtain the exact leading constants in practical regimes.


\begin{theorem}[Impossibility]\label{thm:general-lower-main}
There is a universal constant $c>0$ and a threshold $\varepsilon = \varepsilon(K,\nu,\sigma_0)>0$ such that for every $\delta>0$, $\alpha\ge0$:
\begin{enumerate}[label=(\roman*)]
    \item  (\textbf{Uniform lower bound}) Whenever $\alpha+\delta<\varepsilon$,
\begin{equation}\label{eq:general-lower-main}
 \Tstar(\delta,\alpha)
 \ge c\cdot \mathfrak{R}_\nu(\delta,\alpha).
\end{equation}

\item (\textbf{Fixed positive abstention}) Furthermore, for every fixed $\alpha\in(0,1)$,
\begin{equation}\label{eq:sharp-lower-fixed-main}
 \liminf_{\delta\downarrow0}
 \frac{\Tstar(\delta,\alpha)}{\log(1/\delta)}\ge V_\nu(\alpha).
\end{equation}

\item (\textbf{Vanishing abstention-dominant regime}) Along every path with $\alpha\downarrow0$ and $\alpha/\delta\to\infty$,
\begin{equation}\label{eq:sharp-lower-separated-main}
 \liminf_{ \alpha,\delta\downarrow0, ~\alpha/\delta\to\infty}
 \frac{\alpha\Tstar(\delta,\alpha)}{\log(\alpha/\delta)}\ge8\kap^2.
\end{equation}
 In particular, for every fixed $p\in(0,1)$, when $\alpha=\delta^p$,
\begin{equation}\label{eq:lower-power-value}
 \liminf_{\delta\downarrow0}
 \frac{\delta^p\Tstar(\delta,\delta^p)}{\log(1/\delta)}
 \geq 8(1-p)\kap^2.
\end{equation}

\end{enumerate}

\end{theorem}

\emph{Proof Sketch.} 
For \eqref{eq:general-lower-main}, Lemma~\ref{lb:flat-subprior} uses the flat local subprior construction of \citet[Appendix~C.1]{HHT26}
to obtain a subprior whose density is constant in the signed top-two gap. We restrict it to a near-tie gap interval of width proportional to
$(\alpha+\delta)/\kap$. Normalize the positive-gap and negative-gap parts of this subprior
to obtain two distributions over instances, each with unnormalized
mass $4(\alpha+\delta)$. Under the positive-gap distribution,
feasibility forces the policy to recommend the best arm with
probability at least $3/4$. Under the paired negative-gap
distribution, that same arm is incorrect, so its recommendation
probability is at most $\delta/[4(\alpha+\delta)]$.
In Proposition~\ref{lower:all-path}, a change-of-measure inequality converts this
separation into \eqref{eq:general-lower-main}.

For \eqref{eq:sharp-lower-fixed-main} and for fixed $\alpha$,
Theorem~\ref{lower:fixed} compares each instance $\mu$ with
alternative instances in a small prior region $B$ where $\mu$'s best
arm is not optimal. These alternatives are drawn according to
$P_\nu(\cdot\mid B)$. Since recommending $\mu$'s best arm is an
error throughout $B$, the Bayesian error constraint bounds the
average probability under these alternatives by $\delta/P_\nu(B)$.
This comparison then yields an instance-wise bound on $\E_\mu[\tau^\pi]$ using a change-of-measure inequality (Lemma~\ref{lower:mixture-lemma}), which is then
integrated against $P_\nu$.
Feasibility requires total correct-recommendation probability at
least $1-\alpha-\delta$. Trimming by Lemma~\ref{geom:trimming}
therefore yields $V_\nu(\alpha)$ as $\delta\downarrow0$.

For \eqref{eq:sharp-lower-separated-main},  as $\alpha \downarrow0$, given any $(\delta,\alpha)$-feasible FC policy, design a fixed budget (FB) algorithm by truncating the FC policy at time $T$ and abstain
if $\tau>T$. This gives error at most $\delta$ and abstention at most
$\alpha+\Pp_\nu(\tau>T)$.
A finite-horizon form of the Bayesian fixed-budget lower-bound
argument of \citet{HHT26}, established in
Lemma~\ref{lb:finite-horizon}, then forces a lower bound on
$\Pp_\nu(\tau>T)$; otherwise, the
truncated policy would have both error and abstention too small
for its sampling horizon (Lemma~\ref{lb:finite-horizon}).
Theorem~\ref{lower:separated} integrates these tail bounds over $T$
to obtain \eqref{eq:sharp-lower-separated-main}.

Appendix~\ref{sec:lower-proofs} gives the full proofs.

\begin{corollary}[Optimality of \textsc{Screen-and-Confirm}]
\label{thm:rate}

\begin{enumerate}[label=(\roman*)]
\item (\textbf{Uniform bounds}) \textup{\textsc{Screen-and-Confirm}} is rate-optimal for all sufficiently small $(\delta,\alpha)$, i.e., there exist a universal constant $c>0$, a constant
$C=C(K,\nu,\sigma_0)<\infty$, and a threshold
$\varepsilon=\varepsilon(K,\nu,\sigma_0)>0$ such that, for every $\delta>0$, $\alpha\ge0$, whenever $\alpha+\delta<\varepsilon$,
\begin{equation}\label{eq:main-global}
 \begin{aligned}
 c\cdot \mathfrak{R}_\nu(\delta,\alpha) 
 \le\Tstar(\delta,\alpha)\le\E_\nu[\tau_{\delta,\alpha}^{\SCalg}]
 \le C\cdot \mathfrak{R}_\nu(\delta,\alpha). 
 \end{aligned}
\end{equation}

\item (\textbf{Fixed positive abstention}) For every fixed $\alpha\in(0,1)$,
\begin{equation}\label{eq:main-lower-fixed}
 \lim_{\delta\downarrow0}\frac{\Tstar(\delta,\alpha)}{\log(1/\delta)}
 =\lim_{\delta\downarrow0}\frac{\E_\nu[\tau_{\delta,\alpha}^{\SCalg}]}{\log(1/\delta)}
 =V_\nu(\alpha).
\end{equation}
Subsequently, taking $\delta\downarrow0$ first and then $\alpha\downarrow0$,
\begin{equation}\label{eq:main-iterated-value}
 \lim_{\alpha\downarrow0}\alpha
 \bigg[\lim_{\delta\downarrow0}
 \frac{\Tstar(\delta,\alpha)}{\log(1/\delta)}\bigg]
 =\lim_{\alpha\downarrow0}\alpha
 \bigg[\lim_{\delta\downarrow0}
 \frac{\E_\nu[\tau_{\delta,\alpha}^{\SCalg}]}{\log(1/\delta)}\bigg]
 =8\kap^2.
\end{equation}
 
\item (\textbf{Vanishing abstention-dominant regime}) Along every path with $\alpha\downarrow0$ and $\alpha/\delta\to\infty$,
\begin{equation}\label{eq:main-lower-separated}
 \lim_{\alpha,\delta\downarrow0,~\alpha/\delta\to\infty}
 \frac{\alpha\Tstar(\delta,\alpha)}{\log(\alpha/\delta)}
 = \lim_{\alpha,\delta\downarrow0,~\alpha/\delta\to\infty}
 \frac{\alpha\E_\nu[\tau_{\delta,\alpha}^{\SCalg}]}{\log(\alpha/\delta)}
 =8\kap^2.
\end{equation}
In particular, for every fixed $p\in(0,1)$, when $\alpha=\delta^p$,
\begin{equation}\label{eq:main-power-value}
 \lim_{\delta\downarrow0}
 \frac{\delta^p\Tstar(\delta,\delta^p)}{\log(1/\delta)}
 =\lim_{\delta\downarrow0}
 \frac{\delta^p\E_\nu[\tau_{\delta,\delta^p}^{\SCalg}]}{\log(1/\delta)}
 =8(1-p)\kap^2.
\end{equation}

\end{enumerate}

\end{corollary}

Corollary~\ref{thm:rate} shows that the achievable rates discussed in
Theorem~\ref{thm:upper-rate} cannot be improved in order, and that the leading constants are \emph{sharp} in the fixed positive abstention and
vanishing abstention-dominant regimes.

We further compare our result under the Bayesian setup with the \textbf{frequentist} guarantees in Appendix~\ref{sec:frequentist}. In particular,  we show that, for every fixed
$\alpha\in[0,1)$, the optimal expected sample complexity under
uniform frequentist constraints is asymptotic to
$(1-\alpha)H(\mu)\log(1/\delta)$ as $\delta\downarrow0$.
This leading constant is attained by abstaining initially with
probability $\alpha$ and otherwise running an asymptotically
optimal FC-BAI policy with error budget $\delta/(1-\alpha)$.
Fixed abstention therefore improves the classical frequentist
benchmark of \citet{GK16} only by the factor $1-\alpha$.
The distinction lies in the constraints: the frequentist
abstention probability is bounded by $\alpha$ at every instance,
whereas the Bayesian constraint averages over the prior and
allows abstention to concentrate on difficult instances.
Detailed discussions are presented at Appendix~\ref{sec:frequentist}.

\section{Numerical Experiments}\label{sec:experiments}

We illustrate the benefits of abstention in
Bayesian FC-BAI. We compare the sampling
cost of $\SCalg$ at several abstention
budgets with a no abstention baseline. We also report the
empirical error and abstention probabilities.

\paragraph{Experimental Setup.}
We use the Gaussian prior and reward model, with $K=5$,
$\sigma_0=1$, and $\nu=(5,5,3,3,2)$. The arm means are drawn
independently as $\mu_i\sim\mathcal N(\nu_i,1)$, and rewards satisfy
$Y_{i,s}\mid\mu_i\sim\mathcal N(\mu_i,1)$.
We evaluate S\&C with $\alpha\in\{10^{-1}, 10^{-2},10^{-3}\}$ and $\delta\in\{10^{-2},3\times 10^{-3}, 10^{-3}, 3\times 10^{-4}, 10^{-4}, 3\times 10^{-5}, 10^{-5}\}$.
We also compare it with a no-abstention baseline:
Successive Elimination with Early-Stopping from \citet{JKY24}, denoted as
SE-early.

Each setting uses $M=10^5$ independent prior-generated trials.
Let $\tau_m$, $D_m$, and
$a_m^\star$ denote the sample size, terminal decision, and true best arm
in trial $m$; we report the following empirical mean sample size, error probability and abstention probability for each $\alpha$ and $\delta$
\[
\begin{aligned}
 \widehat T=\frac1M\sum_{m=1}^M\tau_m,\qquad
 \widehat E=\frac1M\sum_{m=1}^M\ind\{D_m\notin\{a_m^\star,?\}\},\qquad
 \widehat A=\frac1M\sum_{m=1}^M\ind\{D_m=?\}.
 \end{aligned}
\]

\begin{figure}[t]
 \centering

\includegraphics[width=\linewidth]{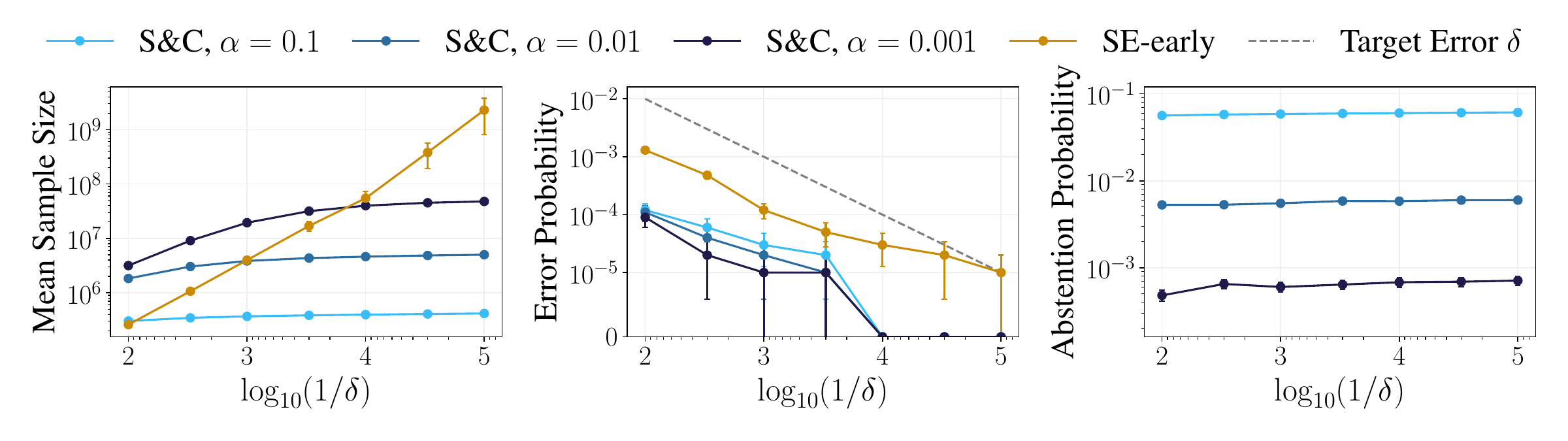}
 \caption{Effect of abstention on Bayesian FC-BAI.
 The panels show $\widehat{T}$ (left), $\widehat{E}$
 (middle) and $\widehat{A}$ (right) as the target error
 probability $\delta$ decreases. Error bars
 show $1\times$ standard error.}
 \label{fig:abstention-comparison}
\end{figure}

\paragraph{Benefit of Abstention.}
Figure~\ref{fig:abstention-comparison} (left) illustrates the reduction in
sampling cost obtained by allowing abstention.
At every tested $\delta$, increasing $\alpha$ reduces the observed mean
sample size of S\&C. 

Crucially, the mean sample size of SE-early increases rapidly as
$\delta$ decreases, at the order of $1/\delta$. In contrast, the mean sample size of S\&C grows much more slowly (in fact, logarithmically) once
$\delta$ becomes small relative to $\alpha$. This behavior is consistent
with the theoretical sample-complexity scale 
\[
 \frac{1+\log((\alpha+\delta)/\delta)}{\alpha+\delta}
 \sim \frac{1+\log(\alpha/\delta)}{\alpha},
 \qquad \delta\downarrow0
\]
for fixed $\alpha>0$. In this regime, the abstention budget controls
the contribution of nearly tied instances, avoiding the $1/\delta$
sample cost associated with requiring a recommendation on every instance.
Tightening the error requirement therefore incurs only logarithmic
growth in the rate.

The middle and right panels of Figure~\ref{fig:abstention-comparison} show that the empirical error and abstention
probabilities satisfy the given $\delta$ and $\alpha$ constraint.
In the middle panel, the gray dashed line marks the target error
probability $\delta$, and the empirical error probabilities of all
algorithms lie below this line at every tested $\delta$.
The right panel shows that the empirical abstention probability
remains below $\alpha$ for every tested $\delta$.

\section{Conclusion}\label{sec:conclusion}

We characterize the optimal expected sample complexity of Bayesian
fixed-confidence best-arm identification with abstention under
independent Gaussian priors and Gaussian rewards.
Our \textup{\textsc{Screen-and-Confirm}} algorithm achieves the optimal
rate up to constants along every path with $\alpha+\delta\downarrow0$,
and matches the exact leading constants for fixed positive abstention
and for shrinking abstention with $\alpha/\delta\to\infty$.
For fixed $\alpha$, the optimal coefficient $V_\nu(\alpha)$ corresponds
to retaining the easiest prior fraction $1-\alpha$, as measured by
the characteristic time $H$. In the shrinking-abstention regime,
the near-tie density $\kap$ determines the sharp constant $8\kap^2$.

In particular, any fixed positive abstention budget, however small,
reduces the optimal dependence on $\delta$ from $1/\delta$ to
$\log(1/\delta)$. This improvement is possible because Bayesian
constraints allow abstention to concentrate on difficult instances.
Under uniform frequentist constraints, by comparison, fixed abstention
improves only the leading constant, and vanishing abstention gives
no leading-order improvement. Our numerical experiments illustrate
the resulting reductions in sampling cost under Bayesian constraints.
Extending these results to non-Gaussian rewards and dependent priors
is a natural direction for future work.

\section*{AI Disclosure}
 The authors used Anthropic’s Claude Code (Opus 5.5) and OpenAI’s GPT-6 Astra during the development and preparation of this manuscript. These systems assisted in developing and refining mathematical arguments, exploring proof strategies, and drafting and revising the exposition. The author reviewed the resulting material and assumes full responsibility for the contents of the final manuscript. 

\label{main:last-page}

\bibliography{references}

\clearpage\appendix

\section{Prior Distribution of the Characteristic Time}
\label{geom:section}
\label{sec:geometry}

This appendix proves the properties of $H$, $h_a$, and $V_\nu(a)$ that will be later. Throughout, the prior of the arms are independent with
distributions $N(\nu_i,\sigma_0^2)$, and rewards have unit variance.
Write $f_i$ and $F_i$ for the prior density and distribution function
of coordinate $i$.

Order the means as $\mu_{(1)}\ge\cdots\ge\mu_{(K)}$, so that
$\Gamma=\mu_{(1)}-\mu_{(2)}$. All means are distinct $P_\nu$-almost
surely. For $K\ge3$, also define the gap to the third-best arm by
\[
 \Delta_{(3)}=\mu_{(1)}-\mu_{(3)}.
\]

\subsection{The Density of the Top-Two Gap}

\begin{lemma}[Density and Probability of a Small Gap]\label{geom:gap}
The top-two gap $\Gamma$ has a continuous, strictly positive
density $f_\Gamma$ on $(0,\infty)$, with
$\lim_{x\downarrow0}f_\Gamma(x)=\kappa_\nu>0$.
Defining $f_\Gamma(0):=\kappa_\nu$, its density formula on $[0,\infty)$ is given by
\begin{equation}\label{geom:gap-density}
 f_\Gamma(x)=\sum_{i\ne j}\int_\R
 f_i(y+x)f_j(y)\prod_{k\notin\{i,j\}}F_k(y)\,\dd y,
 \qquad x\ge0.
\end{equation}
In particular,
\begin{equation}\label{geom:kappa-formula}
 \kap=f_\Gamma(0)
 =2\sum_{i<j}\int_\R
 f_i(y)f_j(y)\prod_{k\notin\{i,j\}}F_k(y)\,\dd y
 \in(0,\infty).
\end{equation}
With $D_\Gamma:=K(K-1)/(2\sigma_0\sqrt\pi)$, for every $x\ge0$,
\begin{equation}\label{geom:density-bound}
 0<f_\Gamma(x)\le D_\Gamma,
 \qquad
 P_\nu(\Gamma\le x)\le D_\Gamma x.
\end{equation}
Moreover,
\begin{equation}\label{geom:gap-remainder}
 \begin{aligned}
 |f_\Gamma(x)-\kap|
 &\le\frac{K(K-1)}{\pi\sigma_0^2}x,\\
 \left|P_\nu(\Gamma\le x)-\kap x\right|
 &\le\frac{K(K-1)}{2\pi\sigma_0^2}x^2.
 \end{aligned}
\end{equation}
Consequently,
\begin{equation}\label{geom:gap-small}
 \lim_{x\downarrow0}\frac{P_\nu(\Gamma\le x)}{x}=\kap.
\end{equation}
\end{lemma}

\begin{proof}

\textbf{Step 1: The Density Formula.}\quad
Since ties have prior probability zero, we partition the event
$\{\Gamma\le x\}$ according to the best arm $i$ and second-best
arm $j$. If $\mu_j=y$, this event requires
\[
 y<\mu_i\le y+x,
 \qquad
 \mu_k<y\quad\text{for every }k\notin\{i,j\}.
\]
By independence, the probability that all remaining means are
below $y$ is $\prod_{k\notin\{i,j\}}F_k(y)$. Thus, for $x\ge0$,
\begin{align*}
 F_\Gamma(x)
 &=\sum_{i\ne j}\int_\R
 f_j(y)\prod_{k\notin\{i,j\}}F_k(y)
 \left[\int_0^x f_i(y+u)\,\dd u\right]\dd y\\
 &=\int_0^x
 \left[
 \sum_{i\ne j}\int_\R
 f_i(y+u)f_j(y)\prod_{k\notin\{i,j\}}F_k(y)\,\dd y
 \right]\dd u,
\end{align*}
where the second equality follows from Tonelli's theorem.
The expression in brackets is therefore a density of $\Gamma$,
giving \eqref{geom:gap-density}. For $K=2$, the empty product
equals one.

For each pair, the integrand is continuous in $u\ge0$ and bounded by
$f_j(y)/(\sigma_0\sqrt{2\pi})$, which is integrable in $y$.
Dominated convergence therefore shows that $f_\Gamma$ is continuous
on $(0,\infty)$ and that
\[
 \lim_{x\downarrow0}f_\Gamma(x)
 =2\sum_{i<j}\int_\R
 f_i(y)f_j(y)\prod_{k\notin\{i,j\}}F_k(y)\,\dd y
 =\kappa_\nu,
\]
where the factor two arises because the terms for $(i,j)$ and
$(j,i)$ coincide at zero.
All factors in the integrands are strictly positive, so
$f_\Gamma(x)>0$ for $x>0$ and $\kappa_\nu>0$.
Defining $f_\Gamma(0):=\kappa_\nu$ makes
\eqref{geom:gap-density} valid also at $x=0$.

\paragraph{Step 2: The Density Bound.}
Since each distribution function is at most one,
\begin{align*}
 f_\Gamma(x)
 &\le\sum_{i\ne j}\int_\R f_i(y+x)f_j(y)\,\dd y\\
 &=\sum_{i\ne j}\frac{1}{2\sigma_0\sqrt\pi}
   \exp\!\left(-\frac{(x-\nu_i+\nu_j)^2}{4\sigma_0^2}\right)\\
 &\le\frac{K(K-1)}{2\sigma_0\sqrt\pi}=D_\Gamma.
\end{align*}
Integrating over $[0,x]$ gives the probability bound in
\eqref{geom:density-bound}.

\paragraph{Step 3: The Remainder at Zero.}
The fundamental theorem of calculus and Tonelli's theorem give
\[
 \int_\R|f_i(y+x)-f_i(y)|\,\dd y
 \le\int_0^x\int_\R|f_i'(y+u)|\,\dd y\,\dd u
 =x\|f_i'\|_1.
\]
For a $N(\nu_i,\sigma_0^2)$ density,
\[
 \|f_i'\|_1=\frac{\sqrt{2/\pi}}{\sigma_0},
 \qquad
 \|f_i\|_\infty=\frac{1}{\sigma_0\sqrt{2\pi}}.
\]
Subtracting \eqref{geom:gap-density} at zero from its value at $x$
therefore yields
\begin{align*}
 |f_\Gamma(x)-\kap|
 &\le\sum_{i\ne j}\|f_j\|_\infty
       \int_\R|f_i(y+x)-f_i(y)|\,\dd y\\
 &\le\frac{K(K-1)}{\pi\sigma_0^2}x.
\end{align*}
Integrating this bound proves the second inequality in
\eqref{geom:gap-remainder}. Dividing that inequality by $x$ and
letting $x\downarrow0$ proves \eqref{geom:gap-small}.
\end{proof}

\subsection{Bounds and Continuity of the Characteristic Time}

\begin{sloppypar}
For an instance with unique best arm $b$, let
$\Delta_i=\mu_b-\mu_i$ for $i\ne b$. Recall that $\Delta_K:=\left\{w\in[0,1]^K:\sum_{i=1}^K w_i=1\right\}$, and
\end{sloppypar}
\begin{equation}\label{geom:H-definition}
 \begin{aligned}
 I(\mu,w):=\min_{i\ne b}
 \frac{w_bw_i}{w_b+w_i}\frac{\Delta_i^2}{2}, 
 \qquad
 H(\mu):=\left(\max_{w\in\Delta_K}I(\mu,w)\right)^{-1}.
 \end{aligned}
\end{equation}
The fraction is defined to be zero when $w_b=w_i=0$. Set
$H=\infty$ when the best arm is not unique, this event has prior
probability zero.

\begin{lemma}[Properties of the Characteristic Time]\label{geom:hardness}
For every unique-best instance, the maximum in
\eqref{geom:H-definition} is attained, and every maximizing
allocation has strictly positive weights. Moreover,
\begin{equation}\label{geom:gap-sandwich}
 \frac{8}{\Gamma^2}\le H(\mu)\le\frac{4K}{\Gamma^2}.
\end{equation}
In particular, for $K=2$, $H(\mu)=8/\Gamma^2$.

\noindent Regarded as a function of the positive gap vector, i.e., $H$ is a function of $\Delta:=(\mu_b-\mu_1,...,\mu_b-\mu_K)$, $H$ satisfies
\begin{equation}\label{geom:gap-perturbation}
 H(r\Delta)=r^{-2}H(\Delta)\ \forall r>0,
 \qquad
 H(\Delta)\le H(\ell)\le c^{-2}H(\Delta)
\end{equation}
whenever $c\Delta_i\le\ell_i\le\Delta_i$ for every $i\ne b$,
where $0<c\le1$. The function $H$ is continuous on each region
$\{\mu:\mu_b>\max_{i\ne b}\mu_i\}$.
\end{lemma}

\begin{proof}
\textbf{Step 1: Existence and Positivity of an Optimal Allocation.}\quad
The function $uv/(u+v)$ is continuous on $[0,\infty)^2$ under
the zero-denominator convention: at $(0,0)$ this follows from
$0\le uv/(u+v)\le\min\{u,v\}$. Thus $I(\mu,w)$ is continuous
in $w$ on the compact simplex, and its maximum is attained.
For uniform weights,
\[
 I(\mu,(1/K,\ldots,1/K))=\frac{\Gamma^2}{4K}>0.
\]
This proves $H(\mu)\le4K/\Gamma^2$, and any weight $w_i$ of a optimal allocation cannot be zero.

\paragraph{Step 2: The Lower Bound on $H$.}
Let $i$ be the second-best arm. For every $w\in\Delta_K$,
\[
 \frac{w_bw_i}{w_b+w_i}
 \le\frac{w_b+w_i}{4}\le\frac14,
\]
where the first inequality follows from
$4w_bw_i\le(w_b+w_i)^2$. Consequently,
\[
 I(\mu,w)
 \le\frac{w_bw_i}{w_b+w_i}\frac{\Gamma^2}{2}
 \le\frac{\Gamma^2}{8}.
\]
Taking the maximum and then its reciprocal proves the lower
bound in \eqref{geom:gap-sandwich}. When $K=2$, the two bounds coincide.

\paragraph{Step 3: Comparison of Gap Vectors.}
Directly from the definition,
\[
 H(r\Delta)^{-1}
 =\max_{w\in\Delta_K}\min_{i\ne b}
   \frac{w_bw_i}{w_b+w_i}\frac{r^2\Delta_i^2}{2}
 =r^2H(\Delta)^{-1}.
\]
For positive vectors $u,v$ with $u_i\le v_i$ for every $i\ne b$,
each allocation satisfies
\[
 \min_{i\ne b}\frac{w_bw_i}{w_b+w_i}\frac{u_i^2}{2}
 \le
 \min_{i\ne b}\frac{w_bw_i}{w_b+w_i}\frac{v_i^2}{2}.
\]
Maximizing both sides over the same simplex and taking reciprocals
gives $H(u)\ge H(v)$.
It follows that
\[
 H(\Delta)\le H(\ell)\le H(c\Delta)=c^{-2}H(\Delta),
\]
proving \eqref{geom:gap-perturbation}.

Also, since $H$ depends only on
the gaps, adding a common constant to all means leaves it unchanged.

\paragraph{Step 4: Continuity.}
Fix $\mu$ with unique best arm $b$. The best arm remains $b$ in
a sufficiently small neighborhood of $\mu$. For $\lambda$ in
that neighborhood,
\[
 \sup_{w\in\Delta_K}|I(\mu,w)-I(\lambda,w)|
 \le\frac18\max_{i\ne b}
       |(\mu_b-\mu_i)^2-(\lambda_b-\lambda_i)^2|.
\]
Here we used $|\min_i u_i-\min_i v_i|\le\max_i|u_i-v_i|$ and $\frac{w_bw_i}{w_b+w_i}
 \le\frac14$.
The right side tends to zero as $\lambda\to\mu$. For every \(w\),
\[
I(\mu,w)
\le I(\lambda,w)+\sup_v|I(\mu,v)-I(\lambda,v)|
\le \max_v I(\lambda,v)+\sup_v|I(\mu,v)-I(\lambda,v)|,
\]
Taking the maximum over \(w\) gives
\[
 \left|\max_w I(\mu,w)-\max_w I(\lambda,w)\right|
 \le\sup_w|I(\mu,w)-I(\lambda,w)|.
\]
Hence the maximum information 
$\max_{w\in\Delta_K} I(\mu,w)$
is continuous. It is strictly positive
on the unique-best region, so its reciprocal $H$ is continuous there.
\end{proof}

\subsection{The Distribution and Quantiles of the Characteristic Time}

\begin{lemma}[Distribution and Quantiles of $H$]\label{geom:regularity}
The random variable $H$ is finite and strictly positive
$P_\nu$-almost surely. It has no atoms, and
$P_\nu(u<H<v)>0$ whenever $0<u<v<\infty$.
For each $a\in(0,1)$ there is therefore a unique $h_a\in(0,\infty)$
satisfying
\begin{equation}\label{geom:quantile-definition}
 P_\nu(H>h_a)=a.
\end{equation}
The function $a\mapsto h_a$ is continuous and strictly decreasing,
with
\[
 \lim_{a\downarrow0}h_a=\infty,
 \qquad
 \lim_{a\uparrow1}h_a=0.
\]
The function
\begin{equation}\label{geom:V-definition}
 V_\nu(a)=\E_\nu[H\ind\{H\le h_a\}]
\end{equation}
is finite, strictly positive, continuous, and nonincreasing on
$(0,1)$. It satisfies
$V_\nu(a)\le(1-a)h_a$ and $\lim_{a\uparrow1}V_\nu(a)=0$.
\end{lemma}

\begin{proof}
\textbf{Step 1: $H$ Has No Atoms.}\quad
Finiteness and positivity of $H$ follow from the almost-sure uniqueness
of the best arm and \eqref{geom:gap-sandwich}.

Fix $h>0$, an arm $b$, and the other means $(\mu_i)_{i\ne b}$.
On the interval $\mu_b>\max_{i\ne b}\mu_i$, the function $H(\mu)$
is strictly decreasing in $\mu_b$. Indeed, increasing $\mu_b$
strictly increases every comparison information at the original
optimal allocation, whose weights are all positive
(Lemma~\ref{geom:hardness}). The maximum information therefore
strictly increases, and its reciprocal $H$ strictly decreases.

Consequently, $H(\mu)=h$ has at most one solution for $\mu_b$
on this interval. Conditional on the other means, $\mu_b$ has
density $f_b$, so
\[
 P_\nu\!\left(H=h,\ a^\star=b
       \,\middle|\,(\mu_i)_{i\ne b}\right)=0.
\]
Integrating over the other means and summing over $b\in[K]$
gives $P_\nu(H=h)=0$.

\paragraph{Step 2: Every Positive Interval Has Positive Probability.}
Fix $0<u<v$ and $h\in(u,v)$. For any unique-best vector $z$, let
$r=\sqrt{H(z)/h}$. Then $H(rz)=h$. By continuity there is an open
ball around $rz$ consisting of unique-best instances with $H\in(u,v)$.
The prior density is strictly positive everywhere, so that ball has
positive prior probability.

\paragraph{Step 3: The Quantiles.}
Since $P_\nu(H=t)=0$ for every $t>0$, the indicators
$\ind\{H>s\}$ converge to $\ind\{H>t\}$ almost surely as
$s\to t$. Dominated convergence therefore shows that
$t\mapsto P_\nu(H>t)$ is continuous on $(0,\infty)$.
Moreover, for $0<u<v$, Step 2 gives
\[
 P_\nu(H>u)-P_\nu(H>v)
 =P_\nu(u<H\le v)
 \ge P_\nu(u<H<v)>0,
\]
so the function $t\mapsto P_\nu(H>t)$ is strictly decreasing.
Since $0<H<\infty$
almost surely, its limits at zero and infinity are one and zero,
respectively. Hence it takes every value $a\in(0,1)$ exactly once.
To prove continuity of $a\mapsto h_a$, fix $a\in(0,1)$ and
$0<\varepsilon<h_a$. Strict decrease of the survival function gives
\[
 P_\nu(H>h_a+\varepsilon)
 <a<
 P_\nu(H>h_a-\varepsilon).
\]
If $a_n\to a$, then $a_n$ eventually lies between these same
two bounds. Since $P_\nu(H>h_{a_n})=a_n$, strict decrease implies
\[
 h_a-\varepsilon<h_{a_n}<h_a+\varepsilon.
\]
As $\varepsilon$ can be arbitrarily small, $h_{a_n}\to h_a$.

The inverse is strictly decreasing because a larger tail
probability requires a smaller threshold. Finally, for every
fixed $t>0$, we have $0<P_\nu(H>t)<1$. Thus $h_a>t$ for all
sufficiently small $a$, and $h_a<t$ for $a$ sufficiently close
to one. Since $t>0$ is arbitrary, this proves
\[
 \lim_{a\downarrow0}h_a=\infty,
 \qquad
 \lim_{a\uparrow1}h_a=0.
\]

\paragraph{Step 4: The Trimmed Expectation.}
The event $\{H\le h_a\}$ has probability $1-a$, and on this event values
of $H$ are positive and bounded by $h_a$. Therefore
$0<V_\nu(a)\le(1-a)h_a$. Take any sequence $a_n\to a\in(0,1)$, the quantile
continuity and atomlessness imply
\[
 H\ind\{H\le h_{a_n}\}\longrightarrow H\ind\{H\le h_a\}
 \quad P_\nu\text{-almost surely}.
\]
The cutoffs $h_{a_n}$ are bounded for sufficiently large $n$,
so dominated convergence gives continuity of $V_\nu$. Its
monotonicity follows from the decreasing cutoffs. Finally,
$V_\nu(a)\le h_a\to0$ as $a\uparrow1$.
\end{proof}

\begin{lemma}[Variational Characterization of the Trimmed Expectation]
\label{geom:trimming}
For every $a\in(0,1)$,
\begin{equation}\label{geom:trimming-variational}
 V_\nu(a)=\inf\left\{\E_\nu[q(\mu)H(\mu)]:
 0\le q\le1,\ q\text{ measurable},\ \E_\nu[q]\ge1-a\right\}.
\end{equation}
The infimum is attained by $q(\mu)=\ind\{H(\mu)\le h_a\}$.
In particular, $V_\nu(a)=(1-a)\E_\nu[H\mid H\le h_a]$.
\end{lemma}

\begin{proof}
For any $q$ satisfying the constraints in \eqref{geom:trimming-variational},
\[
 \bigl(q-\ind\{H\le h_a\}\bigr)(H-h_a)\ge0.
\]
since both factors are nonpositive on $\{H\le h_a\}$ and
nonnegative on its complement. If $\E_\nu[qH]=\infty$, the
required lower bound is immediate. Otherwise, integrating and
using $P_\nu(H\le h_a)=1-a$ gives
\begin{align*}
 \E_\nu[qH]-V_\nu(a)
 &=\E_\nu\!\left[
   \bigl(q-\ind\{H\le h_a\}\bigr)(H-h_a)\right]
   +h_a\bigl(\E_\nu[q]-(1-a)\bigr)\\
 &\ge0.
\end{align*}
This proves \eqref{geom:trimming-variational}. 

And, it is easy to verify that the stated indicator $\ind\{H(\mu)\le h_a\}$ satisfies the constraints in
\eqref{geom:trimming-variational} and its expected cost is $V_\nu(a)$. Finally, since $P_\nu(H\le h_a)=1-a$,
\[
 \E_\nu[H\mid H\le h_a]
 =\frac{\E_\nu[H\ind\{H\le h_a\}]}{P_\nu(H\le h_a)}
 =\frac{V_\nu(a)}{1-a},
\]
which proves the final identity in the statement.
\end{proof}

\subsection{The Characteristic Time when the Third-Best Arm is Separated}

\begin{lemma}[An Allocation for a Separated Third-Best Arm]
\label{geom:isolated-pair}
Suppose $K\ge3$ and the best arm is unique. Write $\Delta_j=\mu_{(1)}-\mu_{(j)}$ for $j\ge3$, and let
\[
 \rho=2\Gamma^2\sum_{j=3}^K\Delta_j^{-2}.
\]
If $\rho<1$, then
\begin{equation}\label{geom:isolated-bound}
 \frac{8}{\Gamma^2}\le H(\mu)
 \le\frac{8}{(1-\rho)\Gamma^2}.
\end{equation}
In particular, if $\Delta_{(3)}\ge \varepsilon>0$ and
$2(K-2)\Gamma^2/\varepsilon^2<1$, then
\begin{equation}\label{geom:isolated-uniform}
 1\le\frac{H(\mu)\Gamma^2}{8}
 \le\left(1-\frac{2(K-2)\Gamma^2}{\varepsilon^2}\right)^{-1}.
\end{equation}
\end{lemma}

\begin{proof}
Choose
\[
 w_j=\frac{2\Gamma^2}{\Delta_j^2}\quad(3\le j\le K),
 \qquad
 w_1=w_2=\frac{1-\rho}{2}.
\]
These weights are positive and sum to one; $\rho$ is the total
weight assigned to arms $3,\ldots,K$. The comparison between the
best and second-best arms has information
\[
 \frac{w_1w_2}{w_1+w_2}\frac{\Gamma^2}{2}
 =\frac{(1-\rho)\Gamma^2}{8}.
\]
For $j\ge3$, substitution of $w_j$ gives
\begin{align*}
 \frac{w_1w_j}{w_1+w_j}\frac{\Delta_j^2}{2}
 &=\frac{w_1\Gamma^2}{w_1+w_j}\\
 &\ge w_1\Gamma^2
 =\frac{(1-\rho)\Gamma^2}{2}
 \ge\frac{(1-\rho)\Gamma^2}{8},
\end{align*}
because $w_1+w_j\le1$. Thus the minimum information of this
allocation is $(1-\rho)\Gamma^2/8$. Maximizing over allocations
and taking the reciprocal proves the upper bound in
\eqref{geom:isolated-bound}; its lower bound is from
\eqref{geom:gap-sandwich}.

If $\Delta_{(3)}\ge \varepsilon$, then
$\Delta_j\ge \varepsilon$ for every $j\ge3$, and hence
$\rho\le2(K-2)\Gamma^2/\varepsilon^2<1$. This gives
\eqref{geom:isolated-uniform} uniformly over all such instances.
\end{proof}

\begin{lemma}[Probability That the Third-Best Arm Is Close]
\label{geom:triple-volume}
For $K\ge3$ and every $x,\varepsilon>0$,
\begin{equation}\label{geom:triple-bound}
 P_\nu(\Gamma\le x,\ \Delta_{(3)}\le \varepsilon)
 \le\frac{K(K-1)(K-2)}{2\pi\sigma_0^2}x\varepsilon.
\end{equation}
In particular,
$P_\nu(\Delta_{(3)}\le \varepsilon)
\le K(K-1)(K-2)\varepsilon^2/(2\pi\sigma_0^2)$.
\end{lemma}

\begin{proof}
Fix distinct indices $i,j,k$. If these are the best, second-best,
and third-best arms, respectively, then the two gap conditions ($\Gamma\le x,\ \Delta_{(3)}\le \varepsilon$) imply
\[
 \mu_j\in[\mu_i-x,\mu_i],
 \qquad
 \mu_k\in[\mu_i-\varepsilon,\mu_i].
\]
For an upper bound, we can drop the requirements that $\mu_j>\mu_k$ or
$\mu_\ell<\mu_k$ for $\ell\notin\{i,j,k\}$.
Conditioning on $\mu_i=y$ and using independence between the arm means $\mu$ therefore gives
\begin{align*}
 &P_\nu\bigl(i,j,k\text{ are the first three arms in order},
              \ \Gamma\le x,\ \Delta_{(3)}\le \varepsilon\bigr)\\
 &\quad\le\int_\R f_i(y)
       \left(\int_{y-x}^y f_j(z)\,\dd z\right)
       \left(\int_{y-\varepsilon}^y f_k(z)\,\dd z\right)\,\dd y\\
 &\quad\le\int_\R f_i(y)
       \frac{x}{\sigma_0\sqrt{2\pi}}
       \frac{\varepsilon}{\sigma_0\sqrt{2\pi}}\,\dd y
 =\frac{x\varepsilon}{2\pi\sigma_0^2}.
\end{align*}
Summing over the $K(K-1)(K-2)$ ordered triples proves
\eqref{geom:triple-bound}. The final assertion follows by taking
$x=\varepsilon$, since $\Gamma\le\Delta_{(3)}$.
\end{proof}

\subsection{The Tail, Quantile, and Trimmed-Expectation Limits}

\begin{proposition}[Tail Probability of the Characteristic Time]
\label{geom:tail}
The characteristic time satisfies
\begin{equation}\label{geom:H-tail}
 \lim_{t\to\infty}\sqrt t\,P_\nu(H(\mu)>t)=\sqrt8\,\kap.
\end{equation}
\end{proposition}

\begin{proof}
\textbf{Step 1: The Lower Bound.}\quad
By \eqref{geom:gap-sandwich},
\[
 \left\{\Gamma<\frac{\sqrt8}{\sqrt t}\right\}
 \subseteq\{H(\mu)>t\}.
\]
Using \eqref{geom:gap-small} at $x=\sqrt8/\sqrt t$ gives
\begin{equation}\label{geom:tail-liminf}
 \liminf_{t\to\infty}\sqrt t\,P_\nu(H>t)
 \ge\sqrt8\,\kap.
\end{equation}
For $K=2$, $H=8/\Gamma^2$, so the same calculation gives
equality. Assume $K\ge3$ for the remaining steps.

\paragraph{Step 2: Instances with a Small Third-Best Gap.}
Fix $\varepsilon>0$. The upper bound in \eqref{geom:gap-sandwich} implies
\begin{equation}\label{geom:tail-gap-implication}
 \{H>t\}\subseteq
 \left\{\Gamma<\frac{2\sqrt K}{\sqrt t}\right\}.
\end{equation}
Consequently, \eqref{geom:triple-bound} gives
\begin{equation}\label{geom:tail-triple-part}
 \begin{aligned}
 P_\nu(H>t,\ \Delta_{(3)}<\varepsilon)
 &\le P_\nu\!\left(
       \Gamma\le\frac{2\sqrt K}{\sqrt t},\ \Delta_{(3)}\le \varepsilon
       \right)\\
 &\le\frac{K(K-1)(K-2)\sqrt K}{\pi\sigma_0^2}
       \frac{\varepsilon}{\sqrt t}.
 \end{aligned}
\end{equation}

\paragraph{Step 3: Instances with a Separated Third-Best Arm.}
On $\{H>t,\ \Delta_{(3)}\ge\varepsilon\}$, the bound
$H\le4K/\Gamma^2$ implies $\Gamma^2<4K/t$.
Thus, for $t>8K(K-2)/\varepsilon^2$,
\[
 \frac{2(K-2)\Gamma^2}{\varepsilon^2}
 <\frac{8K(K-2)}{t\varepsilon^2}<1.
\]
\eqref{geom:isolated-uniform} then gives
\[
 H
 \le\frac{8}{\Gamma^2}
 \left(1-\frac{2(K-2)\Gamma^2}{\varepsilon^2}\right)^{-1}
 \le\frac{8}{\Gamma^2}
 \left(1-\frac{8K(K-2)}{t\varepsilon^2}\right)^{-1}.
\]
It follows that for $t>8K(K-2)/\varepsilon^2$,
\[
 \{H>t,\ \Delta_{(3)}\ge \varepsilon\}
 \subseteq
 \left\{\Gamma<
 \frac{\sqrt8}{\sqrt t\sqrt{1-8K(K-2)/(t\varepsilon^2)}}\right\}.
\]
For fixed $\varepsilon$, as $t\to\infty$ the threshold on the right tends to $\sqrt{8/t}$. Applying
\eqref{geom:gap-small} yields
\begin{equation}\label{geom:tail-isolated-part}
 \limsup_{t\to\infty}\sqrt t\,
 P_\nu(H>t,\ \Delta_{(3)}\ge \varepsilon)\le\sqrt8\,\kap.
\end{equation}
For every $\varepsilon>0$, splitting according to whether
$\Delta_{(3)}<\varepsilon$ gives
\[
 P_\nu(H>t)
 =P_\nu(H>t,\ \Delta_{(3)}<\varepsilon)
  +P_\nu(H>t,\ \Delta_{(3)}\ge\varepsilon).
\]
Multiplying by $\sqrt t$, taking the upper limit, and applying
the bounds from Steps 2 and 3 yields
\begin{align*}
 \limsup_{t\to\infty}\sqrt t\,P_\nu(H>t)
 &\le
 \limsup_{t\to\infty}\sqrt t\,
 P_\nu(H>t,\ \Delta_{(3)}<\varepsilon)+
 \limsup_{t\to\infty}\sqrt t\,
 P_\nu(H>t,\ \Delta_{(3)}\ge\varepsilon)\\
 &\le
 \frac{K(K-1)(K-2)\sqrt K}{\pi\sigma_0^2}\varepsilon
 +\sqrt8\,\kap.
\end{align*}
This holds for every $\varepsilon>0$. Letting $\varepsilon\downarrow0$ and combining
with \eqref{geom:tail-liminf} proves \eqref{geom:H-tail}.
\end{proof}

\begin{corollary}[Quantile and Truncated-Expectation Limits]
\label{geom:quantile}
For $0<a<1$,
\begin{equation}\label{geom:quantile-global}
 h_a\le\frac{4KD_\Gamma^2}{a^2},
\end{equation}
and
\begin{equation}\label{geom:quantile-asymptotic}
 \lim_{a\downarrow0}a^2h_a=8\kap^2.
\end{equation}
Furthermore,
\begin{align}
 \lim_{t\to\infty}
 \frac{\E_\nu[H\ind\{H\le t\}]}{\sqrt t}
 &=\sqrt8\,\kap,\label{geom:truncated-moment}\\
 \lim_{a\downarrow0}aV_\nu(a)&=8\kap^2.
 \label{geom:V-asymptotic}
\end{align}
The function $aV_\nu(a)/(8\kap^2)$ is bounded on $(0,1)$
and extends continuously to zero with value one.
\end{corollary}

\begin{proof}
\textbf{Step 1: The Quantile Bound and Limit.}\quad
Equations \eqref{geom:gap-sandwich} and \eqref{geom:density-bound} give
\[
 a=P_\nu(H>h_a)
 \le P_\nu\!\left(\Gamma<\frac{2\sqrt K}{\sqrt{h_a}}\right)
 \le\frac{2\sqrt K D_\Gamma}{\sqrt{h_a}},
\]
which proves \eqref{geom:quantile-global}. Since $h_a\to\infty$
as $a\downarrow0$, \eqref{geom:H-tail} also gives
\[
 \lim_{a\downarrow0}a\sqrt{h_a}
 =\lim_{a\downarrow0}\sqrt{h_a}\,P_\nu(H>h_a)
 =\sqrt8\,\kap.
\]
Squaring proves \eqref{geom:quantile-asymptotic}.

\paragraph{Step 2: The Truncated Expectation.}
Pointwise,
\[
 \int_0^t\ind\{H>x\}\,\dd x
 =\min\{H,t\}
 =H\ind\{H\le t\}+t\ind\{H>t\}.
\]
Taking expectations and applying Tonelli's theorem yields
\begin{equation}\label{geom:layer-cake}
 \E_\nu[H\ind\{H\le t\}]
 =\int_0^tP_\nu(H>x)\,\dd x-tP_\nu(H>t).
\end{equation}
For each $\varepsilon\in(0,1)$, \eqref{geom:H-tail} allows us to pick
$T(\varepsilon)<\infty$ such that, for all $t\ge T(\varepsilon)$,
\[
 (1-\varepsilon)\frac{\sqrt8\,\kap}{\sqrt t}
 \le P_\nu(H>t)
 \le(1+\varepsilon)\frac{\sqrt8\,\kap}{\sqrt t}.
\]
Thus for $t\ge T(\varepsilon)$, integrate the above from $T(\varepsilon)$ to $t$, and bounding $\int_0^{T(\varepsilon)}P_\nu(H>x)\,\dd x$ by $T(\varepsilon)$ gives
\begin{align*}
 2(1-\varepsilon)\sqrt8\,\kap
   (\sqrt t-\sqrt{T(\varepsilon)})
 &\le\int_0^tP_\nu(H>x)\,\dd x\\
 &\le T(\varepsilon)+
 2(1+\varepsilon)\sqrt8\,\kap
   (\sqrt t-\sqrt{T(\varepsilon)}).
\end{align*}
Divide by $\sqrt t$, let $t\to\infty$, and then let
$\varepsilon\downarrow0$. This gives
\[
 \lim_{t\to\infty}\frac{1}{\sqrt t}
       \int_0^tP_\nu(H>x)\,\dd x=2\sqrt8\,\kap.
\]
On the other hand,
$tP_\nu(H>t)/\sqrt t\to\sqrt8\,\kap$ by
\eqref{geom:H-tail}. Subtracting these limits in
\eqref{geom:layer-cake} proves \eqref{geom:truncated-moment}.

\paragraph{Step 3: The Trimmed Expectation.}
Since $h_a\to\infty$ as $a\downarrow0$,
\[
 aV_\nu(a)
 =\left(a\sqrt{h_a}\right)
   \frac{\E_\nu[H\ind\{H\le h_a\}]}{\sqrt{h_a}}
 \longrightarrow(\sqrt8\,\kap)^2=8\kap^2.
\]
Here the first factor converges by \eqref{geom:quantile-asymptotic} and the second by \eqref{geom:truncated-moment}.
This proves \eqref{geom:V-asymptotic}
\end{proof}

\section{Proof of the Error and Abstention Guarantees}\label{sec:risk}

We first show what the screen guarantees when its block means are
accurate. We then control the prior probability of any inaccurate
reached stage. The final step bounds contribution from the confirmation
batch independently.

For the analysis, assume all screening blocks of samples are generated in advance, independently
conditional on $\mu$, and leave unused blocks unobserved. With the empirical 
leaders $b_j$ from \eqref{alg:screen-mean}, define the event that \emph{stage $j$ is reached} as
\begin{equation}\label{screen:reach-event}
 \mathsf R_j:=\bigcap_{k=j+1}^J
 \left\{\overline X_{b_k,k}-\max_{i\ne b_k}\overline X_{i,k}\le d_k\right\}.
\end{equation}
An empty intersection is the whole sample space, so stage $J$ is always
reached. Define the event that \emph{every reached screen stage is accurate} by
\begin{equation}\label{screen:event}
 \mathcal E:=\bigcap_{j=0}^J
 \left(\mathsf R_j^c\ \cup\
 \left\{\max_i|\overline X_{i,j}-\mu_i|\le\eta d_j\right\}\right).
\end{equation}
The pre-generated observations make every block mean well-defined, including
those at unreached stages. The final screening stage and its candidate are
\begin{equation}\label{screen:final-stage}
 \widehat\jmath:=\min\{j\in\{0,\ldots,J\}:\mathsf R_j\text{ occurs}\},
 \qquad b:=b_{\widehat\jmath}.
\end{equation}
Set $\ell_i=1$, $w_i=1/K$, and $n=M_i=1$ whenever the corresponding
quantity is not computed.

\begin{lemma}[Consequences of an Accurate Screen]\label{screen:lemma}
On $\mathcal E$, a reached successful stage has candidate $b=a^\star$.
Writing $\Delta_i:=\mu_{a^\star}-\mu_i$, its gap estimates satisfy
\begin{equation}\label{screen:gaps}
 (1-8\eta)\Delta_i\le\ell_i\le\Delta_i,\qquad i\ne a^\star,
\end{equation}
and therefore
\begin{equation}\label{screen:hardness}
 H(\mu)\le H(\ell)\le (1-8\eta)^{-2} H(\mu).
\end{equation}
For $j<J$, if stage $j$ is reached and all preceding reached stages
were accurate, then
\begin{equation}\label{screen:reach}
 \Gamma\le(1+2\eta)d_{j+1}\le3d_j.
\end{equation}
Finally, on $\mathcal E$, rejection before confirmation implies
$H(\mu)\ge h_a$. Every instance with $H(\mu)<h_a$ is accepted for
confirmation on $\mathcal E$.
\end{lemma}

\begin{proof}
\textbf{Step 1: A Successful Accurate Stage Identifies the Best Arm.}\quad
If its leader $b$ were wrong, accuracy would give
\[
 \overline X_{b,j}-\overline X_{a^\star,j}
 \le\mu_b-\mu_{a^\star}+2\eta d_j<2\eta d_j<d_j.
\]
This contradicts \eqref{alg:pass}, since the true best is among the
challengers. Hence $b=a^\star$.

\paragraph{Step 2: The Recorded Gaps Are Relatively Accurate.}
On $\mathcal E$, the error of each empirical mean difference is at most $2\eta d_j$.
Subtracting the margin in \eqref{alg:lowergaps} gives
\[
 \Delta_i-4\eta d_j\le\ell_i\le\Delta_i.
\]
Success also gives $\ell_i>(1-2\eta)d_j$, and thus
$\Delta_i>(1-2\eta)d_j$, $d_j<\Delta_i/(1-2\eta)$. Consequently
\[
 \ell_i\ge\left(1-\frac{4\eta}{1-2\eta}\right)\Delta_i
 \ge(1-8\eta)\Delta_i,
\]
where the final inequality uses \(\eta\le1/10\), which implies \(1-2\eta\ge1/2\), so
$\frac{4\eta}{1-2\eta}\le8\eta$.
The characteristic time decreases when any gap increases and satisfies
$H(c\Delta)=c^{-2}H(\Delta)$ for $c>0$
(Lemma~\ref{geom:hardness}). Thus
\[
 H(\Delta)\le H(\ell)
 \le H((1-8\eta)\Delta)
 =(1-8\eta)^{-2}H(\Delta),
\]
which proves \eqref{screen:hardness}.

\paragraph{Step 3: Reaching a Finer Stage Implies a Small True Gap.}
At an accurate stage $k$, each estimate differs from its true
mean by at most $\eta d_k$. Hence, for every $i\ne a^\star$,
\[
 \overline X_{a^\star,k}-\overline X_{i,k}
 \ge(\mu_{a^\star}-\eta d_k)-(\mu_i+\eta d_k)
 \ge\Gamma-2\eta d_k.
\]
If $\Gamma>(1+2\eta)d_k$, every displayed difference is strictly
larger than $d_k$. Thus $a^\star$ is the empirical leader and stage $k$ would have succeeded as per \eqref{alg:pass}.

Now suppose stage $j<J$ is reached and all preceding reached
stages were accurate. Stage $j+1$ must have failed, since otherwise
screening would have stopped there. By the preceding implication,
its failure and accuracy require
\[
 \Gamma\le(1+2\eta)d_{j+1}
 =2(1+2\eta)d_j
 \le3d_j,
\]
where the last inequality uses $\eta\le1/10$.
This proves \eqref{screen:reach}.

\paragraph{Step 4: An Accurate Rejection Has Large Characteristic Time.}
If all stages fail accurately, then
$\Gamma\le(1+2\eta)d_0=\sqrt{8/h_a}$.
The left inequality in \eqref{eq:hardness-sandwich-intro} implies
$H(\mu)\ge h_a$. If a successful accurate screen instead fails
the hardness check, \eqref{screen:hardness} gives
$(1-8\eta)^{-2} H(\mu)\ge H(\ell)>(1-8\eta)^{-2} h_a$, again implying
$H(\mu)>h_a$. Hence, on $\mathcal E$, every $H(\mu)<h_a$ must reach
confirmation. The boundary $H=h_a$ has prior probability zero.
\end{proof}

\begin{lemma}[Probability of an Inaccuracy]
\label{risk:screening}
Over the screening phase,
\begin{equation}\label{risk:bad}
 \Pp_\nu(\mathcal E^c)\le\eta {(\alpha+\delta)}.
\end{equation}
\end{lemma}
\begin{proof}
Fix $\mu$. Using the independent sample blocks generated in advance,
the stage-$j$ means $\overline X_{i,j}$ are independent
$N(\mu_i,1/m_j)$ variables and are independent of the screening
history before stage $j$. The Gaussian tail bound and a union
bound therefore give
\begin{equation}\label{risk:stage-tail}
 \Pp_\mu\!\left(
   \max_i|\overline X_{i,j}-\mu_i|>\eta d_j
 \right)
 \le 2K\exp\!\left(-\frac{m_j\eta^2d_j^2}{2}\right)
 \le q_j.
\end{equation}
Let $B_j$ be the event that stage $j$ is reached and every
preceding reached stage was accurate. For fixed $\mu$, this
event is determined by the history before stage $j$. Independence
of the fresh sample block and \eqref{risk:stage-tail} therefore give
\[
 \Pp_\mu\!\left(
 B_j\cap\left\{\max_i|\overline X_{i,j}-\mu_i|>\eta d_j\right\}
 \right)
 \le q_j\,\Pp_\mu(B_j).
\]
The event on the left means stage $j$ is the first
inaccurate reached stage. These events are disjoint and their
union is $\mathcal E^c$.

The initial stage $J$ is always reached, so $\Pp(B_J)=1$. For $j<J$, Lemma~\ref{screen:lemma} shows that
$B_j$ implies $\Gamma\le3d_j$. Hence, for every fixed $\mu$,
\[
 \Pp_\mu(B_j)\le\ind\{\Gamma(\mu)\le3d_j\},
 \qquad j<J.
\]
Summing over $j$ and integrating with
respect to $P_\nu(\dd\mu)$ therefore yields
\begin{align}
 \Pp_\nu(\mathcal E^c)
 &\le q_J+\sum_{j=0}^{J-1}q_j P_\nu(\Gamma\le3d_j)
 \notag\\
 &\le\frac{\eta(\alpha+\delta)}{Z}
 \left[
 4^{-J}+\sum_{j=0}^{J-1}4^{-j}F_\Gamma(3d_j)
 \right].
 \label{risk:first-bad-sum}
\end{align}
Recall
\[
Z=\min\left\{\frac43,\;4^{-J}+6D_\Gamma d_0\right\},
\qquad d_j=2^j d_0.
\]
First, using \(F_\Gamma\le1\),
\[
4^{-J}+\sum_{j=0}^{J-1}4^{-j}F_\Gamma(3d_j)
\le \sum_{j=0}^{J}4^{-j}
\le\frac43.
\]
Second, using \(F_\Gamma(x)\le D_\Gamma x\),
\[
\begin{aligned}
4^{-J}+\sum_{j=0}^{J-1}4^{-j}F_\Gamma(3d_j)
&\le4^{-J}+3D_\Gamma d_0\sum_{j=0}^{J-1}2^{-j}\\
&\le4^{-J}+6D_\Gamma d_0.
\end{aligned}
\]
It follows that $\left[
 4^{-J}+\sum_{j=0}^{J-1}4^{-j}F_\Gamma(3d_j)
 \right]\leq Z$. Substituting this bound in \eqref{risk:first-bad-sum} proves the claim.

\end{proof}

\subsection{Conditioning on the Complete Screen}

Let $\mathcal H_{\mathrm{scr}}$ be the sigma-field generated by all
screening observations that were collected, their stage indices,
the recorded candidate, and the allocation decision. Let $\widehat\jmath$ be the last screening stage reached and
$b=b_{\widehat\jmath}$ its empirical leader, as defined in
\eqref{screen:final-stage}.
Acceptance for confirmation is the event
\begin{equation}\label{risk:acceptance-event}
 \mathsf A:=
 \left\{\overline X_{b,\widehat\jmath}
       -\max_{i\ne b}\overline X_{i,\widehat\jmath}>d_{\widehat\jmath},\quad
       H(\ell)\le(1-8\eta)^{-2}h_a\right\}.
\end{equation}
Thus $\mathsf A\in\mathcal H_{\mathrm{scr}}$. And the values $\widehat\jmath,b,\ell,w,n,M_1,\ldots,M_K$ are determined by the screening
record. They are therefore
$\mathcal H_{\mathrm{scr}}$-measurable. Set $\overline Y_i=0$ on
$\mathsf A^c$; the following confirmation analysis is restricted to
$\mathsf A$.

The next two calculations condition on both $\mu$ and
$\mathcal H_{\mathrm{scr}}$, which together determine the means and sample
counts. On $\mathsf A$, the fresh confirmation batch has the conditional
product law
\begin{equation}\label{risk:fresh-law}
 \mathcal L\bigl((\overline Y_i)_{i=1}^K
                  \mid\mu,\mathcal H_{\mathrm{scr}}\bigr)
 =\bigotimes_{i=1}^K N(\mu_i,1/M_i).
\end{equation}

\begin{lemma}[Confirmation Succeeds on Accurate Accepted Screens]
\label{risk:confirmation}
The probability of confirmation failure on $\mathcal E$ is at most
\begin{equation}\label{risk:confirm-fail}
 \Pp_\nu(\mathcal E\cap\mathsf A\cap
              \{\text{confirmation fails}\})\le\eta {(\alpha+\delta)}.
\end{equation}
\end{lemma}
\begin{proof}
Condition on an accepted accurate
screen at stage $j$. Lemma~\ref{screen:lemma} gives $b=a^\star$
and $\Delta_i\ge\ell_i$. By definition and construction of $w$, for every $i\ne b$,
\begin{equation}
 \frac{\ell_i^2}{2(1/w_b+1/w_i)}
 \ge
 \min_{j\ne b}\frac{\ell_j^2}{2(1/w_b+1/w_j)}
 =I(\mu^\ell,w)
 =\frac1{H(\ell)}. \label{risk:signal_0}
\end{equation}
Rounding sample counts upward can only decrease the comparison
variance. Consequently by the definition of $n$ \eqref{alg:allocation}, 
\begin{equation} 
 \frac{\Delta_i}{\sqrt{1/M_b+1/M_i}}
 \ge\frac{\ell_i\sqrt n}{\sqrt{1/w_b+1/w_i}}
 \ge\sqrt{2n/H(\ell)}
 \ge\sqrt{2\beta}+\sqrt{2\log((K-1)/q_j)}. \label{risk:signal}
\end{equation}
By \eqref{risk:fresh-law}, conditional on $\mu$ and
$\mathcal H_{\mathrm{scr}}$, the confirmation means
$\overline Y_b$ and $\overline Y_i$ are independent normals.
Since $b=a^\star$ on an accurate accepted screen,
\[
 \mathcal L\!\left(
 \frac{\overline Y_b-\overline Y_i}{\sqrt{1/M_b+1/M_i}}
 \,\middle|\,\mu,\mathcal H_{\mathrm{scr}}
 \right)
 =
 N\!\left(
 \frac{\Delta_i}{\sqrt{1/M_b+1/M_i}},\,1
 \right).
\]
Its conditional mean is bounded below by \eqref{risk:signal}.
The conditional probability that confirmation fails (as per \eqref{alg:certificate}) is therefore at most
\[
 \Phi\!\left(\sqrt{2\beta}
     -\frac{\Delta_i}{\sqrt{1/M_b+1/M_i}}\right)
 \le\Phi(-\sqrt{2\log((K-1)/q_j)})\le \frac{q_j}{K-1}.
\]
The union of the $K-1$ failed comparisons has conditional probability
at most $(K-1)\frac{q_j}{K-1}=q_j$. 

Now average over $\mathcal H_{\mathrm{scr}}$. For a successful stage $j<J$ on
$\mathcal E$, \eqref{screen:reach} implies $\Gamma\le3d_j$.
On $\mathsf A$, $\widehat\jmath$ is the stage at which screening
succeeds and the hardness check accepts.
For each $j$, the event
$\mathcal E\cap\mathsf A\cap\{\widehat\jmath=j\}$
is determined by $\mu$ and the $\mathcal H_{\mathrm{scr}}$. On this event,
the conditional probability of confirmation failure is at most
$q_j$, as shown above. Therefore, the tower property gives
\begin{align*}
 \Pp_\nu(\mathcal E\cap\mathsf A
                  \cap\{\text{confirmation fails}\})
 &=\sum_{j=0}^{J}\E_\nu\!\left[
   \ind_{\mathcal E\cap\mathsf A\cap\{\widehat\jmath=j\}}
   \Pp_\nu\!\left(\text{confirmation fails}
       \mid\mu,\mathcal H_{\mathrm{scr}}\right)
   \right]\\
 &\le\sum_{j=0}^{J}q_j
       \Pp_\nu(\mathcal E\cap\mathsf A
                        \cap\{\widehat\jmath=j\})\\
 &\le q_J+\sum_{j=0}^{J-1}q_j F_\Gamma(3d_j)\\
 &\le\eta(\alpha+\delta).
\end{align*}
For the penultimate inequality, the probability for $j=J$ is
at most one. For $j<J$, reaching stage $j$ on $\mathcal E$
implies $\Gamma\le3d_j$ by Lemma~\ref{screen:lemma}, so
 $\Pp_\nu(\mathcal E\cap\mathsf A\cap\{\widehat\jmath=j\})
 \le P_\nu(\Gamma\le3d_j)=F_\Gamma(3d_j)$.
The last inequality follows from the same bounds on $q_j$ and
the geometric sum as \eqref{risk:first-bad-sum}.
\end{proof}

Let $\mathsf C\subseteq\mathsf A$ be the event that all
confirmation comparisons pass:
\[
 \mathsf C
 :=\left\{\omega\in\mathsf A:
 \overline Y_b-\overline Y_i
 >\sqrt{2\beta\left(\frac1{M_b}+\frac1{M_i}\right)}
 \ \text{for every }i\ne b
 \right\}.
\]
On $\mathsf C$, the algorithm recommends $b$ directly, without
using the terminal rejection coin.

\begin{lemma}[Wrong Certified Recommendations]\label{risk:wrong}
The probability of an incorrect recommendation through
certification satisfies
\[
 \Pp_\nu\bigl(\mathsf C\cap\{b\ne a^\star\}\bigr)
 \le\eta\delta.
\]
\end{lemma}
\begin{proof}

Conditional on $\mu$ and $\mathcal H_{\mathrm{scr}}$, the fresh
means are independent. On $\mathsf A\cap\{b\ne a^\star\}$,
\[
 \mathcal L\!\left(
 \overline Y_b-\overline Y_{a^\star}
 \,\middle|\,\mu,\mathcal H_{\mathrm{scr}}
 \right)
 =
 N\!\left(
 \mu_b-\mu_{a^\star},
 \frac1{M_b}+\frac1{M_{a^\star}}
 \right).
\]
Since $\mu_b<\mu_{a^\star}$, standardizing gives
\begin{align*}
 \Pp_\mu\!\left(
 \frac{\overline Y_b-\overline Y_{a^\star}}
      {\sqrt{1/M_b+1/M_{a^\star}}}
 >\sqrt{2\beta}
 \,\middle|\,\mathcal H_{\mathrm{scr}}
 \right)
 &=
 \Phi\!\left(
 \frac{\mu_b-\mu_{a^\star}}
      {\sqrt{1/M_b+1/M_{a^\star}}}
 -\sqrt{2\beta}
 \right)\\
 &\le \Phi(-\sqrt{2\beta}) \\
 &\le \exp\!\left(-\frac{(\sqrt{2\beta})^2}{2}\right)\\
 &=e^{-\beta}.
\end{align*}
The first inequality uses the negative mean and monotonicity
of $\Phi$; the second uses the Gaussian tail bound
$\Phi(-x)\le e^{-x^2/2}$ for $x\ge0$.

Recall that $\mathsf C\subseteq\mathsf A$ is the event of
successful certification. Conditional on $\mu$ and
$\mathcal H_{\mathrm{scr}}$, both $\mathsf A$ and $\{b\ne a^\star\}$
are determined. On $\mathsf A\cap\{b\ne a^\star\}$,
certification requires the comparison with $a^\star$ to pass,
so the preceding calculation gives
\[
 \Pp_\nu(\mathsf C\mid\mu,\mathcal H_{\mathrm{scr}})
 \le e^{-\beta}.
\]
Since $\mathsf C\subseteq\mathsf A$, the tower property yields
\begin{align*}
 \Pp_\nu(\mathsf C\cap\{b\ne a^\star\})
 &=\E_\nu\!\left[
   \E_\nu\!\left[
     \ind_{\mathsf C}\ind_{\mathsf A\cap\{b\ne a^\star\}}
     \,\middle|\,\mu,\mathcal H_{\mathrm{scr}}
   \right]
 \right]\\
 &=\E_\nu\!\left[
   \ind_{\mathsf A\cap\{b\ne a^\star\}}
   \Pp_\nu(\mathsf C\mid\mu,\mathcal H_{\mathrm{scr}})
 \right]\\
 &\le e^{-\beta}
       \Pp_\nu(\mathsf A\cap\{b\ne a^\star\}).
\end{align*}
By Lemma~\ref{screen:lemma}, an accurate successful screen
selects the true best arm. Therefore
\[
 \mathcal E\cap\mathsf A\subseteq\{b=a^\star\},
 \qquad
 \mathsf A\cap\{b\ne a^\star\}\subseteq\mathcal E^c.
\]
Combining this inclusion with \eqref{risk:bad} and
$\beta\ge\log((\alpha+\delta)/\delta)$ gives
\begin{align}
 \Pp_\nu(\mathsf C\cap\{b\ne a^\star\})
 &\le e^{-\beta}\Pp_\nu(\mathcal E^c)\notag\\
 &\le e^{-\beta}\eta(\alpha+\delta)\notag\\
 &\le\frac{\delta}{\alpha+\delta}\eta(\alpha+\delta)
 =\eta\delta.
 \label{risk:wrong-bound}
\end{align}
\end{proof}

\subsection{The Rejection Coin and Completion of the Risk Proof}

\begin{proof}\textbf{of feasibility of Theorem~\ref{thm:upper-rate}.} 
Let $\mathsf B$ be the event that the terminal rejection rule
is entered. This happens either when screening does not accept
an instance for confirmation, or when the confirmation batch
fails. Thus
\[
 \mathsf B
 =\mathsf A^c\cup(\mathsf A\cap\mathsf C^c).
\]
Here $\mathsf A^c$ includes both failure to pass any screening
stage and rejection by the hardness check.

On $\mathcal E$, Lemma~\ref{screen:lemma} shows that every
instance with $H<h_a$ is accepted for confirmation. Consequently,
\[
 \mathcal E\cap\mathsf A^c\subseteq\{H\ge h_a\}.
\]
Splitting the two rejection events according to whether
$\mathcal E$ holds therefore gives
\[
 \mathsf B
 \subseteq
 \{H\ge h_a\}
 \cup\mathcal E^c
 \cup(\mathcal E\cap\mathsf A\cap\mathsf C^c).
\]
Since $P_\nu(H>h_a)=a$ and $H$ has an atomless distribution,
$P_\nu(H\ge h_a)=a$. Applying a union bound, followed by
\eqref{risk:bad} and \eqref{risk:confirm-fail}, yields
\begin{align}
 \Pp_\nu(\mathsf B)
 &\le P_\nu(H\ge h_a)
      +\Pp_\nu(\mathcal E^c)
      +\Pp_\nu(\mathcal E\cap\mathsf A\cap\mathsf C^c)
      \notag\\
 &\le a+\eta(\alpha+\delta)+\eta(\alpha+\delta)
      \notag\\
 &=(1-2\eta)(\alpha+\delta),
 \label{risk:reject}
\end{align}
where the last equality uses $a=(1-4\eta)(\alpha+\delta)$.

Let $U$ be uniform on $(0,1)$ and independent of $\mu$ and all
sampling data. We may generate $U$ in advance and use it only
on $\mathsf B$. On that event, the terminal rule recommends $b$
if $U\le\delta/(\alpha+\delta)$ and abstains otherwise.

Thus an incorrect recommendation occurs either through
certification or through the terminal coin:
\begin{align*}
 \{D\in[K],D\ne a^\star\}
 =
 \bigl(\mathsf C\cap\{b\ne a^\star\}\bigr)
 \cup
 \left(
 \mathsf B\cap
 \left\{U\le\frac{\delta}{\alpha+\delta}\right\}
 \cap\{b\ne a^\star\}
 \right).
\end{align*}
These two events are disjoint. Dropping $\{b\ne a^\star\}$
from the second event and using independence of $U$ and
$\mathsf B$, we obtain
\begin{align}
 \Pp_\nu(D\in[K],D\ne a^\star)
 &\le
 \Pp_\nu(\mathsf C\cap\{b\ne a^\star\})
 +\frac{\delta}{\alpha+\delta}\Pp_\nu(\mathsf B)
 \notag\\
 &\le\eta\delta
 +\frac{\delta}{\alpha+\delta}(1-2\eta)(\alpha+\delta)
 \notag\\
 &=(1-\eta)\delta\le\delta.
 \label{risk:total-error}
\end{align}
Abstention occurs only through the terminal coin, so
\[
 \{D=?\}
 =\mathsf B\cap
   \left\{U>\frac{\delta}{\alpha+\delta}\right\}.
\]
Independence and \eqref{risk:reject} therefore give
\begin{align}
 \Pp_\nu(D=?)
 &=\frac{\alpha}{\alpha+\delta}\Pp_\nu(\mathsf B)
 \notag\\
 &\le\frac{\alpha}{\alpha+\delta}
       (1-2\eta)(\alpha+\delta)
 \notag\\
 &=(1-2\eta)\alpha\le\alpha.
 \label{risk:total-abstention}
\end{align}
\paragraph{A Deterministic Sampling Bound.}
For fixed admissible budgets, $0<a<1$, and Lemma~\ref{geom:regularity}
gives $0<h_a<\infty$. Thus $J$ and every $m_j$ are finite, and
$q_j\ge q_J>0$. On acceptance, the hardness cap and \eqref{alg:allocation}
bound $n$ by a deterministic constant. Since
$\sum_iM_i=\sum_i\lceil nw_i\rceil\le n+K$, the total sample size satisfies
\begin{equation}\label{risk:finite-budget}
 \tau_{\delta,\alpha}^{\SCalg}\le
 K\sum_{j=0}^Jm_j+
 \left\lceil\frac{h_a}{(1-8\eta)^2}
 \left(\sqrt\beta+\sqrt{\log\frac{K-1}{q_J}}\right)^2\right\rceil+K
 <\infty.
\end{equation}

\end{proof}


\section{Expected Sampling Cost for Arbitrary Budgets}
\label{cost:section}
\label{sec:cost}

Throughout this section, assume $\delta>0$, $\alpha\ge0$, and
$\alpha+\delta\le1$. Constants $c,C_1,\ldots,C_{12},C>0$ may depend on
$(K,\nu,\sigma_0)$ but not on $(\alpha,\delta)$. Each constant is fixed throughout this section. By the definitions of $\eta,a$, and $\beta$,
\[
 0<\eta\le\frac1{10},\qquad
 \frac35(\alpha+\delta)\le a<\alpha+\delta,\qquad
 1\le\beta\le1+\log\frac{\alpha+\delta}{\delta}.
\]
Recall that $\mathsf R_j$ is the event that screening reaches
stage $j$, and $\mathcal E$ is the event that every reached
screening stage is accurate.

\subsection{Bounds for the Screening Scales}

\begin{lemma}[Screening Thresholds and Error Allowances]
\label{cost:scale-lemma}
There exist constants $c,C_1,C_2>0$ such that, for every pair of budgets $(\alpha,\delta)$ satisfying $\delta>0$, $\alpha\ge0$, and
$\alpha+\delta\le1$, 
\begin{align}
 d_0&\ge c(\alpha+\delta), \qquad
 \frac{Z}{\alpha+\delta}\le C_1,
 \label{cost:scale}\\
 \log\frac{K-1}{q_j}
 \le \log\frac{2K}{q_j}
 &\le C_2\bigl(1+\log(1/\eta)+j\bigr)\quad
 \forall 0\le j\le J.
 \label{cost:logs}
\end{align}

\end{lemma}

\begin{proof}
The inequalities $8/\Gamma^2\le H\le4K/\Gamma^2$ imply
\[
 \left\{\Gamma<\sqrt{8/h_a}\right\}
 \subseteq\{H>h_a\}
 \subseteq\left\{\Gamma<\sqrt{4K/h_a}\right\}.
\]
Since $\Gamma$ has a density and $P_\nu(H>h_a)=a$, it follows that
\begin{equation*}
 F_\Gamma\!\left(\sqrt{8/h_a}\right)
 \le a
 \le F_\Gamma\!\left(\sqrt{4K/h_a}\right).
\end{equation*}
Using $F_\Gamma(x)\le D_\Gamma x$ and the definition of $d_0$,
\[
 a\le D_\Gamma\sqrt{4K/h_a}
   =D_\Gamma\sqrt{K/2}\,(1+2\eta)d_0.
\]
Using $a\ge3(\alpha+\delta)/5$ and $1+2\eta\le6/5$ give the
lower bound for $d_0$. The same inequality also gives
\begin{equation}
 h_a\le\frac{4KD_\Gamma^2}{a^2}
      \le\frac{C_3}{(\alpha+\delta)^2}.
 \label{cost:quantile-cap}
\end{equation}
Here $C_3:=100KD_\Gamma^2/9$.

To bound $Z$, first note that there is a constant $C_4>0$ such that
$F_\Gamma(x)\ge x/C_4$ for $0<x\le1$. Indeed,
$F_\Gamma(x)/x\to\kap>0$ at zero by \eqref{geom:gap-small}; on
every compact subinterval of $(0,1]$, $F_\Gamma(x)/x$ is continuous and
strictly positive. Since $\sqrt{8/h_a}=(1+2\eta)d_0$, we have
\[
 a=P_\nu(H>h_a) \ge P_\nu\!\left(\Gamma<\sqrt{8/h_a}\right)=F_\Gamma\!\left(\sqrt{8/h_a}\right)= F_\Gamma((1+2\eta)d_0)\ge F_\Gamma(d_0).
\]
If $d_0\le1$, this implies $d_0\le C_4(\alpha+\delta)$.
Also, $2^Jd_0\ge1$ implies $4^{-J}\le d_0^2$, so
\[
 Z\le4^{-J}+6D_\Gamma d_0
   \le d_0^2+6D_\Gamma d_0
   \le C_1(\alpha+\delta).
\]
If $d_0>1$, then $\alpha+\delta\ge a\ge F_\Gamma(1)>0$, and
\[
 Z\le\frac43
   \le\frac{4(\alpha+\delta)}{3F_\Gamma(1)}.
\]
This proves the asserted bound for $Z$ in both cases, with a fixed constant $C_1$.

Finally, by the definition of $q_j$,
\[
 \frac1{q_j}
 =\max\left\{4,\frac{Z4^j}{\eta(\alpha+\delta)}\right\}
 \le\frac{\max\{4,C_1\}4^j}{\eta}.
\]
Taking logarithms proves \eqref{cost:logs} for a suitable constant $C_2$.
\end{proof}

\subsection{Probabilities of Reaching the Screening Stages}

\begin{lemma}[Expected Sums over Reached Stages]
\label{cost:geometric-lemma}
There exists a constant $C_5>0$ such that, for every pair of budgets $(\alpha,\delta)$ satisfying $\delta>0$, $\alpha\ge0$, and
$\alpha+\delta\le1$,
\begin{align}
 \sum_{j=0}^Jd_j^{-2}\Pp_\nu(\mathsf R_j)
 &\le\frac{C_5}{\alpha+\delta},
 \label{cost:sum-unweighted}\\
 \sum_{j=0}^J\frac{1+\log(1/\eta)+j}{d_j^2}
                  \Pp_\nu(\mathsf R_j)
 &\le\frac{C_5(1+\log(1/\eta))}{\alpha+\delta},
 \label{cost:sum-weighted}\\
 J+1&\le C_5\left[1+\log\frac1{\alpha+\delta}\right]\le\frac{C_5}{\alpha+\delta}.
 \label{cost:rounding-stages}
\end{align}
\end{lemma}

\begin{proof}
For $j<J$, reaching stage $j$ on $\mathcal E$ implies
$\Gamma\le3d_j$ by \eqref{screen:reach}. Hence
\[
 \mathsf R_j\subseteq\{\Gamma\le3d_j\}\cup\mathcal E^c,
\]
and a union bound gives
\[
 \Pp_\nu(\mathsf R_j)
 \le F_\Gamma(3d_j)+\Pp_\nu(\mathcal E^c)
 \le3D_\Gamma d_j+\eta(\alpha+\delta).
\]
At the initial stage $J$, $\Pp_\nu(\mathsf R_J)=1$ and $d_J\ge1$.
With $C_6:=\max\{1,3D_\Gamma\}$, both cases are therefore covered by
\begin{equation}
 \Pp_\nu(\mathsf R_j)
 \le C_6d_j+\eta(\alpha+\delta),\qquad 0\le j\le J.
 \label{cost:reach}
\end{equation}

Since $d_j=2^jd_0$, the identities
\[
 \sum_{j\ge0}2^{-j}=2,\qquad
 \sum_{j\ge0}j2^{-j}=2,\qquad
 \sum_{j\ge0}4^{-j}=\frac43,\qquad
 \sum_{j\ge0}j4^{-j}=\frac49
\]
give
\begin{align*}
 \sum_{j=0}^Jd_j^{-1}&\le\frac2{d_0},&
 \sum_{j=0}^Jd_j^{-2}&\le\frac4{3d_0^2},\\
 \sum_{j=0}^J\frac{1+\log(1/\eta)+j}{d_j}
 &\le\frac{4(1+\log(1/\eta))}{d_0},&
 \sum_{j=0}^J\frac{1+\log(1/\eta)+j}{d_j^2}
 &\le\frac{16(1+\log(1/\eta))}{9d_0^2}.
\end{align*}
Multiplying \eqref{cost:reach} by the corresponding weights and
summing therefore yields
\begin{align*}
 \sum_{j=0}^Jd_j^{-2}\Pp_\nu(\mathsf R_j)
 &\le\frac{2C_6}{d_0}
       +\frac{4\eta(\alpha+\delta)}{3d_0^2},\\
 \sum_{j=0}^J\frac{1+\log(1/\eta)+j}{d_j^2}
                  \Pp_\nu(\mathsf R_j)
 &\le (1+\log(1/\eta))
       \left[\frac{4C_6}{d_0}
             +\frac{16\eta(\alpha+\delta)}{9d_0^2}\right].
\end{align*}
Substituting $d_0\ge c(\alpha+\delta)$ and $\eta\le1/10$ proves
\eqref{cost:sum-unweighted}--\eqref{cost:sum-weighted} for a sufficiently large $C_5$.

For \eqref{cost:rounding-stages}, if $d_0<1$ then
$J+1\le2+\log(1/d_0)/\log2$; otherwise $J+1=1$.
The same bound on $d_0$ gives
\[
 J+1\le C_5\left[1+\log\frac1{\alpha+\delta}\right]
       \le\frac{C_5}{\alpha+\delta},
\]
where the last inequality uses $1+\log(1/x)\le1/x$ for $0<x\le1$. Choose $C_5$ large enough that all three bounds hold.
\end{proof}

\subsection{Expected Number of Screening Observations}
\begin{lemma}[Expected time from screening]
\label{cost:screening_phase}
Let $N_{\mathrm{screen}}$ denote the number of individual pulls used
for screening. There exists a constant $C_7>0$ such that, for every pair of budgets $(\alpha,\delta)$ satisfying $\delta>0$, $\alpha\ge0$, and
$\alpha+\delta\le1$,
$$ \E_\nu[N_{\mathrm{screen}}] 
\le\frac{C_7\eta^{-2}(1+\log(1/\eta))}{\alpha+\delta}
       .$$
\end{lemma}

\begin{proof}
Stage $j$ contributes $Km_j$ pulls when
$\mathsf R_j$ occurs, so
\[
 N_{\mathrm{screen}}=K\sum_{j=0}^Jm_j\ind_{\mathsf R_j}.
\]
The block sizes $m_j$ are deterministic. Taking expectations and
using $\lceil x\rceil\le x+1$ gives
\begin{align}
 \E_\nu[N_{\mathrm{screen}}]
 &=K\sum_{j=0}^Jm_j\Pp_\nu(\mathsf R_j)\notag\\
 &\le\frac{2K}{\eta^2}\sum_{j=0}^J
       \frac{\log(2K/q_j)}{d_j^2}\Pp_\nu(\mathsf R_j)
       +K\sum_{j=0}^J\Pp_\nu(\mathsf R_j)\notag\\
 &\le 2KC_2\eta^{-2}\sum_{j=0}^J
       \frac{1+\log(1/\eta)+j}{d_j^2}\Pp_\nu(\mathsf R_j)
       +K(J+1)\notag\\
 &\le\frac{2KC_2C_5\eta^{-2}(1+\log(1/\eta))}{\alpha+\delta}
       +\frac{KC_5}{\alpha+\delta}.
 \label{cost:screening}
\end{align}
The last two inequalities use Lemmas~\ref{cost:scale-lemma}
and~\ref{cost:geometric-lemma}, respectively. Since $\eta^{-2}(1+\log(1/\eta))\ge1$, the second term can be absorbed into the first term, with $C_7:=KC_5(2C_2+1)$.
\end{proof}

\subsection{Expected Number of Confirmation Observations}

Recall that $\mathsf A$ is the event that screening succeeds and
the hardness check accepts, and $\widehat\jmath$ is the final
reached screening stage defined in \eqref{screen:final-stage}. Define
\[
 \widehat H:=
 \begin{cases}
 H(\ell),&\text{on }\mathsf A,\\
 0,&\text{on }\mathsf A^c.
 \end{cases}
\]
Thus $\widehat H$ is the characteristic time used to size the
confirmation batch, and is zero when no batch is collected.

\begin{lemma}[Moments of the Confirmation Characteristic Time]
\label{cost:hardness-lemma}
At every successful screening stage $j$,
\begin{equation}
 H(\ell)\le\frac{4K}{(1-2\eta)^2d_j^2}
           \le\frac{C_8}{d_j^2}.
 \label{cost:stage}
\end{equation}
where $C_8:=25K/4$. Consequently there exists a constant $C_9>0$ such that, for every pair of budgets $(\alpha,\delta)$ satisfying $\delta>0$, $\alpha\ge0$, and
$\alpha+\delta\le1$,
\begin{align}
 \E_\nu[\widehat H]&\le\frac{C_9}{\alpha+\delta},
 \label{cost:Hmoment}\\
 \E_\nu\!\left[\widehat H\log\frac{K-1}{q_{\widehat\jmath}}\right]
 &\le\frac{C_9(1+\log(1/\eta))}{\alpha+\delta},
 \label{cost:Hcmoment}\\
 \E_\nu\!\left[\widehat H
             \sqrt{\log\frac{K-1}{q_{\widehat\jmath}}}\right]
 &\le\frac{C_9\sqrt{1+\log(1/\eta)}}{\alpha+\delta}.
 \label{cost:mixed-moment}
\end{align}
\end{lemma}

\begin{proof}
Successful screening implies $\ell_i>(1-2\eta)d_j$ for every
$i\ne b$, whether or not the screen is accurate. Evaluating the
information at the uniform allocation $w_i=1/K$ gives
\[
 \max_{w\in\Delta_K}\min_{i\ne b}
       \frac{w_bw_i}{w_b+w_i}\frac{\ell_i^2}{2}
 \ge\min_{i\ne b}\frac{\ell_i^2}{4K}
 \ge\frac{(1-2\eta)^2d_j^2}{4K}.
\]
Taking reciprocals proves \eqref{cost:stage}.

The events $\mathsf A\cap\{\widehat\jmath=j\}$ partition
$\mathsf A$ and are contained in $\mathsf R_j$. Therefore
\begin{align*}
 \E_\nu[\widehat H]
 &=\sum_{j=0}^J
   \E_\nu[\widehat H\ind_{\mathsf A\cap\{\widehat\jmath=j\}}]
 \le \sum_{j=0}^JC_8 d_j^{-2}\Pp_\nu(\mathsf A\cap\{\widehat\jmath=j\})
 \le C_8\sum_{j=0}^Jd_j^{-2}\Pp_\nu(\mathsf R_j),
\end{align*}
where the first inequality follows from \eqref{cost:stage}.
Furthermore
\begin{align*}
     \E_\nu\!\left[\widehat H\log\frac{K-1}{q_{\widehat\jmath}}\right]
 =\sum_{j=0}^J\log\frac{K-1}{q_j}
   \E_\nu[\widehat H\ind_{\mathsf A\cap\{\widehat\jmath=j\}}]
 \le C_2C_8\sum_{j=0}^J
       \frac{1+\log(1/\eta)+j}{d_j^2}\Pp_\nu(\mathsf R_j).
\end{align*}
Here the last inequality uses \eqref{cost:logs} and \eqref{cost:stage}.
Using Lemma~\ref{cost:geometric-lemma} proves \eqref{cost:Hmoment} and \eqref{cost:Hcmoment} with $C_9:=C_8C_5\max\{1,C_2\}$. Applying Cauchy--Schwarz to $\sqrt{\widehat H}$ and
$\sqrt{\widehat H\log((K-1)/q_{\widehat\jmath})}$ gives
\begin{align*}
 \E_\nu\!\left[\widehat H
             \sqrt{\log\frac{K-1}{q_{\widehat\jmath}}}\right]
 \le\sqrt{\E_\nu[\widehat H]\,
       \E_\nu\!\left[\widehat H
       \log\frac{K-1}{q_{\widehat\jmath}}\right]}
 \le\frac{C_9\sqrt{1+\log(1/\eta)}}{\alpha+\delta}.
\end{align*}
\end{proof}

\begin{proposition}[Expected Cost in Terms of the Confirmation Characteristic Time]
\label{cost:master-proposition}
There exists a constant $C_{10}>0$ such that, for every pair of budgets $(\alpha,\delta)$ satisfying $\delta>0$, $\alpha\ge0$, and
$\alpha+\delta\le1$,
\begin{equation}
 \begin{aligned}
 \E_\nu[\tau]\le{}&\beta\E_\nu[\widehat H]+(K+1)\\
 &{}+\frac{C_{10}}{\alpha+\delta}\left[
     \eta^{-2}(1+\log(1/\eta))
     +\sqrt{\beta(1+\log(1/\eta))}
     +1+\log(1/\eta)\right].
 \end{aligned}
 \label{cost:master}
\end{equation}
\end{proposition}

\begin{proof}
Let $N_{\mathrm{confirm}}$ count the confirmation pulls, with value
zero on $\mathsf A^c$. On $\mathsf A$, \eqref{alg:allocation}
and $\sum_iw_i=1$ imply
\[
 N_{\mathrm{confirm}}=\sum_{i=1}^K\lceil nw_i\rceil
 \le n+K
 \le\widehat H
       \left(\sqrt\beta+
             \sqrt{\log\frac{K-1}{q_{\widehat\jmath}}}\right)^2
       +K+1.
\]
Since $\widehat H=0$ on $\mathsf A^c$, expanding the square, we get the following bound on the whole sample space:
\begin{align*}
 N_{\mathrm{confirm}}
 \le{}\beta\widehat H
       +2\sqrt\beta\,\widehat H
     \sqrt{\log\frac{K-1}{q_{\widehat\jmath}}}
 +\widehat H\log\frac{K-1}{q_{\widehat\jmath}}
       +(K+1)\ind_{\mathsf A}.
\end{align*}
Taking expectations and applying
Lemma~\ref{cost:hardness-lemma} yields
\begin{align}
 \E_\nu[N_{\mathrm{confirm}}]
 &\le{}\beta\E_\nu[\widehat H]\\
       &\quad+\frac{2C_9}{\alpha+\delta}
         \left[\sqrt{\beta(1+\log(1/\eta))}
         +1+\log(1/\eta)\right]
 +(K+1)\Pp_\nu(\mathsf A).
 \label{cost:confirmation}
\end{align}

Since
$\tau=N_{\mathrm{screen}}+N_{\mathrm{confirm}}$. Add
\eqref{cost:screening} and \eqref{cost:confirmation} gives the claimed bound with $C_{10}:=\max\{C_7,2C_9\}$.
\end{proof}

\subsection{The Uniform Upper Bound}

\begin{proposition}[Upper Bound for Arbitrary Budgets]
\label{cost:uniform-proposition}
There exists a finite constant $C=C(K,\nu,\sigma_0)$, independent of
$(\alpha,\delta)$, such that for every $\delta>0$, $\alpha\ge0$ with $\alpha+\delta\le1$,
\begin{equation}
 \E_\nu[\tau]\le
 C\frac{\kap^2}{\alpha+\delta}
 \left[1+\log\frac{\alpha+\delta}{\delta}\right].
 \label{cost:uniform}
\end{equation}
\end{proposition}

\begin{proof}
The prescribed value of $\eta$ gives
\begin{equation}
 1+\log(1/\eta)
 =1+\log10+\frac18\log\left[1+\log\frac{\alpha+\delta}{\delta}\right].
 \label{cost:B}
\end{equation}
Also,
\begin{equation}
 \eta^{-2}=100\left[1+\log\frac{\alpha+\delta}{\delta}\right]^{1/4}.
 \label{asym:explicit-slack}
\end{equation}
Together with $\beta\le1+\log((\alpha+\delta)/\delta)$, these
identities imply, for a fixed constant $C_{11}>0$,
\begin{align*}
 \eta^{-2}(1+\log(1/\eta))
 &\le C_{11}\left[1+\log\frac{\alpha+\delta}{\delta}\right],\\
 \sqrt{\beta(1+\log(1/\eta))}+1+\log(1/\eta)
 &\le C_{11}\left[1+\log\frac{\alpha+\delta}{\delta}\right].
\end{align*}
These inequalities follow from boundedness of
$x^{-3/4}(1+\log x)$ and $x^{-1}(1+\log x)$ on $x\ge1$,
applied at $x=1+\log((\alpha+\delta)/\delta)$.
Also, \eqref{cost:Hmoment} gives
\[
 \beta\E_\nu[\widehat H]
 \le\frac{C_9}{\alpha+\delta}
        \left[1+\log\frac{\alpha+\delta}{\delta}\right].
\]
Substituting these three bounds into \eqref{cost:master} proves
\[
 \E_\nu[\tau]\le\frac{C_{12}}{\alpha+\delta}
           \left[1+\log\frac{\alpha+\delta}{\delta}\right],
\]
where $C_{12}:=C_9+2C_{10}C_{11}+K+1$; the term $K+1$ can be absorbed because the expression multiplying $C_{12}$
is at least one. Since $\kap>0$ is fixed, setting
$C=C_{12}/\kap^2$ proves \eqref{cost:uniform}. Note this constant is independent of $(\alpha,\delta)$.
\end{proof}

\section{Asymptotic Upper Bounds on the Expected Cost}
\label{asym:section}
\label{sec:asymptotics}

We first bound $\E_\nu[\widehat H]$ by truncated $H$. We then use Proposition~\ref{cost:master-proposition}
to show that the remaining terms in the expected-cost bound are
of smaller order when $\alpha/\delta\to\infty$.

\begin{lemma}[Upper Bound on the Expected Confirmation Characteristic Time]
\label{asym:sandwich}
For every $\delta>0$ and $\alpha\ge0$ satisfying
$\alpha+\delta\le1$,
\begin{equation}
 \E_\nu[\widehat H]
 \le\frac{\E_\nu\!\left[H\ind\{H\le h_a/(1-8\eta)^2\}\right]}
              {(1-8\eta)^2}
       +\frac{h_a\eta(\alpha+\delta)}{(1-8\eta)^2}.
 \label{asym:hardness-sandwich}
\end{equation}
\end{lemma}

\begin{proof}
On $\mathcal E\cap\mathsf A$, Lemma~\ref{screen:lemma} gives
$H\le H(\ell)\le H/(1-8\eta)^2$. The acceptance check also
gives $H(\ell)\le h_a/(1-8\eta)^2$ on $\mathsf A$,
regardless of screening accuracy. Hence
\[
 \mathcal E\cap\mathsf A
 \subseteq\{H\le h_a/(1-8\eta)^2\}.
\]
Since $\widehat H=H(\ell)$ on $\mathsf A$ and zero otherwise,
\begin{align*}
 \widehat H\ind_{\mathcal E}
 &\le\frac{H}{(1-8\eta)^2}
      \ind\{H\le h_a/(1-8\eta)^2\},\\
 \widehat H\ind_{\mathcal E^c}
 &\le\frac{h_a}{(1-8\eta)^2}\ind_{\mathcal E^c}.
\end{align*}
Add these inequalities, take expectations, and use
$\Pp_\nu(\mathcal E^c)\le\eta(\alpha+\delta)$ gives the claimed bound.
\end{proof}

\begin{lemma}[Asymptotic Reduction to the Confirmation Cost]
\label{asym:cost-lemma}
Along every sequence of budgets satisfying $\delta>0$, $\alpha\ge0$,
$\alpha+\delta\le1$, and $\alpha/\delta\to\infty$,
\begin{equation}
 \E_\nu[\tau]
 \le\beta\E_\nu[\widehat H]
   +o\!\left(\frac{1+\log((\alpha+\delta)/\delta)}
                    {\alpha+\delta}\right).
 \label{asym:cost-reduction}
\end{equation}
Consequently,
\begin{equation}
 \E_\nu[\tau]
 \le\frac{\beta}{(1-8\eta)^2}
 \E_\nu\!\left[
 H\ind\!\left\{H\le\frac{h_a}{(1-8\eta)^2}\right\}
 \right]
 +o\!\left(\frac{\beta}{\alpha+\delta}\right).
 \label{asym:truncated-cost}
\end{equation}
\end{lemma}

\begin{proof}
For $\alpha/\delta$ sufficiently large,
$\beta=\log((\alpha+\delta)/\delta)$; and $\beta\to\infty$ as $\alpha/\delta\to \infty$.
The definition of $\eta$ therefore gives
\[
 \eta^{-2}=10^2(1+\beta)^{1/4},\qquad
 1+\log(1/\eta)=1+\log10+\frac18\log(1+\beta).
\]
Substitution gives
\begin{align}
 \frac{\eta^{-2}(1+\log(1/\eta))}
      {1+\log((\alpha+\delta)/\delta)}&\longrightarrow0,\notag\\
 \frac{\sqrt{\beta(1+\log(1/\eta))}+1+\log(1/\eta)}
      {1+\log((\alpha+\delta)/\delta)}&\longrightarrow0.
 \label{cost:negligible}
\end{align}
The rounding term also satisfies
\[
 0\le\frac{(K+1)(\alpha+\delta)}
              {1+\log((\alpha+\delta)/\delta)}
 \le\frac{K+1}{1+\log((\alpha+\delta)/\delta)}
 \longrightarrow0.
\]
Applying these limits to \eqref{cost:master} proves
\eqref{asym:cost-reduction}.

For the second conclusion, note that as $\alpha/\delta\to\infty$,
$\log((\alpha+\delta)/\delta)\to\infty$, hence 
$\beta/[1+\log((\alpha+\delta)/\delta)]\to1$.
By \eqref{cost:quantile-cap},
$h_a\le C_3/(\alpha+\delta)^2$. Also, $\eta\le1/10$ implies
$(1-8\eta)^{-2}\le25$. Consequently,
\[
 \frac{\beta h_a\eta(\alpha+\delta)}{(1-8\eta)^2}
 \le 25C_3\,
      \frac{\eta\beta}{\alpha+\delta}.
\]
Since $\alpha/\delta\to\infty$, the definition of $\eta$ gives
$\eta\to0$. Dividing the last bound by
$\beta/(\alpha+\delta)$ gives $25C_3\eta\to0$, proving
\[
 \frac{\beta h_a\eta(\alpha+\delta)}{(1-8\eta)^2}
 =o\!\left(\frac{\beta}{\alpha+\delta}\right).
\]
Combining \eqref{asym:cost-reduction} with
Lemma~\ref{asym:sandwich} proves \eqref{asym:truncated-cost}.
\end{proof}

\subsection{Fixed Positive Abstention}

\begin{theorem}[Upper Bound at Fixed Positive Abstention]
\label{asym:fixed-alpha}
For every fixed $\alpha\in(0,1)$,
\begin{equation}
 \limsup_{\delta\downarrow0}
       \frac{\E_\nu[\tau_{\delta,\alpha}^{\SCalg}]}{\log(1/\delta)}
       \le V_\nu(\alpha).
 \label{asym:fixed-alpha-limit}
\end{equation}
\end{theorem}

\begin{proof}
As $\delta\downarrow0$ with $\alpha$ fixed,
\[
 \alpha+\delta\longrightarrow\alpha,\qquad
 \eta\longrightarrow0,\qquad
 a=(1-4\eta)(\alpha+\delta)\longrightarrow\alpha.
\]
Continuity of $a\mapsto h_a$ gives
\[
 h_a\longrightarrow h_\alpha,
 \qquad \frac{h_a}{(1-8\eta)^2}\longrightarrow h_\alpha.
\]
Since $0<h_\alpha<\infty$ and
$h_a/(1-8\eta)^2\to h_\alpha$, for all sufficiently small
$\delta$ we have
 $\frac{h_a}{(1-8\eta)^2}\le2h_\alpha$.
Consequently,
\[
 0\le H\ind\{H\le h_a/(1-8\eta)^2\}\le2h_\alpha.
\]
The constant $2h_\alpha$ is integrable and independent of $\delta$. Since $P_\nu(H=h_\alpha)=0$,
dominated convergence yields
\[
 \E_\nu\!\left[H\ind\{H\le h_a/(1-8\eta)^2\}\right]
 \longrightarrow V_\nu(\alpha).
\]
Also, $h_a\eta(\alpha+\delta)\to0$ and $(1-8\eta)^{-2}\to1$.
Thus \eqref{asym:hardness-sandwich} gives
\[
 \limsup_{\delta\downarrow0}\E_\nu[\widehat H]
 \le V_\nu(\alpha).
\]
For total cost $\tau$, observe that as $\delta\downarrow 0$,
\[
 \frac{\beta}{\log(1/\delta)}\longrightarrow1,
 \qquad
 \frac{1+\log((\alpha+\delta)/\delta)}
      {(\alpha+\delta)\log(1/\delta)}
 \longrightarrow\frac1\alpha.
\]
Divide \eqref{asym:cost-reduction} by $\log(1/\delta)$ to obtain
\[
 \frac{\E_\nu[\tau]}{\log(1/\delta)}
 \le\frac{\beta}{\log(1/\delta)}\E_\nu[\widehat H]+o(1).
\]
Taking limsup proves \eqref{asym:fixed-alpha-limit}.
\end{proof}

\subsection{Shrinking Abstention with \texorpdfstring{$\alpha/\delta\to\infty$}{Alpha/Delta Tending to Infinity}}

\begin{theorem}[Upper Bound When $\alpha\downarrow0$ and $\alpha/\delta\to\infty$]
\label{asym:separated}
The algorithm satisfies
\begin{equation}
 \limsup_{\substack{\alpha,\delta\downarrow0\\\alpha/\delta\to\infty}}
 \frac{\alpha\E_\nu[\tau_{\delta,\alpha}^{\SCalg}]}
      {\log(\alpha/\delta)}\le 8\kap^2.
 \label{asym:separated-limit}
\end{equation}
\end{theorem}

\begin{proof}
Along every sequence with $\alpha\downarrow0$ and
$\alpha/\delta\to\infty$, we have $ \eta\rightarrow0$, and
\begin{align}
 \frac{\alpha+\delta}{\alpha}&\longrightarrow1,&
 \frac a\alpha=\frac{(1-4\eta)(\alpha+\delta)}{\alpha}&\longrightarrow1,\notag\\
 \frac{1+\log((\alpha+\delta)/\delta)}{\log(\alpha/\delta)}
 &\longrightarrow1,&
 \frac{\beta}{\log(\alpha/\delta)}&\longrightarrow1.
 \label{asym:separated-scales}
\end{align}
Recall \eqref{geom:quantile-asymptotic} gives
$\lim_{a\downarrow0}a^2h_a=8\kap^2$
and \eqref{geom:truncated-moment} gives
\begin{equation}
 \lim_{t\to\infty}
 \frac{\E_\nu[H\ind\{H\le t\}]}{\sqrt{8\kap^2t}}=1.
 \label{asym:truncated-mean}
\end{equation}
In particular, as $\alpha \downarrow 0$ and
$\alpha/\delta\to\infty$, $h_a\to\infty$ and
$h_a/(1-8\eta)^2\to\infty$.

Multiply \eqref{asym:hardness-sandwich} by $\alpha/(8\kap^2)$. For the first term, write
\begin{align*}
 \frac{\alpha\E_\nu[H\ind\{H\le h_a/(1-8\eta)^2\}]}
       {8\kap^2(1-8\eta)^2}
 =
 \frac{\E_\nu[H\ind\{H\le h_a/(1-8\eta)^2\}]}
      {\sqrt{8\kap^2h_a/(1-8\eta)^2}}
 \cdot\frac\alpha a
 \cdot\sqrt{\frac{a^2h_a}{8\kap^2}}
 \cdot\frac1{(1-8\eta)^3}.
\end{align*}
Applying \eqref{asym:truncated-mean} with
$t=h_a/(1-8\eta)^2$ shows that this term tends to one, since every factor on the
right tends to one. The second term vanishes because
\[
 \alpha h_a\eta(\alpha+\delta)
 =(a^2h_a)\frac{\alpha(\alpha+\delta)}{a^2}\eta
 \longrightarrow0,
 \qquad (1-8\eta)^{-2}\longrightarrow1.
\]
Since the sequence was arbitrary,
\[
 \limsup_{\substack{\alpha,\delta\downarrow0\\\alpha/\delta\to\infty}}
 \frac{\alpha\E_\nu[\widehat H]}{8\kap^2}\le1.
\]

Now use \eqref{asym:cost-reduction}, 
the remainder in \eqref{asym:cost-reduction} is negligible, since
\[
 \frac{\alpha}{\alpha+\delta}
 \frac{1+\log((\alpha+\delta)/\delta)}{\log(\alpha/\delta)}
 \longrightarrow1.
\]
Consequently,
\[
 \frac{\alpha\E_\nu[\tau]}{8\kap^2\log(\alpha/\delta)}
 \le\frac{\beta}{\log(\alpha/\delta)}
       \frac{\alpha\E_\nu[\widehat H]}{8\kap^2}+o(1).
\]
Using
$\beta/\log(\alpha/\delta)\to1$, taking limsup proves the result.
\end{proof}

\begin{corollary}[Upper Bound on Power Paths with $0<p<1$]
\label{asym:power}
For every fixed $p\in(0,1)$,
\begin{equation}
 \limsup_{\delta\downarrow0}
 \frac{\E_\nu[\tau_{\delta,\delta^p}^{\SCalg}]}
      {8(1-p)\kap^2\delta^{-p}\log(1/\delta)}\le1.
 \label{asym:power-limit}
\end{equation}
\end{corollary}

\begin{proof}
With $\alpha=\delta^p$, we have
$\alpha\downarrow0$, $\alpha/\delta=\delta^{p-1}\to\infty$, and
$\log(\alpha/\delta)=(1-p)\log(1/\delta)$.
Substitution into Theorem~\ref{asym:separated} proves the result.
\end{proof}

\begin{corollary}[Upper Bound with Iterated limit, $\delta\downarrow0$ before $\alpha\downarrow0$]
\label{asym:iterated}
The algorithm satisfies
\begin{equation}
 \limsup_{\alpha\downarrow0}\alpha
   \left[\limsup_{\delta\downarrow0}
      \frac{\E_\nu[\tau_{\delta,\alpha}^{\SCalg}]}{\log(1/\delta)}\right]
       \le8\kap^2.
 \label{asym:iterated-limit}
\end{equation}
\end{corollary}

\begin{proof}
For each fixed $\alpha\in(0,1)$,
Theorem~\ref{asym:fixed-alpha} gives
\[
 \alpha\left[\limsup_{\delta\downarrow0}
       \frac{\E_\nu[\tau_{\delta,\alpha}^{\SCalg}]}{\log(1/\delta)}\right]
 \le\alpha V_\nu(\alpha).
\]
Taking the upper limit as $\alpha\downarrow0$ and using
$\lim_{\alpha\downarrow0}\alpha V_\nu(\alpha)=8\kap^2$
from \eqref{geom:V-asymptotic} proves the claim.
\end{proof}

\section{Proofs of the Lower Bounds}
\label{sec:lower-proofs}

All lower bounds apply to the adaptive randomized policies in
\eqref{eq:optimal-value}. With constrained Bayesian error and abstention probabilities.

\subsection{The Lower Bound at Fixed Positive Abstention}
\label{lower:fixed-section}

\begin{theorem}[Fixed Positive Abstention: Lower Bound]
\label{lower:fixed}
For every fixed $\alpha\in(0,1)$,
\begin{equation}\label{lower:fixed-bound}
 \liminf_{\delta\downarrow0}
 \frac{\Tstar(\delta,\alpha)}{\log(1/\delta)}
 \ge V_\nu(\alpha).
\end{equation}
\end{theorem}

Fix an instance $\mu$ and let $i=a^\star(\mu)$.
Choose a measurable set $B$ of alternative instances at which
arm $i$ is not best, with $P_\nu(B)>0$.
We compare running the policy at $\mu$ with running the policy at
$\lambda$ drawn from $P_\nu(\cdot\mid B)$.
In the second experiment, recommending $i$ is always an error.
The Bayesian error constraint therefore bounds its probability by
\[
 \int_B \Pp_\lambda(D=i)\,
       \frac{P_\nu(\dd\lambda)}{P_\nu(B)}
 \le \frac{\delta}{P_\nu(B)}.
\]
We use this comparison to obtain a lower bound
on the expected sample size at $\mu$.

We first prove the comparison inequality used in the theorem.  Let
$N_a(\tau)$ denote the total number of observations from arm $a$.

\begin{lemma}[Comparison with an Alternative Prior Mixture]
\label{lower:mixture-lemma}
Fix a $(\delta,\alpha)$-feasible policy with recommendation $D$, an arm $i$, and a measurable set
\[
 B\subset\{\lambda\in\R^K:
                 \max_{j\ne i}\lambda_j>\lambda_i\}
 \quad\text{with}\quad P_\nu(B)>0.
\]
At every instance $\mu$ at which the policy terminates almost surely,
\begin{equation}\label{lower:mixture-comparison}
 \Pp_\mu(D=i)\log\frac{P_\nu(B)}{\delta}-\log2
 \le
 \sum_{a=1}^K\E_\mu[N_a(\tau)]
       \int_B\frac{(\mu_a-\lambda_a)^2}{2}
                     \frac{P_\nu(\dd\lambda)}{P_\nu(B)}.
\end{equation}
\end{lemma}

\begin{proof}
Let $U$ collect the policy's internal randomization, independently of
$\mu$ and the reward sequences $(Y_{i,s})_{i,s}$. Let $A_t$ and $X_t$
denote the arm selected and reward observed at round $t$, and define
\[
 \mathcal F_0:=\sigma(U),\qquad
 \mathcal F_t:=\sigma(U,A_1,X_1,\ldots,A_t,X_t),\quad t\ge1.
\]
The sampling rule makes $A_t$ measurable with respect to
$\mathcal F_{t-1}$, $\tau$ is an $(\mathcal F_t)$-stopping time, and
$D$ is $\mathcal F_\tau$-measurable.

It suffices to consider $\E_\mu[\tau]<\infty$; otherwise, the right
side of \eqref{lower:mixture-comparison} is infinite. Put
$$q:=\Pp_\mu(D=i)$$, and write
\[
 \operatorname{kl}(q,p)
 :=q\log\frac{q}{p}+(1-q)\log\frac{1-q}{1-p}
\]
for the binary relative entropy, with the usual limiting values at
$p,q\in\{0,1\}$.

Feasibility implies that the policy terminates almost surely at
$P_\nu$-almost every $\lambda$, hence also at
$P_\nu(\cdot\mid B)$-almost every $\lambda$.
For each such $\lambda$, apply Lemma~1 of \citet{KCG16} to the two
instances $\mu,\lambda$, the stopping time $\tau$, and the event
$\{D=i\}\in\mathcal F_\tau$. Since
$\operatorname{KL}(\mathcal N(\mu_a,1)\|\mathcal N(\lambda_a,1))
=(\mu_a-\lambda_a)^2/2$, this gives
\begin{equation}\label{lower:pairwise-comparison}
 \sum_{a=1}^K\E_\mu[N_a(\tau)]
                 \frac{(\mu_a-\lambda_a)^2}{2}
 \ge \operatorname{kl}\!\left(q,\Pp_\lambda(D=i)\right).
\end{equation}
The independent random seed $U$ is included in $\mathcal F_t$;
its law is the same under both instances and contributes zero
relative entropy.

Let $Q_B$ be the probability law obtained by first drawing
$\lambda$ from $P_\nu(\cdot\mid B)$ and then running the policy at
$\lambda$. Integrating \eqref{lower:pairwise-comparison} over this
conditional prior and using convexity of $\operatorname{kl}(q,\cdot)$
yields
\begin{align}
 \sum_{a=1}^K\E_\mu[N_a(\tau)]
       \int_B\frac{(\mu_a-\lambda_a)^2}{2}
        \frac{P_\nu(\dd\lambda)}{P_\nu(B)}\notag
 &\ge \int_B
       \operatorname{kl}\!\left(q,\Pp_\lambda(D=i)\right)
        \frac{P_\nu(\dd\lambda)}{P_\nu(B)}\\
 &\ge \operatorname{kl}\!\left(q,Q_B(D=i)\right).
 \label{lower:averaged-comparison}
\end{align}
Because recommending $i$ is an error at every $\lambda\in B$,
\begin{equation}\label{lower:mixture-error}
 Q_B(D=i)
 =\frac{1}{P_\nu(B)}\int_B\Pp_\lambda(D=i)\,P_\nu(\dd\lambda)
 \le\frac{\delta}{P_\nu(B)}.
\end{equation}
Finally, binary relative entropy satisfies
\begin{align*}
 \operatorname{kl}(q,p)
 &=q\log\frac1p+(1-q)\log\frac1{1-p}
   +q\log q+(1-q)\log(1-q)\\
 &\ge q\log\frac1p-\log2,
\end{align*}
since the second term is nonnegative and
$q\log q+(1-q)\log(1-q)\ge-\log2$.
Consequently, \eqref{lower:mixture-error} gives
\[
 \operatorname{kl}\!\left(q,Q_B(D=i)\right)
 \ge q\log\frac{P_\nu(B)}{\delta}-\log2.
\]
Combining this with \eqref{lower:averaged-comparison} proves
\eqref{lower:mixture-comparison}.
\end{proof}

\begingroup
\renewcommand{\proofname}{Proof of Theorem~\ref{lower:fixed}}
\begin{proof}
\textbf{Step 1: The Prior Mass of Correct Recommendations.}\quad
Fix $\alpha\in(0,1)$ and $\delta\in(0,1)$. Consider an arbitrary
$(\delta,\alpha)$-feasible policy with finite
Bayesian expected sample size.  Policies with infinite expected sample
size satisfy every lower bound below automatically.  For prior-almost
every $\mu$, the best arm is unique, the policy terminates, and
$\E_\mu[\tau]<\infty$.  At these instances define
\[
 t(\mu):=\E_\mu[\tau],
 \qquad q(\mu):=\Pp_\mu(D=a^\star(\mu)).
\]
The total probability of a correct recommendation is at least
$1-\alpha-\delta$, so
\begin{equation}\label{lower:correct-mass}
 0\le q\le1,
 \qquad \E_\nu[q]\ge1-\alpha-\delta.
\end{equation}

\paragraph{Step 2: Construct Alternatives with Small Information Cost.}
Fix $R>0$ and $\varepsilon\in(0,1]$, independently of the budgets.
Take an instance with $\|\mu\|_\infty\le R$, and put
$i=a^\star(\mu)$.  If $t(\mu)>0$, let
\[
 w_a:=\frac{\E_\mu[N_a(\tau)]}{t(\mu)},\qquad a\in[K].
\]
Since $\sum_aN_a(\tau)=\tau$, these numbers form a vector in
$\Delta_K$.  If $t(\mu)=0$, choose any $w\in\Delta_K$; all the
expected counts are then zero.

Next, we adapt the Gaussian construction of \citet[Lemma~3]{GK16}
to obtain a set of alternatives with positive prior mass and
nearly the same information cost. Choose a challenger $j\ne i$ attaining
the minimum in $I(\mu,w)$. Form a vector $\lambda^0$ by keeping every coordinate of $\mu$
except $i$ and $j$, and setting these two coordinates equal to
\[
 \lambda_i^0=\lambda_j^0
 =\frac{w_i\mu_i+w_j\mu_j}{w_i+w_j}.
\]
If $w_i+w_j=0$, choose any common value between $\mu_j$ and $\mu_i$.
A direct expansion of the two changed coordinates gives
\begin{equation}\label{lower:closest-tie}
 \sum_{a=1}^K w_a\frac{(\mu_a-\lambda_a^0)^2}{2}
 =\frac{w_iw_j}{w_i+w_j}\frac{(\mu_i-\mu_j)^2}{2}
 =I(\mu,w)\le\frac{1}{H(\mu)}.
\end{equation}
The fraction is defined to be zero when $w_i+w_j=0$.

Let $e_i,e_j$ denote the corresponding coordinate vectors, and let
$B$ be the open cube of side length $\varepsilon$ centered at
$\lambda^0+\varepsilon(e_j-e_i)$.  Every $\lambda\in B$ satisfies
$\lambda_j>\lambda_i$, so $B$ is an admissible alternative set for
Lemma~\ref{lower:mixture-lemma}.  Moreover,
\[
 B\subset[-R-2,R+2]^K,
 \qquad
 |\lambda_a-\lambda_a^0|\le\frac{3\varepsilon}{2}
 \quad(a\in[K]).
\]
The prior $P_\nu$ on the mean vector has density
$\lambda\mapsto\prod_{a=1}^K f_a(\lambda_a)$, which is continuous
and strictly positive. Therefore, define
\[
 m_R:=
 \min_{\lambda\in[-R-2,R+2]^K}
       \prod_{a=1}^K f_a(\lambda_a)>0.
\]
Since $B\subset[-R-2,R+2]^K$ and $B$ has volume
$\varepsilon^K$, we can have
\begin{equation}\label{lower:cube-mass}
 P_\nu(B)\ge m_R\varepsilon^K.
\end{equation}
Note this bound is uniform in $\mu$ and $w$, and $m_R\varepsilon^K\le1$.

Both $\mu_a$ and $\lambda_a^0$ lie in $[-R,R]$.  For every $\lambda\in B$ and $a\in[K]$, we have
$|\mu_a-\lambda_a^0|\le2R$ and
$|\lambda_a-\lambda_a^0|\le3\varepsilon/2$. Hence
\begin{align*}
 \frac{(\mu_a-\lambda_a)^2}{2}
 &=\frac{(\mu_a-\lambda_a^0)^2}{2}
   +(\mu_a-\lambda_a^0)(\lambda_a^0-\lambda_a)
   +\frac{(\lambda_a^0-\lambda_a)^2}{2}\\
 &\le\frac{(\mu_a-\lambda_a^0)^2}{2}
   +(2R)\frac{3\varepsilon}{2}
   +\frac12\left(\frac{3\varepsilon}{2}\right)^2.
\end{align*}
Multiplying by $w_a\ge0$, summing over $a$, and using
$\sum_a w_a=1$ and \eqref{lower:closest-tie} gives
\begin{align}
 \sum_{a=1}^K w_a\frac{(\mu_a-\lambda_a)^2}{2}
 &\le
 \sum_{a=1}^K w_a\frac{(\mu_a-\lambda_a^0)^2}{2}
       +3R\varepsilon+\frac{9\varepsilon^2}{8}\notag\\
 &\le \frac{1}{H(\mu)}
       +3R\varepsilon+\frac{9\varepsilon^2}{8}.\label{lower:cube-information}
\end{align}
Since $\E_\mu[N_a(\tau)]=t(\mu)w_a$,
Lemma~\ref{lower:mixture-lemma} and \eqref{lower:cube-information} give
\begin{align*}
 q(\mu)\log\frac{P_\nu(B)}{\delta}-\log2
 &\le t(\mu)\int_B
       \left[\sum_{a=1}^K w_a\frac{(\mu_a-\lambda_a)^2}{2}\right]
       \frac{P_\nu(\dd\lambda)}{P_\nu(B)}\\
 &\le t(\mu)\left(\frac{1}{H(\mu)}
              +3R\varepsilon+\frac{9\varepsilon^2}{8}\right).
\end{align*}
To further lower bound the left side, \eqref{lower:cube-mass} and $0\le q\le1$
imply
\[
 q(\mu)\log P_\nu(B)
 \ge q(\mu)\log(m_R\varepsilon^K)
 \ge \log(m_R\varepsilon^K),
\]
where the last inequality uses $\log(m_R\varepsilon^K)\le0$.
Expanding yields
\begin{equation}\label{lower:uniform-instance}
 t(\mu)\left(\frac{1}{H(\mu)}
              +3R\varepsilon+\frac{9\varepsilon^2}{8}\right)
 \ge q(\mu)\log\frac{1}{\delta}
       -\log\frac{2}{m_R\varepsilon^K}.
\end{equation}

\paragraph{Step 3: Integrate Using the Fixed Cutoff.}
To integrate \eqref{lower:uniform-instance}, multiply it by
$\min\{H(\mu),h_\alpha\}$ on $\{\|\mu\|_\infty\le R\}$, note
 $\min\{H,h_\alpha\}
 \left(H^{-1}+3R\varepsilon+\frac{9\varepsilon^2}{8}\right)
 \le1+h_\alpha\left(3R\varepsilon+
\frac{9\varepsilon^2}{8}\right)$.
Hence, on $\|\mu\|_\infty\le R$,
\begin{align*}
 \left[1+h_\alpha\left(3R\varepsilon+
   \frac{9\varepsilon^2}{8}\right)\right]t(\mu)
 \ge
 q(\mu)\min\{H(\mu),h_\alpha\}\log\frac1\delta
 -\min\{H(\mu),h_\alpha\}\log\frac{2}{m_R\varepsilon^K}.
\end{align*}
Integrating the preceding pointwise inequality on
$\{\|\mu\|_\infty\le R\}$ gives
\begin{align*}
 &\left[1+h_\alpha\left(3R\varepsilon+
  \frac{9\varepsilon^2}{8}\right)\right]
       \E_\nu\!\left[t(\mu)\ind\{\|\mu\|_\infty\le R\}\right]\\
 &\qquad\ge
 \log(1/\delta)\,
 \E_\nu\!\left[q\min\{H,h_\alpha\}
                 \ind\{\|\mu\|_\infty\le R\}\right]
-
 \log\frac{2}{m_R\varepsilon^K}\,
 \E_\nu\!\left[\min\{H,h_\alpha\}
    \ind\{\|\mu\|_\infty\le R\}\right].
\end{align*}
Since $t(\mu)\ge0$, the expectation on the left satisfies
\[
 \E_\nu\!\left[t(\mu)\ind\{\|\mu\|_\infty\le R\}\right]
 \le \E_\nu[t(\mu)]=\E_\nu[\tau].
\]
For the negative term
on the right, note
 $\E_\nu\!\left[\min\{H,h_\alpha\}
\ind\{\|\mu\|_\infty\le R\}\right]
 \le h_\alpha$.
Since $\log(2/(m_R\varepsilon^K))\ge0$, 
$$
-\log\frac{2}{m_R\varepsilon^K}\,
 \E_\nu\!\left[\min\{H,h_\alpha\}
    \ind\{\|\mu\|_\infty\le R\}\right]
\geq
-h_\alpha\log(2/(m_R\varepsilon^K))
$$
For the first term on the right, recall that, by Lemma~\ref{geom:regularity}, $h_\alpha$ is finite and
$P_\nu(H>h_\alpha)=\alpha$. Consequently,
 $\E_\nu[\min\{H,h_\alpha\}]
 =\E_\nu[H\ind\{H\le h_\alpha\}]
   +h_\alpha P_\nu(H>h_\alpha)
 =V_\nu(\alpha)+\alpha h_\alpha$.
Combining this with the $(\delta,\alpha)$-feasibility constraint
\eqref{lower:correct-mass} gives
\begin{align*}
 \E_\nu\!\left[q\min\{H,h_\alpha\}
                 \ind\{\|\mu\|_\infty\le R\}\right]
 &\ge \E_\nu[\min\{H,h_\alpha\}]
       -h_\alpha\E_\nu[1-q]
       -h_\alpha P_\nu(\|\mu\|_\infty>R)\\
 &\ge V_\nu(\alpha)
       -h_\alpha\left(\delta+P_\nu(\|\mu\|_\infty>R)\right).
\end{align*}
Substituting these bounds and dividing by $\log(1/\delta)>0$
gives
\begin{align}
 &\left[1+h_\alpha\left(3R\varepsilon+
     \frac{9\varepsilon^2}{8}\right)\right]
       \frac{\E_\nu[\tau]}{\log(1/\delta)}
\ge
 V_\nu(\alpha)
       -h_\alpha\left(\delta+P_\nu(\|\mu\|_\infty>R)\right)
 -\frac{h_\alpha\log(2/(m_R\varepsilon^K))}{\log(1/\delta)}.
 \label{lower:capped-cost}
\end{align}
The right-hand side and the coefficient multiplying
$\E_\nu[\tau]$ are independent of the policy. We may therefore take the infimum over feasible
policies. Keeping $R$ and $\varepsilon$ fixed and letting
$\delta\downarrow0$ then gives
\[
 \liminf_{\delta\downarrow0}
 \frac{\Tstar(\delta,\alpha)}{\log(1/\delta)}
 \ge
 \frac{V_\nu(\alpha)-h_\alpha P_\nu(\|\mu\|_\infty>R)}
      {1+h_\alpha(3R\varepsilon+9\varepsilon^2/8)}.
\]
Finally, let $\varepsilon\downarrow0$ for fixed $R$, and then let
$R\uparrow\infty$. Since $P_\nu(\|\mu\|_\infty>R)\to0$, the
right side tends to $V_\nu(\alpha)$, proving
\eqref{lower:fixed-bound}.
\end{proof}
\endgroup

\subsection{A Submeasure: Flat Local Subprior}

To prove lower bounds for the other two regimes, we adapt the flat local subprior construction of
\citet[Appendix~C.1]{HHT26}. It is used both in the lower bound for
arbitrary budget ratios and in the sharp lower bound as abstention
vanishes. 
For a fixed slack parameter $\rho\in(0,1)$, we construct a measure
$Q_\rho$ dominated by the true Gaussian prior $P_\nu$.

\begin{definition}[Local Change of Variables]
For a fixed pair of arms $i<j$, we reparameterize the mean vector
using the midpoint $m:=(\mu_i+\mu_j)/2$, the signed gap
$d:=\mu_i-\mu_j$, and the remaining means
$z:=(\mu_k)_{k\notin\{i,j\}}$.
We consolidate the background variables into
$\xi:=(i,j,m,z)$. Given $\xi$ and $d$, the inverse map
$\mu(\xi,d)\in\R^K$ is
\begin{equation}\label{lb:flat-coordinate-map}
 \mu_i=m+d/2,\qquad \mu_j=m-d/2,\qquad
 \mu_k=z_k\quad(k\notin\{i,j\}).
\end{equation}
The absolute Jacobian determinant of
$(m,d)\mapsto(\mu_i,\mu_j)$ is $1$.
\end{definition}

\begin{remark}
To see why the restricted true prior is difficult to work with, let
\[
 \Omega_{ij}:=
 \{\mu:\min\{\mu_i,\mu_j\}>\mu_k
                    \text{ for every }k\notin\{i,j\}\}.
\]
These regions partition the parameter space up to prior-null ties.
For any nonnegative measurable function $\psi$, the change of
variables in \eqref{lb:flat-coordinate-map} gives
\begin{align*}
 \int_{\R^K}\psi(\mu)\,P_\nu(\dd\mu)
 &=\sum_{i<j}\int_{\Omega_{ij}}
      \psi(\mu)\prod_{a=1}^K f_a(\mu_a)\,\dd\mu\\
 &=\sum_{i<j}\int_\R\int_{\mathcal D_{ij}(d)}
      \psi\bigl(\mu(\xi,d)\bigr)
      f_i(m+d/2)f_j(m-d/2)
      \prod_{k\notin\{i,j\}}f_k(z_k)\,\dd m\,\dd z\,\dd d,
\end{align*}
where
\[
 \mathcal D_{ij}(d):=
 \{(m,z):z_k<m-|d|/2
                  \text{ for every }k\notin\{i,j\}\}.
\]
The domain $\mathcal D_{ij}(d)$ shrinks as $|d|$ grows, and the
densities $f_i$ and $f_j$ vary with both the magnitude and sign of
$d$. We eliminate these dependencies by restricting $(m,z)$ to a
smaller set and lower bounding $f_i(m+d/2)f_j(m-d/2)$ by
$(1-\rho)f_i(m)f_j(m)$, which does not depend on $d$.
\end{remark}

For $\ell>0$, $M>0$, and each pair $i<j$, define the
\emph{$(\ell,M)$-restricted (compact) set}
\[
 \mathcal C_{ij}(\ell,M):=
 \{(m,z):|m|\le M,\ |z_k|\le M,\ z_k\le m-\ell
                    \text{ for every }k\notin\{i,j\}\}.
\]
Recall that
\begin{equation}\label{lb:tie-face-integral}
 \kappa_\nu
 =2\sum_{i<j}\int_\R
    \int_{\{z_k<m:\ k\notin\{i,j\}\}}
      f_i(m)f_j(m)\prod_{k\notin\{i,j\}}f_k(z_k)\,\dd z\,\dd m.
\end{equation}
By monotone convergence, we can choose $\ell>0$ sufficiently small
and $M>0$ sufficiently large that
\begin{equation}\label{lb:restricted-tie-mass}
 2\sum_{i<j}\int_{\mathcal C_{ij}(\ell,M)}
      f_i(m)f_j(m)\prod_{k\notin\{i,j\}}f_k(z_k)\,\dd m\,\dd z
 \ge(1-\rho)\kappa_\nu.
\end{equation}
Let
\[
 \mathcal C_\rho:=
 \bigsqcup_{i<j}\{(i,j)\}\times\mathcal C_{ij}(\ell,M)
\]
be the disjoint union, with the arm pair included in $\xi$.
Define a finite measure $\Lambda_\rho$ on $\mathcal C_\rho$ by
specifying its restriction to each pair:
\begin{equation}\label{lb:background-measure}
 \Lambda_\rho(\dd\xi):=
 (1-\rho)f_i(m)f_j(m)
       \prod_{k\notin\{i,j\}}f_k(z_k)\,\dd m\,\dd z.
\end{equation}
Equations \eqref{lb:tie-face-integral} and
\eqref{lb:restricted-tie-mass} imply
\begin{equation}\label{lb:flat-kappa}
 (1-\rho)^2\kappa_\nu
 \le\kappa_\rho:=2\Lambda_\rho(\mathcal C_\rho)
 \le(1-\rho)\kappa_\nu.
\end{equation}
In particular, $\kappa_\rho\to\kappa_\nu$ as $\rho\downarrow0$.

\begin{lemma}[Uniform Density Lower Bound]\label{lb:uniform-density}
There exists $\varepsilon_\rho\in(0,2\ell)$ such that, for every
pair $i<j$, every $(m,z)\in\mathcal C_{ij}(\ell,M)$, and every
$|d|\le\varepsilon_\rho$,
\[
 (1-\rho)f_i(m)f_j(m)
 \le f_i(m+d/2)f_j(m-d/2).
\]
Throughout this region, arms $i$ and $j$ are the top two arms.
\end{lemma}

\begin{proof}
For each pair, the Gaussian densities are continuous and strictly
positive. Therefore the ratio
\[
 (m,d)\longmapsto
 \frac{f_i(m+d/2)f_j(m-d/2)}{f_i(m)f_j(m)}
\]
is continuous on the compact set $[-M,M]\times[-1,1]$ and equals
one when $d=0$. Uniform continuity gives a sufficiently small
positive cutoff for which the ratio is at least $1-\rho$
uniformly over $|m|\le M$. Since there are finitely many pairs,
choose $\varepsilon_\rho$ below all their cutoffs and below
$2\ell$. Moreover, for every remaining arm,
\[
 \min\{\mu_i,\mu_j\}=m-|d|/2
 \ge m-\varepsilon_\rho/2>m-\ell\ge z_k.
\]
Thus the stated pair consists of the top two arms.
\end{proof}

\begin{definition}[Flat Local Subprior]\label{lb:flat-definition}
Define the \emph{flat local subprior} $Q_\rho$ by
\begin{equation}\label{lb:flat-definition-equation}
 Q_\rho(E):=
 \sum_{i<j}\int_{\mathcal C_{ij}(\ell,M)}
       \int_{-\varepsilon_\rho}^{\varepsilon_\rho}
       \ind_E\bigl(\mu(\xi,d)\bigr)\,\dd d\,\Lambda_\rho(\dd\xi),
 \qquad E\in\mathcal B(\R^K).
\end{equation}
Here the integral for each pair uses the corresponding restriction
of $\Lambda_\rho$.
\end{definition}

\begin{lemma}[Domination]\label{lb:subprior-domination}
The constructed measure is dominated by the true prior:
$Q_\rho(E)\le P_\nu(E)$ for every Borel set $E\subseteq\R^K$.
\end{lemma}

\begin{proof}
For each $i<j$, let
\[
 \mathcal S_{ij}:=
 \{\mu(\xi,d):(m,z)\in\mathcal C_{ij}(\ell,M),\
                         |d|\le\varepsilon_\rho\},
 \qquad
 \mathcal S:=\bigcup_{i<j}\mathcal S_{ij}.
\]
Substituting \eqref{lb:background-measure} into
\eqref{lb:flat-definition-equation} gives
\begin{align*}
 Q_\rho(E)
 &=\sum_{i<j}\int_{\mathcal C_{ij}(\ell,M)}
       \int_{-\varepsilon_\rho}^{\varepsilon_\rho}
       \ind_E\bigl(\mu(\xi,d)\bigr)(1-\rho)f_i(m)f_j(m)
       \prod_{k\notin\{i,j\}}f_k(z_k)\,\dd d\,\dd m\,\dd z.
\end{align*}
For each pair, the coordinate map has absolute Jacobian determinant
one. The top-two pair is uniquely determined on $\mathcal S$,
so the sets $\mathcal S_{ij}$ are disjoint. Consequently, $Q_\rho$
has Lebesgue density
\[
 q_\rho\bigl(\mu(\xi,d)\bigr)
 =(1-\rho)f_i(m)f_j(m)
       \prod_{k\notin\{i,j\}}f_k(z_k)
\]
almost everywhere on $\mathcal S_{ij}$, and density zero outside
$\mathcal S$.

At the same instance, the true prior density is
\[
 p_\nu\bigl(\mu(\xi,d)\bigr)
 =f_i(m+d/2)f_j(m-d/2)
       \prod_{k\notin\{i,j\}}f_k(z_k).
\]
Lemma~\ref{lb:uniform-density} gives $q_\rho\le p_\nu$ almost
everywhere on $\mathcal S$. The inequality also holds outside
$\mathcal S$, where $q_\rho=0$. Hence, for every Borel set $E$,
\[
 Q_\rho(E)=\int_E q_\rho(\mu)\,\dd\mu
 \le\int_E p_\nu(\mu)\,\dd\mu=P_\nu(E).
\]
In particular, $Q_\rho(\R^K)\le1$.
\end{proof}

\begin{lemma}[A Subprior with Constant Gap Density]
\label{lb:flat-subprior}
For every $\rho\in(0,1)$, the construction above gives a measure
$Q_\rho\le P_\nu$ whose local-coordinate representation is
\begin{equation}\label{lb:flat-product}
 Q_\rho(\dd\xi,\dd d)=\Lambda_\rho(\dd\xi)\,\dd d,
 \qquad \xi\in\mathcal C_\rho,\quad |d|<\varepsilon_\rho,
\end{equation}
with $\kappa_\rho=2\Lambda_\rho(\mathcal C_\rho)$ satisfying
\eqref{lb:flat-kappa}. The pair specified by $\xi$ consists of the
top two arms throughout the support of $Q_\rho$.
The following properties hold:
\begin{enumerate}
 \item \textbf{Fair Sign.} After normalizing $Q_\rho$ to a
 probability measure, the sign of $d$ is equally likely to be
 positive or negative, conditionally on the background and $|d|$.
 \item \textbf{Flat Gap.} The unnormalized density of the top-two
 gap $\Gamma=|d|$ is constant:
 \[
  Q_\rho(\Gamma\in\dd g)=\kappa_\rho\,\dd g,
  \qquad 0<g<\varepsilon_\rho.
 \]
\end{enumerate}
\end{lemma}

\begin{proof}
The representation \eqref{lb:flat-product} is the definition of
$Q_\rho$ in local coordinates. In terms of the absolute gap and
sign, it gives
\[
 Q_\rho(\dd\xi,\ |d|\in\dd g,\ \operatorname{sgn}(d)=s)
 =\Lambda_\rho(\dd\xi)\,\dd g,
 \qquad 0<g<\varepsilon_\rho,\quad s\in\{-1,+1\}.
\]
The two signs therefore have equal mass for each background and
absolute gap. Since $Q_\rho(\R^K)=\kappa_\rho\varepsilon_\rho>0$,
normalization yields conditional sign probabilities $1/2$.
Integrating out the background and summing over the two signs gives
$2\Lambda_\rho(\mathcal C_\rho)\,\dd g=\kappa_\rho\,\dd g$.
\end{proof}

\begin{lemma}[Risk and Cost Domination Under $Q_\rho$]
\label{lb:risk-domination}
For every policy with terminal decision $D$ and sample size $\tau$
satisfying $\Pp_\nu(\tau<\infty)=1$,
\begin{align*}
 \int\Pp_\mu(D=?)\,Q_\rho(\dd\mu)
 &\le\Pp_\nu(D=?),\\
 \int\Pp_\mu(D\in[K],D\ne a^\star)\,Q_\rho(\dd\mu)
 &\le\Pp_\nu(D\in[K],D\ne a^\star),\\
 \int\E_\mu[\tau]\,Q_\rho(\dd\mu)
 &\le\E_\nu[\tau].
\end{align*}
In particular, for a $(\delta,\alpha)$-feasible policy, the
abstention and error masses under $Q_\rho$ are at most $\alpha$
and $\delta$, respectively.
\end{lemma}

\begin{proof}
Each integrand is nonnegative, so the inequalities follow from
Lemma~\ref{lb:subprior-domination}.
\end{proof}
Thus, to lower bound the original Bayesian error, it suffices to
lower bound the error mass under $Q_\rho$, subject to the same
abstention budget. These comparisons apply to both fixed-budget
policies and policies with a random stopping time.

\subsection{The Lower Bound for Arbitrary Budget Ratios}

The following argument compares two prior regions having opposite best
arms. Their masses are proportional to the total risk budget
$\alpha+\delta$. The Bayesian constraints then force a constant
probability of the correct decision on one region and a much smaller
probability of the same decision on the other.

\begin{proposition}[A Lower Bound for All Sufficiently Small Budgets]
\label{lower:all-path}
For the fixed Gaussian prior, there is $\varepsilon>0$ such that every
Bayesian $(\delta,\alpha)$-feasible policy satisfies
\begin{equation}\label{lower:all-path-bound}
 \E_\nu[\tau]
 \ge \frac{\log 2}{64}
       \frac{\kappa_\nu^2}{\alpha+\delta}
       \left[1+\log\frac{\alpha+\delta}{\delta}\right]
\end{equation}
whenever $\delta>0$, $\alpha\ge0$, and $\alpha+\delta<\varepsilon$.
Consequently, the same lower bound holds for
$\Tstar(\delta,\alpha)$. The numerical constant does not depend on
$K,\nu$, or $\sigma_0$; the threshold $\varepsilon$ may depend on them.
\end{proposition}

\begin{proof}
If $\E_\nu[\tau]=\infty$, the asserted lower bound is immediate.
Hence assume $\E_\nu[\tau]<\infty$.
\paragraph{Step 1: Define Two Probability Laws with Opposite Best Arms.}
Apply Lemma~\ref{lb:flat-subprior} with $\rho=1/4$, so that
$\kappa_\rho\ge\kappa_\nu/2$, and set
\[
 r:=\frac{8(\alpha+\delta)}{\kappa_\rho}.
\]
Choose $\varepsilon=\kappa_\rho\varepsilon_\rho/8$. The assumption $\alpha+\delta<\varepsilon$ then gives $r<\varepsilon_\rho$. For
$\xi=(i,j,m,z)\in\mathcal C_\rho$ and $0<d<r$, define the instances
\[
 \mu^+(\xi,d):=\mu(\xi,d),\qquad
 \mu^-(\xi,d):=\mu(\xi,-d),
\]
where $\mu(\xi,d)$ defines an instance given by \eqref{lb:flat-coordinate-map}.
Their best arms are $i$ and $j$, respectively. Define the two sets of instances
\[
 \mathcal S_\pm(r):=
 \{\mu^\pm(\xi,d):\xi\in\mathcal C_\rho,\ 0<d<r\}.
\]
The product representation \eqref{lb:flat-product} gives
\begin{equation}\label{lower:paired-region-mass}
 Q_\rho(\mathcal S_\pm(r))
 =\int_{\mathcal C_\rho}\int_0^r\dd d\,\Lambda_\rho(\dd\xi)
 =\frac{r\kappa_\rho}{2}=4(\alpha+\delta).
\end{equation}

Let $U$ collect the policy's internal randomization, independently
of the instance and all arm reward sequences. Let $A_t$ and $X_t$
denote the arm selected and reward observed at round $t$.
Write $\Pp_\mu(\dd\omega)$ for the law at instance $\mu$ of the reward history
$\omega:=\bigl(U,(A_t,X_t)_{1\le t\le\tau},\tau,D\bigr)$.
Define the joint probability laws of $(\xi,d,\omega)$ by
\begin{equation}\label{lower:paired-joint-laws}
 \mathbb Q_\pm(\dd\xi,\dd d,\dd\omega)
 :=\frac{\Lambda_\rho(\dd\xi)\,\dd d}{4(\alpha+\delta)}
       \Pp_{\mu^\pm(\xi,d)}(\dd\omega),
 \qquad \xi\in\mathcal C_\rho,\quad 0<d<r.
\end{equation}
The factor $\Lambda_\rho(\dd\xi)\,\dd d/[4(\alpha+\delta)]$
is a probability measure on $\mathcal C_\rho\times(0,r)$, since
\[
 \int_{\mathcal C_\rho}\int_0^r
       \frac{\dd d\,\Lambda_\rho(\dd\xi)}{4(\alpha+\delta)}
 =\frac{r\Lambda_\rho(\mathcal C_\rho)}{4(\alpha+\delta)}
 =1
\]
by \eqref{lower:paired-region-mass}. Integrating the conditional law
$\Pp_{\mu^\pm(\xi,d)}(\dd\omega)$ over the policy record $\omega$
also gives one. Thus \eqref{lower:paired-joint-laws} defines
probability laws on $(\xi,d,\omega)$. Conditional on $(\xi,d)$,
the policy is run at $\mu^+(\xi,d)$ under $\mathbb Q_+$ and at
$\mu^-(\xi,d)$ under $\mathbb Q_-$. For every Borel set
$B\subseteq\R^K$, the resulting instance distributions satisfy
\[
 \mathbb Q_\pm\bigl(\mu^\pm(\xi,d)\in B\bigr)
 =\frac{Q_\rho(B\cap\mathcal S_\pm(r))}
        {Q_\rho(\mathcal S_\pm(r))}.
\]
Thus these instance distributions are precisely the normalized
restrictions of $Q_\rho$ to $\mathcal S_+(r)$ and $\mathcal S_-(r)$.
\paragraph{Step 2: Translate the Bayesian Risks into a Binary Test.}
For $\xi=(i,j,m,z)$, define $i(\xi):=i$ and $j(\xi):=j$, and set
\[
 A:=\{(\xi,d,\omega):D(\omega)=i(\xi)\}.
\]
Since $D\in[K]\cup\{?\}$, the complement of $A$ is the disjoint union
\[
 A^c=\{D=?\}\mathbin{\cup}
       \{D\in[K],\ D\ne i(\xi)\}.
\]
We have $a^\star(\mu^+(\xi,d))=i(\xi)$. Thus
\eqref{lower:paired-joint-laws} gives
\begin{align}
 \mathbb Q_+(A^c)
 &=\frac{1}{4(\alpha+\delta)}
   \int_{\mathcal S_+(r)}
       \bigl[\Pp_\mu(D=?)+
       \Pp_\mu(D\in[K],D\ne a^\star(\mu))\bigr]Q_\rho(\dd\mu)
       \notag\\
 &\le\frac{
   \Pp_\nu(D=?)+\Pp_\nu(D\in[K],D\ne a^\star)}
   {4(\alpha+\delta)}
       \notag\\
 &\le\frac{\alpha+\delta}{4(\alpha+\delta)}=\frac14.
 \label{lower:paired-correct}
\end{align}
Where the second inequality is by Lemma~\ref{lb:risk-domination}, and the last inequality is the policy's $(\delta,\alpha)$-feasibility.

At $\mu^-(\xi,d)$, the best arm is $j(\xi)\ne i(\xi)$, so
$\{D=i(\xi)\}\subseteq\{D\in[K],D\ne a^\star(\mu^-(\xi,d))\}$.
Consequently,
\begin{align}
 \mathbb Q_-(A)
 &=\frac{1}{4(\alpha+\delta)}
   \int_{\mathcal C_\rho}\int_0^r
       \Pp_{\mu^-(\xi,d)}(D=i(\xi))\,\dd d\,\Lambda_\rho(\dd\xi)
       \notag\\
 &\le\frac{1}{4(\alpha+\delta)}
   \int_{\mathcal S_-(r)}
       \Pp_\mu(D\in[K],D\ne a^\star(\mu))\,Q_\rho(\dd\mu)
       \notag\\
 &\le\frac{\Pp_\nu(D\in[K],D\ne a^\star)}{4(\alpha+\delta)}
       \qquad\text{(Lemma~\ref{lb:risk-domination})}\notag\\
 &\le\frac{\delta}{4(\alpha+\delta)}.
 \label{lower:paired-error}
\end{align}
For either sign, the expected sample size satisfies
\begin{align}
 \E_{\mathbb Q_\pm}[\tau]
 &=\frac{1}{4(\alpha+\delta)}
   \int_{\mathcal S_\pm(r)}\E_\mu[\tau]\,Q_\rho(\dd\mu)
       \label{lower:decompose_exp_q}\\
 &\le\frac{\E_\nu[\tau]}{4(\alpha+\delta)}<\infty
       \qquad\text{(Lemma~\ref{lb:risk-domination})}.
 \label{lower:paired-cost}
\end{align}
Here we used $\E_\nu[\tau]<\infty$.
In particular,
 $\mathbb Q_+(\tau<\infty)=\mathbb Q_-(\tau<\infty)=1$.
\paragraph{Step 3: Bound the Information in the Observations.}
Equations \eqref{lower:paired-joint-laws} and \eqref{lower:decompose_exp_q}
give, for either sign,
\[
 \int_{\mathcal C_\rho}\int_0^r
   \E_{\mu^\pm(\xi,d)}[\tau]\,\dd d\,\Lambda_\rho(\dd\xi)
 =4(\alpha+\delta)\E_{\mathbb Q_\pm}[\tau]<\infty.
\]
Thus $\E_{\mu^+(\xi,d)}[\tau]<\infty$ and
$\E_{\mu^-(\xi,d)}[\tau]<\infty$ for
$\Lambda_\rho(\dd\xi)\,\dd d$-almost every $(\xi,d)$ in
$\mathcal C_\rho\times(0,r)$.
Define the probabilities of recommending $i(\xi)$ at the paired
instances by
\[
 p_\pm(\xi,d):=
   \Pp_{\mu^\pm(\xi,d)}(D=i(\xi)).
\]
By \eqref{lower:paired-joint-laws},
\[
 \mathbb Q_\pm(A)
 =\frac{1}{4(\alpha+\delta)}
    \int_{\mathcal C_\rho}\int_0^r
        p_\pm(\xi,d)\,\dd d\,\Lambda_\rho(\dd\xi).
\]
For every arm $a\in[K]$, the Gaussian relative entropy is
\begin{align*}
 \operatorname{KL}\!\left(
   \mathcal N(\mu_a^+(\xi,d),1)
   \,\middle\|\,
   \mathcal N(\mu_a^-(\xi,d),1)\right)
 &=
   \frac{(\mu_a^+(\xi,d)-\mu_a^-(\xi,d))^2}{2}\\
   &=\frac{d^2}{2}\ind\{a\in\{i(\xi),j(\xi)\}\}.
\end{align*}
Apply the change-of-measure inequality
\eqref{lower:pairwise-comparison} to the instances
$\mu^+(\xi,d),\mu^-(\xi,d)$ and the event $\{D=i(\xi)\}$.
For almost every $(\xi,d)$ as above, this gives
\begin{equation}\label{lower:paired-instance-kl}
 \operatorname{kl}\bigl(p_+(\xi,d),p_-(\xi,d)\bigr)
 \le\frac{d^2}{2}
     \E_{\mu^+(\xi,d)}\!\left[
       N_{i(\xi)}(\tau)+N_{j(\xi)}(\tau)\right].
\end{equation}
Joint convexity of $\operatorname{kl}$ and
\eqref{lower:paired-instance-kl} now yield
\begin{align}
 \operatorname{kl}\bigl(\mathbb Q_+(A),\mathbb Q_-(A)\bigr)
 &\le\frac{1}{4(\alpha+\delta)}
    \int_{\mathcal C_\rho}\int_0^r
      \operatorname{kl}\bigl(p_+(\xi,d),p_-(\xi,d)\bigr)
             \,\dd d\,\Lambda_\rho(\dd\xi)\notag\\
 &\le\E_{\mathbb Q_+}\!\left[
       \frac{d^2}{2}
       \{N_{i(\xi)}(\tau)+N_{j(\xi)}(\tau)\}\right]\notag\\
 &\le\frac{r^2}{2}\E_{\mathbb Q_+}[\tau] \\
 &\le\frac{r^2}{8(\alpha+\delta)}\E_\nu[\tau].
 \label{lower:paired-kl}
\end{align}
The third inequality uses $0<d<r$ and
\[
 N_{i(\xi)}(\tau)+N_{j(\xi)}(\tau)
 =\sum_{t=1}^{\tau}\ind\{A_t\in\{i(\xi),j(\xi)\}\}
 \le\tau;
\]
the last inequality is \eqref{lower:paired-cost}.

Equations \eqref{lower:paired-correct} and \eqref{lower:paired-error}
give
\[
 \mathbb Q_+(A)\ge\frac34,
 \qquad
 \mathbb Q_-(A)\le\frac{\delta}{4(\alpha+\delta)}\le\frac14.
\]
By \eqref{lower:paired-kl},
$\operatorname{kl}(\mathbb Q_+(A),\mathbb Q_-(A))<\infty$.
Together with $\mathbb Q_+(A)>0$, this implies $\mathbb Q_-(A)>0$.
Applying the binary entropy bound
$\operatorname{kl}(p,q)\ge p\log(1/q)-\log2$ proved in
Lemma~\ref{lower:mixture-lemma}, we obtain
\begin{align*}
 \frac{r^2}{8(\alpha+\delta)}\E_\nu[\tau]
 &\ge\operatorname{kl}\bigl(\mathbb Q_+(A),\mathbb Q_-(A)\bigr)\\
 &\ge\mathbb Q_+(A)\log\frac{1}{\mathbb Q_-(A)}-\log2\\
 &\ge\frac34\log\frac{4(\alpha+\delta)}{\delta}-\log2\\
 &=\frac34\log\frac{\alpha+\delta}{\delta}
       +\frac12\log2\\
 &\ge\frac{\log2}{2}
       \left[1+\log\frac{\alpha+\delta}{\delta}\right].
\end{align*}
The last inequality uses $\log((\alpha+\delta)/\delta)\ge0$ and
$3/4\ge(\log2)/2$.
\paragraph{Step 4: Substitute the Gap Cutoff.}
Since $r=8(\alpha+\delta)/\kappa_\rho$, the preceding inequality
implies
\begin{align*}
 \E_\nu[\tau]
 &\ge\frac{\log2}{16}
       \frac{\kappa_\rho^2}{\alpha+\delta}
       \left[1+\log\frac{\alpha+\delta}{\delta}\right]\\
 &\ge\frac{\log2}{64}
       \frac{\kappa_\nu^2}{\alpha+\delta}
       \left[1+\log\frac{\alpha+\delta}{\delta}\right].
\end{align*}
This proves \eqref{lower:all-path-bound}. Its right-hand side and its
small-budget threshold are independent of the policy, so taking the
infimum proves the assertion for $\Tstar$.
\end{proof}

\subsection{A Sharp Lower Bound When the Abstention Budget Tends to Zero}
\label{lb:separated}

For $\alpha\downarrow0$ and $\alpha/\delta\to\infty$, we first
bound the error of every policy with a deterministic sampling horizon. That is, we first give a lower bound of error probability for the Bayesian Fixed-Budget Best Arm Identification problem with Abstention as in \citet{HHT26}.
We then use that to obtain a lower bound for Fixed-Confidence case.

The next lemma proves a  finite-horizon inequality in the proof of
\citet[Lemma~C.12]{HHT26}, with the subprior constructed above.

\begin{lemma}[Finite-Horizon Error Bound]
\label{lb:finite-horizon}
Let
\begin{equation}\label{lb:fixed-budget-value}
 \mathcal E_{\nu,T}^\star(\alpha^{\prime})
 :=\inf_{\pi:\,\Pp_\nu^\pi(D=?)\le\alpha^{\prime}}
       \Pp_\nu^\pi(D\in[K],\ D\ne a^\star),
\end{equation}
where the infimum is over all policies taking exactly $T$ individual
pulls. Fix $\rho,\eta\in(0,1)$ and use the construction of
Lemma~\ref{lb:flat-subprior}. There is a finite $T_0(\rho,\eta)$ such that
for every integer $T\ge T_0(\rho,\eta)$ and every
$0<\alpha^{\prime}<(1-\eta)\kappa_\rho\varepsilon_\rho/4$,
\begin{equation}\label{lb:finite-horizon-inequality}
 \mathcal E_{\nu,T}^\star(\alpha^{\prime})
 \ge\frac{2(1-\eta)\kappa_\rho}{\sqrt T}
       \int_{\frac{\alpha^{\prime}\sqrt T}{2(1-\eta)\kappa_\rho}}
            ^{\frac{\varepsilon_\rho\sqrt T}{4}}
                \Phi(-u)\,\dd u.
\end{equation}
The threshold $T_0(\rho,\eta)$ does not depend on $\alpha^{\prime}$.
\end{lemma}

\begin{proof}
The subprior notation of \citet{HHT26} corresponds to ours through
$c_\rho=\kappa_\rho$, $\delta_\rho=\varepsilon_\rho$, and $\Theta=d$.

\paragraph{Step 1: Apply the Gaussian Sign Reduction.}
Fix a policy $\pi$ taking exactly $T$ pulls and satisfying
$\Pp_\nu^\pi(D=?)\le\alpha^{\prime}$.
Reveal the background $\xi=(i,j,m,z)\in\mathcal C_\rho$ to the
learner: the identities of the top two arms, their midpoint, and
the means of all remaining arms. The signed gap $d$ remains unknown.
For $0<|d|<\varepsilon_\rho$, the best arm is
\[
 a^\star(\mu(\xi,d))=
 \begin{cases}
  i,&d>0,\\
  j,&d<0.
 \end{cases}
\]
By \citet[Lemma~C.5]{HHT26}, there exists an oracle policy that uses
this revealed background, takes exactly $T$ pulls in total from
$\{i,j\}$, and outputs $S\in\{-1,+1,?\}$. Where $S$ outputs $+1$ and $-1$ means we recommend
arms $i$ and $j$, respectively, while $?$ denotes abstention.
Let $\Pp_{\xi,d}^{\mathrm{or}}$ denote the oracle policy's law at instance
$\mu(\xi,d)$. This oracle policy, without knowing $d$, satisfies
for every $0<|d|<\varepsilon_\rho$,
\begin{align}
 \Pp_{\xi,d}^{\mathrm{or}}(S=?)
 &=\Pp_{\mu(\xi,d)}^\pi(D=?),\notag\\
 \Pp_{\xi,d}^{\mathrm{or}}(S=-\operatorname{sgn}(d))
 &\le\Pp_{\mu(\xi,d)}^\pi
       (D\in[K],D\ne a^\star(\mu(\xi,d))).
 \label{lb:oracle-risk-comparison}
\end{align}

For this oracle policy, let $A_t\in\{i,j\}$ and $X_t$ be the arm
selected and reward observed at round $t$. Define
\[
 Z_t:=
 \begin{cases}
  X_t-m,&A_t=i,\\
  m-X_t,&A_t=j,
 \end{cases}
 \qquad
 V:=\frac1{\sqrt T}\sum_{t=1}^T Z_t.
\]
By \citet[Lemma~C.6]{HHT26}, under $\Pp_{\xi,d}^{\mathrm{or}}$,
\[
 (Z_1,\ldots,Z_T)\sim\mathcal N(d/2,1)^{\otimes T},
 \qquad V\sim\mathcal N(d\sqrt T/2,1).
\]
By \citet[Lemmas~C.7--C.8]{HHT26}, there are measurable functions
$(\xi,v)\mapsto\psi_s^\xi(v)$ for $s\in\{-1,+1,?\}$,
which do not depend on $d$ and satisfy
\[
 0\le\psi_s^\xi(v)\le1,\qquad
 \psi_{-1}^\xi(v)+\psi_{+1}^\xi(v)+\psi_?^\xi(v)=1,
\]
such that, for every $0<|d|<\varepsilon_\rho$ and
$s\in\{-1,+1,?\}$,
\begin{equation}\label{lb:oracle-output-matching}
 \int_\R\psi_s^\xi(v)\phi(v-d\sqrt T/2)\,\dd v
 =\Pp_{\xi,d}^{\mathrm{or}}(S=s),
\end{equation}
where $\phi$ is the standard gaussian density. This randomized rule $\psi$ observes $(\xi,V)$ and outputs $s$
with probability $\psi_s^\xi(V)$. Essentially, for each fixed \((\xi,d)\), the rule based on \((\xi,V)\) has the same probability of each output as the oracle policy using its full observation history.

Define its abstention and sign-error masses under $Q_\rho$ by
\begin{align}
 \mathcal A_{Q_\rho}(\psi)
 &:=\int_{\mathcal C_\rho}\int_{0<|d|<\varepsilon_\rho}\int_\R
     \psi_?^\xi(v)\phi(v-d\sqrt T/2)
                 \,\dd v\,\dd d\,\Lambda_\rho(\dd\xi),
       \label{lb:oracle-abstention}\\
 \mathcal E_{Q_\rho}(\psi)
 &:=\int_{\mathcal C_\rho}\int_{0<|d|<\varepsilon_\rho}\int_\R
     \psi_{-\operatorname{sgn}(d)}^\xi(v)\phi(v-d\sqrt T/2)
                 \,\dd v\,\dd d\,\Lambda_\rho(\dd\xi).
       \label{lb:oracle-error}
\end{align}
Integrating \eqref{lb:oracle-output-matching}, using \eqref{lb:oracle-abstention} and
\eqref{lb:oracle-risk-comparison} and Lemma~\ref{lb:risk-domination}
gives
\begin{align*}
 \mathcal A_{Q_\rho}(\psi)
 &=\int\Pp_\mu^\pi(D=?)\,Q_\rho(\dd\mu)
 \le\Pp_\nu^\pi(D=?)\le\alpha^{\prime},\\
 \mathcal E_{Q_\rho}(\psi)
 &\le\int\Pp_\mu^\pi(D\in[K],D\ne a^\star(\mu))\,Q_\rho(\dd\mu)
 \le\Pp_\nu^\pi(D\in[K],D\ne a^\star).
\end{align*}

\paragraph{Step 2: Apply the Finite-Integral Bound.}
Define the Gaussian densities integrated over positive and negative
signed gaps, after the substitution $u=|d|\sqrt T/2$, by
\begin{equation}\label{lb:gaussian-weights}
 A_T(v):=\int_0^{\varepsilon_\rho\sqrt T/2}\phi(v-u)\,\dd u,
 \qquad
 B_T(v):=\int_0^{\varepsilon_\rho\sqrt T/2}\phi(v+u)\,\dd u.
\end{equation}
Rearranging \eqref{lb:oracle-abstention}, we can write
\begin{equation}\label{lb:oracle-abstention-weights}
 \mathcal A_{Q_\rho}(\psi)
 =\frac2{\sqrt T}\int_{\mathcal C_\rho}\int_\R
    \psi_?^\xi(v)\{A_T(v)+B_T(v)\}\,\dd v\,\Lambda_\rho(\dd\xi).
\end{equation}

Choose $T_0(\rho,\eta)$ so that, for every integer $T\ge T_0(\rho,\eta)$,
\[
 2\Phi(-\varepsilon_\rho\sqrt T/4)\le\eta,
 \qquad
 \frac{\Phi(-\varepsilon_\rho\sqrt T/2)}
      {\Phi(-\varepsilon_\rho\sqrt T/4)}\le\eta.
\]
Such a finite $T_0(\rho,\eta)$ exists because both left-hand sides
tend to zero as $T\to\infty$. It depends on $\varepsilon_\rho$ and
$\eta$, and is independent of $\alpha^{\prime}$ and the policy $\pi$.
By \citet[Lemma~C.10]{HHT26}, these conditions imply
that, for every $T\ge T_0(\rho,\eta)$ and every
$|v|\le\varepsilon_\rho\sqrt T/4$,
\begin{equation}\label{lb:central-weights}
 A_T(v)+B_T(v)\ge1-\eta,
 \qquad
 \min\{A_T(v),B_T(v)\}\ge(1-\eta)\Phi(-|v|).
\end{equation}

Using $\mathcal A_{Q_\rho}(\psi)\le\alpha^{\prime}$ from Step~1,
\eqref{lb:oracle-abstention-weights}, and the first inequality in
\eqref{lb:central-weights}, we obtain
\begin{align}
 \int_{\mathcal C_\rho}
 \int_{|v|\le\varepsilon_\rho\sqrt T/4}
     \psi_?^\xi(v)\,\dd v\,\Lambda_\rho(\dd\xi)
 &\le\frac{\alpha^{\prime}\sqrt T}{2(1-\eta)}\notag\\
 &=2\Lambda_\rho(\mathcal C_\rho)
      \frac{\alpha^{\prime}\sqrt T}{2(1-\eta)\kappa_\rho}.
 \label{lb:central-abstention-mass}
\end{align}
The equality uses $2\Lambda_\rho(\mathcal C_\rho)=\kappa_\rho$.
The assumed bound on $\alpha^{\prime}$ also gives
\[
 0<\frac{\alpha^{\prime}\sqrt T}{2(1-\eta)\kappa_\rho}
   <\frac{\varepsilon_\rho\sqrt T}{8}
   <\frac{\varepsilon_\rho\sqrt T}{4}.
\]
Apply the abstention rearrangement inequality of
\citet[Lemma~C.11]{HHT26} to the function $\psi_?^\xi(v)$ on
$\mathcal C_\rho\times[-\varepsilon_\rho\sqrt T/4,
\varepsilon_\rho\sqrt T/4]$, with background measure $\Lambda_\rho$
and weight $\Phi(-|v|)$. The weight is even, nonnegative, and
nonincreasing in $|v|$, and \eqref{lb:central-abstention-mass}
is the required constraint on the abstention function. That lemma
gives
\begin{align}
 \int_{\mathcal C_\rho}
   \int_{|v|\le\varepsilon_\rho\sqrt T/4}
      (1-\psi_?^\xi(v))\Phi(-|v|)\,\dd v\,\Lambda_\rho(\dd\xi)
      \ge
      \kappa_\rho
       \int_{\frac{\alpha^{\prime}\sqrt T}{2(1-\eta)\kappa_\rho}}
            ^{\frac{\varepsilon_\rho\sqrt T}{4}}
          \Phi(-u)\,\dd u.
 \label{lb:central-rearrangement}
\end{align}
Finally, Step~1, \citet[Lemma~C.9]{HHT26},
\eqref{lb:central-weights}, and \eqref{lb:central-rearrangement},
respectively, give
\begin{align*}
 \Pp_\nu^\pi(D\in[K],D\ne a^\star)
 &\ge\mathcal E_{Q_\rho}(\psi)\\
 &\ge\frac2{\sqrt T}\int_{\mathcal C_\rho}\int_\R
       (1-\psi_?^\xi(v))\min\{A_T(v),B_T(v)\}
          \,\dd v\,\Lambda_\rho(\dd\xi)\\
 &\ge\frac{2(1-\eta)}{\sqrt T}
       \int_{\mathcal C_\rho}
       \int_{|v|\le\varepsilon_\rho\sqrt T/4}
       (1-\psi_?^\xi(v))\Phi(-|v|)\,\dd v\,\Lambda_\rho(\dd\xi)\\
 &\ge\frac{2(1-\eta)\kappa_\rho}{\sqrt T}
       \int_{\frac{\alpha^{\prime}\sqrt T}{2(1-\eta)\kappa_\rho}}
            ^{\frac{\varepsilon_\rho\sqrt T}{4}}
          \Phi(-u)\,\dd u.
\end{align*}
Taking the infimum over the policies in
\eqref{lb:fixed-budget-value} proves
\eqref{lb:finite-horizon-inequality}.
\end{proof}


\begin{theorem}[Lower Bound for Separated Shrinking Paths]
\label{lower:separated}
For every sequence $(\delta_n,\alpha_n)$ satisfying
$\alpha_n\to0$ and $\alpha_n/\delta_n\to\infty$,
\begin{equation}\label{lower:separated-bound}
 \liminf_{n\to\infty}
 \frac{\alpha_n\Tstar(\delta_n,\alpha_n)}
      {\log(\alpha_n/\delta_n)}
 \ge8\kappa_\nu^2.
\end{equation}
\end{theorem}

\begin{proof}
\paragraph{Step 1: Prove a Fixed-Horizon Impossibility Bound.}
We choose horizons $T_n(r)$ at which every fixed-budget policy
with abstention probability at most $r\alpha_n$ has error
greater than $\delta_n$.
Fix $\rho,\eta\in(0,1)$ and the subprior from
Lemma~\ref{lb:flat-subprior}, and choose
\[
 0<C<8(1-\eta)^2\kappa_\rho^2.
\]
The constants $\kappa_\rho$ and $\varepsilon_\rho$ remain fixed as
$n\to\infty$. For a fixed $r>1$ and all sufficiently large $n$, set
\begin{equation}\label{lb:truncation-horizon}
 T_n(r):=\left\lfloor
    \frac{C\log(\alpha_n/\delta_n)}{r^2\alpha_n^2}
                      \right\rfloor.
\end{equation}
Our goal is to prove
\begin{equation}\label{lb:fixed-budget-impossible}
 \mathcal E_{\nu,T_n(r)}^\star(r\alpha_n)>\delta_n
 \qquad\text{for all sufficiently large }n.
\end{equation}
The parameter $r$ determines the extra abstention allowance
$r\alpha_n-\alpha_n=(r-1)\alpha_n$. Step~2 will use
\eqref{lb:fixed-budget-impossible} to prove
$\Pp_\nu(\tau>T_n(r))>(r-1)\alpha_n$, and Step~3 will integrate
these bounds over $r$. Throughout Step~1, $r$ is fixed as $n\to\infty$.

We prove the stronger statement
\[
 \frac{\mathcal E_{\nu,T_n(r)}^\star(r\alpha_n)}{\delta_n}
 \longrightarrow\infty.
\]
Since $T_n(r)\to\infty$ and $r\alpha_n\to0$, for all sufficiently
large $n$ we have
\[
 T_n(r)\ge T_0(\rho,\eta),\qquad
 0<r\alpha_n<(1-\eta)\kappa_\rho\varepsilon_\rho/4.
\]
Lemma~\ref{lb:finite-horizon} therefore gives the explicit integral
lower bound
\begin{equation}\label{lb:finite-horizon-at-truncation}
 \frac{\mathcal E_{\nu,T_n(r)}^\star(r\alpha_n)}{\delta_n}
 \ge\frac{2(1-\eta)\kappa_\rho}{\delta_n\sqrt{T_n(r)}}
       \int_{\frac{r\alpha_n\sqrt{T_n(r)}}{2(1-\eta)\kappa_\rho}}
            ^{\frac{\varepsilon_\rho\sqrt{T_n(r)}}{4}}
           \Phi(-u)\,\dd u.
\end{equation}
We next bound the integration interval, the Gaussian integral,
and the remaining factors in this expression.

First, denote the lower limit of integration in \eqref{lb:finite-horizon-at-truncation} by
\[
 z_n:=\frac{r\alpha_n\sqrt{T_n(r)}}{2(1-\eta)\kappa_\rho}.
\]
The floor in \eqref{lb:truncation-horizon} gives
\begin{equation}\label{lb:truncation-endpoint-bounds}
 \frac{C\log(\alpha_n/\delta_n)-r^2\alpha_n^2}
      {4(1-\eta)^2\kappa_\rho^2}
 \le z_n^2
 \le\frac{C\log(\alpha_n/\delta_n)}
            {4(1-\eta)^2\kappa_\rho^2}.
\end{equation}
Consequently, for every fixed $r>1$, as $n\to\infty$,
\[
 z_n\longrightarrow\infty,\qquad
 \frac{z_n^2}{2\log(\alpha_n/\delta_n)}
 \longrightarrow\frac{C}{8(1-\eta)^2\kappa_\rho^2}<1.
\]
For the upper limit of integration in \eqref{lb:finite-horizon-at-truncation}, its ratio to $z_n$ satisfies
\[
 \frac{\varepsilon_\rho\sqrt{T_n(r)}/4}{z_n}
 =\frac{\varepsilon_\rho(1-\eta)\kappa_\rho}{2r\alpha_n}
 \longrightarrow\infty.
\]
Thus, eventually $z_n\ge1$ and
\[
 z_n+z_n^{-1}\le2z_n
 \le\frac{\varepsilon_\rho\sqrt{T_n(r)}}4.
\]
We may therefore lower bound the integral in
\eqref{lb:finite-horizon-at-truncation} by its integral over
$[z_n,z_n+z_n^{-1}]$.

Second, integration by parts gives, for $x>0$,
\[
 \Phi(-x)=\frac{\phi(x)}x-\int_x^\infty\frac{\phi(y)}{y^2}\,\dd y
 \ge\frac{\phi(x)}x-\frac{\Phi(-x)}{x^2},
 \qquad
 \Phi(-x)\ge\frac{x\phi(x)}{x^2+1}.
\]
For $z\ge1$, it follows that
\begin{equation}\label{lb:short-gaussian-tail}
 \int_z^{z+z^{-1}}\Phi(-u)\,\dd u
 \ge\frac1z\Phi(-z-z^{-1})
 \ge\frac{e^{-3/2}}{4\sqrt{2\pi}}\,z^{-2}e^{-z^2/2}.
\end{equation}
Here we used
\[
 z+z^{-1}\le2z,\qquad
 \frac{(z+z^{-1})^2}{2}\le\frac{z^2}{2}+\frac32,
 \qquad
 \frac{x}{x^2+1}\ge\frac1{2x}\quad(x\ge1).
\]
Substituting \eqref{lb:short-gaussian-tail} into
\eqref{lb:finite-horizon-at-truncation} yields
\begin{equation}\label{lb:finite-horizon-factor-bound}
 \frac{\mathcal E_{\nu,T_n(r)}^\star(r\alpha_n)}{\delta_n}
 \ge\frac{e^{-3/2}(1-\eta)\kappa_\rho}{2\sqrt{2\pi}}
       \frac1{\delta_n}\,
       \frac1{\sqrt{T_n(r)}}\,
       \frac1{z_n^2}\,e^{-z_n^2/2}.
\end{equation}
Finally, \eqref{lb:truncation-horizon} and
\eqref{lb:truncation-endpoint-bounds} bound these factors by
\begin{align*}
 \frac1{\sqrt{T_n(r)}}
 &\ge\frac{r\alpha_n}{\sqrt C\,[\log(\alpha_n/\delta_n)]^{1/2}},\\
 \frac1{z_n^2}
 &\ge\frac{4(1-\eta)^2\kappa_\rho^2}
             {C\log(\alpha_n/\delta_n)},\\
 e^{-z_n^2/2}
 &\ge\left(\frac{\alpha_n}{\delta_n}\right)^{
          -C/[8(1-\eta)^2\kappa_\rho^2]}.
\end{align*}
Combining these inequalities gives
\[
 \frac{\mathcal E_{\nu,T_n(r)}^\star(r\alpha_n)}{\delta_n}
 \ge
 \frac{2e^{-3/2}(1-\eta)^3\kappa_\rho^3 r}
      {\sqrt{2\pi}\,C^{3/2}}
 \frac{(\alpha_n/\delta_n)^{
             1-C/[8(1-\eta)^2\kappa_\rho^2]}}
      {[\log(\alpha_n/\delta_n)]^{3/2}}
 \longrightarrow\infty,
\]
because $1-C/[8(1-\eta)^2\kappa_\rho^2]>0$ and
$\alpha_n/\delta_n\to\infty$. This proves
\eqref{lb:fixed-budget-impossible} for each fixed $r>1$.

\paragraph{Step 2: Bound the Stopping-Time Tail by Truncation.}
Take any $(\delta_n,\alpha_n)$-feasible policy and run it for at most
$T_n(r)$ pulls. Define its truncated output by
\[
 \widetilde D=
 \begin{cases}
  D,&\tau\le T_n(r),\\
  ?,&\tau>T_n(r).
 \end{cases}
\]
If it stops early, make arbitrary additional pulls without changing
$\widetilde D$. This defines a Fixed-Budget algorithm with exactly $T_n(r)$ pulls.
Its error and abstention probabilities satisfy
\begin{align*}
 \Pp_\nu(\widetilde D\in[K],\widetilde D\ne a^\star)
 &=\Pp_\nu(\tau\le T_n(r),D\in[K],D\ne a^\star)
 \le\delta_n,\\
 \Pp_\nu(\widetilde D=?)
 &=\Pp_\nu(\tau\le T_n(r),D=?)+\Pp_\nu(\tau>T_n(r))
 \le\alpha_n+\Pp_\nu(\tau>T_n(r)).
\end{align*}
If $\Pp_\nu(\tau>T_n(r))\le(r-1)\alpha_n$, then
\[
 \Pp_\nu(\widetilde D=?)
 \le\alpha_n+(r-1)\alpha_n
 =r\alpha_n.
\]
Thus the truncated policy uses exactly $T_n(r)$ pulls and has
abstention probability at most $r\alpha_n$. It is therefore
among the policies defining
$\mathcal E_{\nu,T_n(r)}^\star(r\alpha_n)$, so
\[
 \mathcal E_{\nu,T_n(r)}^\star(r\alpha_n)
 \le\Pp_\nu(\widetilde D\in[K],\widetilde D\ne a^\star)
 \le\delta_n,
\]
contradicting \eqref{lb:fixed-budget-impossible}. Hence, for all sufficiently large $n$,
\begin{equation}\label{lb:stopping-tail}
 \Pp_\nu(\tau>T_n(r))>(r-1)\alpha_n.
\end{equation}
The threshold in this assertion is uniform over all feasible policies,
because \eqref{lb:fixed-budget-impossible} bounds their infimum.

\paragraph{Step 3: Integrate the Stopping-Time Tail.}
Choose a feasible policy $\pi_n$ with
\[
 \E_\nu^{\pi_n}[\tau_n]\le\Tstar(\delta_n,\alpha_n)+1.
\]
Such policies exist because the upper bound gives a finite feasible
expected cost. Fix $R>1$. The tail integral formula and the change of
variable $t=C\log(\alpha_n/\delta_n)/(r^2\alpha_n^2)$ give
\begin{align}
 \E_\nu^{\pi_n}[\tau_n]
 &=\int_0^\infty\Pp_\nu^{\pi_n}(\tau_n>t)\,\dd t\notag\\
 &\ge
\int_{\frac{C\log(\alpha_n/\delta_n)}{R^2\alpha_n^2}}
     ^{\frac{C\log(\alpha_n/\delta_n)}{\alpha_n^2}}
  \mathbb P_\nu^{\pi_n}(\tau_n>t)\,\dd t.\\
 &\ge\frac{2C\log(\alpha_n/\delta_n)}{\alpha_n^2}
       \int_1^R\Pp_\nu^{\pi_n}(\tau_n>T_n(r))\frac{\dd r}{r^3}.
 \label{lb:tail-integration}
\end{align}
Here the first inequality restricted the integral range, the last inequality used
$\{\tau_n>t\}=\{\tau_n>\lfloor t\rfloor\}$, since $\tau_n$ is
integer-valued.
For each fixed $r\in(1,R)$, \eqref{lb:stopping-tail} implies
\[
 \liminf_{n\to\infty}
 \frac{\Pp_\nu^{\pi_n}(\tau_n>T_n(r))}{\alpha_n}\ge r-1.
\]
Multiplying \eqref{lb:tail-integration} by
$\alpha_n/\log(\alpha_n/\delta_n)$ and applying Fatou's lemma gives
\begin{align*}
 \liminf_{n\to\infty}
 \frac{\alpha_n\E_\nu^{\pi_n}[\tau_n]}
      {\log(\alpha_n/\delta_n)}
 &\ge2C\int_1^R\frac{r-1}{r^3}\,\dd r\\
 &=C(1-R^{-1})^2.
\end{align*}
By the choice of $\pi_n$,
\[
 0\le
 \frac{\alpha_n\bigl(\E_\nu^{\pi_n}[\tau_n]
              -\Tstar(\delta_n,\alpha_n)\bigr)}
      {\log(\alpha_n/\delta_n)}
 \le\frac{\alpha_n}{\log(\alpha_n/\delta_n)}
 \longrightarrow0.
\]
Thus the same lower bound holds with
$\E_\nu^{\pi_n}[\tau_n]$ replaced by
$\Tstar(\delta_n,\alpha_n)$.

Let $R\to\infty$, then let
$C\uparrow8(1-\eta)^2\kappa_\rho^2$. Finally let $\eta\downarrow0$
and $\rho\downarrow0$, using \eqref{lb:flat-kappa}. This proves~\eqref{lower:separated-bound}.
\end{proof}

\subsection{Proof of Corollary~\ref{thm:rate}}

Combining the lower bounds with the \textup{\textsc{Screen-and-Confirm}} upper bounds gives the
matching results for the optimal expected sampling cost.

\begin{proof}

\paragraph{Arbitrary path.} Combining \eqref{main:uniform-upper} and \eqref{eq:general-lower-main} gives \eqref{eq:main-global}.

\paragraph{Shrinking Abstention Budgets and Power Paths.}
For every sequence $\alpha_n\to0$ and $\alpha_n/\delta_n\to\infty$,
Theorems~\ref{lower:separated} and \ref{asym:separated} give
\begin{align*}
 8\kap^2
 \le\liminf_{n\to\infty}
       \frac{\alpha_n\Tstar(\delta_n,\alpha_n)}
            {\log(\alpha_n/\delta_n)}
 \le\limsup_{n\to\infty}
       \frac{\alpha_n\Tstar(\delta_n,\alpha_n)}
            {\log(\alpha_n/\delta_n)}
 \le\limsup_{n\to\infty}
       \frac{\alpha_n\E_\nu[\tau_{\delta_n,\alpha_n}^{\SCalg}]}
            {\log(\alpha_n/\delta_n)}
 \le8\kap^2.
\end{align*}
This proves \eqref{eq:main-lower-separated}. For $\alpha=\delta^p$
with fixed $0<p<1$, we have $\alpha/\delta\to\infty$ and
\[
 \frac{\delta^p\Tstar(\delta,\delta^p)}{\log(1/\delta)}
 =(1-p)\frac{\alpha\Tstar(\delta,\alpha)}
                  {\log(\alpha/\delta)}.
\]
Equation~\eqref{eq:main-power-value} follows.

\paragraph{Fixed Positive Abstention.}
\textup{\textsc{Screen-and-Confirm}} is feasible by Theorem~\ref{thm:upper-rate}, so
\begin{equation}\label{lower:value-upper}
 0\le\Tstar(\delta,\alpha)
 \le\E_\nu[\tau_{\delta,\alpha}^{\SCalg}].
\end{equation}
Theorems~\ref{lower:fixed} and \ref{asym:fixed-alpha} give
\begin{align*}
 V_\nu(\alpha)
 \le\liminf_{\delta\downarrow0}
       \frac{\Tstar(\delta,\alpha)}{\log(1/\delta)}
 \le\limsup_{\delta\downarrow0}
       \frac{\Tstar(\delta,\alpha)}{\log(1/\delta)}
 \le\limsup_{\delta\downarrow0}
       \frac{\E_\nu[\tau_{\delta,\alpha}^{\SCalg}]}{\log(1/\delta)}
 \le V_\nu(\alpha).
\end{align*}
Thus the optimal-value limit exists and equals $V_\nu(\alpha)$,
proving \eqref{eq:main-lower-fixed}.

\paragraph{The Iterated Limit.}
By \eqref{eq:main-lower-fixed}, the inner limit exists for every
$\alpha\in(0,1)$. Equation~\eqref{geom:V-asymptotic} then gives
\[
 \lim_{\alpha\downarrow0}\alpha
 \left[\lim_{\delta\downarrow0}
 \frac{\Tstar(\delta,\alpha)}{\log(1/\delta)}\right]
 =\lim_{\alpha\downarrow0}\alpha V_\nu(\alpha)
 =8\kap^2,
\]
which proves \eqref{eq:main-iterated-value}.

\end{proof}


\section{Frequentist Fixed-Confidence Comparison}
\label{sec:frequentist}

We contrast the Bayesian results with the frequentist setting, where $\mu$ is unknown but fixed.
Call $\pi$ \emph{frequentist $(\delta,\alpha)$-feasible} if, for every
$\mu\in\R^K$ with a unique best arm,
\begin{equation}\label{freq:constraints}
 \Pp_\mu(\tau^\pi<\infty)=1,\qquad
 \Pp_\mu(D^\pi\in[K],D^\pi\ne a^\star(\mu))\le\delta,\qquad
 \Pp_\mu(D^\pi=?)\le\alpha.
\end{equation}

\begin{theorem}[Frequentist Sample Complexity With Abstention]
\label{thm:frequentist}
Fix an instance $\mu$ with a unique best arm.
\begin{enumerate}[label=(\roman*)]
\item For $0\le\alpha<1$ and $0<\delta<(1-\alpha)/2$, every frequentist
$(\delta,\alpha)$-feasible policy $\pi$ satisfies
\begin{equation}\label{freq:finite-lower}
 \E_\mu[\tau^\pi]
 \ge H(\mu)(1-\alpha-2\delta)
       \log\frac{1-\alpha-\delta}{\delta}.
\end{equation}
\item For every fixed $\alpha\in[0,1)$ and every family
$(\pi_{\delta,\alpha})_{0<\delta<(1-\alpha)/2}$ of frequentist
$(\delta,\alpha)$-feasible policies,
\begin{equation}\label{freq:fixed-limit}
 \liminf_{\delta\downarrow0}
 \frac{\E_\mu[\tau^{\pi_{\delta,\alpha}}]}
      {\log(1/\delta)}
 \ge(1-\alpha)H(\mu).
\end{equation}
\end{enumerate}
\end{theorem}

\emph{Proof Sketch.}
Restart a feasible policy with fresh observations and independent
randomization whenever it abstains. The resulting policy never abstains
and has error at most $\delta/(1-\alpha)$ at every instance. The lower bound of \citet[Theorem~1]{GK16} to this policy gives
\eqref{freq:finite-lower}. Dividing by $\log(1/\delta)$ and taking
$\delta\downarrow0$ at fixed $\alpha$ gives \eqref{freq:fixed-limit}.
Appendices~\ref{sec:restart} and~\ref{sec:finite_lb} give the full argument.

\paragraph{Remark.} The coefficient $(1-\alpha)H(\mu)$ is sharp
(Proposition~\ref{freq:sharp-fixed}): abstain initially with probability
$\alpha$ and otherwise run a classical FC-BAI policy with error budget
$\delta/(1-\alpha)$ achives this. A fixed abstention budget therefore reduces the
optimal leading sample complexity by the factor $1-\alpha$.
Along every path with $\alpha,\delta\to0$, the ratio of the optimal cost
to $H(\mu)\log(1/\delta)$ tends to one
(Corollary~\ref{freq:consequences}). Thus a vanishing abstention budget
gives no leading-order improvement under uniform frequentist constraints.

The Bayesian constraint permits abstention to concentrate on difficult
instances. In contrast, \eqref{freq:constraints} requires a recommendation
with probability at least $1-\alpha$ even on instances with arbitrarily
small gaps. Under our Gaussian prior, this forces
$\E_\nu[\tau^\pi]=\infty$ whenever $\alpha+2\delta<1$
(Corollary~\ref{freq:consequences}), whereas the Bayesian-feasible
\textup{\textsc{Screen-and-Confirm}} policy has finite expected cost
by Theorem~\ref{thm:upper-rate}.

Throughout this section, \emph{feasibility} means the uniform constraints
\eqref{freq:constraints}, and all instances under consideration have a
unique best arm. We allow abstention at time zero, using the independent
random seed $U$. Define the optimal frequentist expected sample size at
$\mu$ by
\begin{equation}\label{freq:optimal-value}
 \mathcal T_{\mathrm F}^\star(\mu;\delta,\alpha)
 :=\inf_{\text{frequentist $(\delta,\alpha)$-feasible policies }\pi}
       \E_\mu[\tau^\pi].
\end{equation}
The objective is evaluated at $\mu$, but every policy in this infimum
must satisfy \eqref{freq:constraints} at all unique-best instances.
We first relate this optimal value to that of ordinary fixed-confidence BAI.

\subsection{Removing Abstention by Restarting} \label{sec:restart}

\begin{lemma}[Reduction to a Policy Without Abstention]
\label{freq:reduction}
For $0\le\alpha<1$ and $0<\delta<(1-\alpha)/2$, and every unique-best
instance $\mu$,
\begin{equation}\label{freq:value-identity}
 \mathcal T_{\mathrm F}^\star(\mu;\delta,\alpha)
 =(1-\alpha)\,
 \mathcal T_{\mathrm F}^\star\!\left(
       \mu;\frac{\delta}{1-\alpha},0\right).
\end{equation}
\end{lemma}

\begin{proof}
\paragraph{Step 1: Construct Non-Abstaining Algorithm from Abstaining Algorithm - Restart After Each Abstention.}
Fix a frequentist $(\delta,\alpha)$-feasible policy $\pi$.
Run independent copies of $\pi$, each using fresh observations and a
fresh independent random seed, until a copy recommends an arm.
For copy $m$, let $(\tau^{(m)},D^{(m)})$ be its sample size and decision.
Define the number of copies used, the total sample size, and the
terminal recommendation by
\[
 L:=\inf\{m\ge1:D^{(m)}\ne ?\},\qquad
 \tau^{\mathrm{rep}}:=\sum_{m=1}^{L}\tau^{(m)},\qquad
 D^{\mathrm{rep}}:=D^{(L)}.
\]
These rules define an adaptive policy without knowledge of the instance.

Fix any unique-best instance $\lambda$ and write
$r_\lambda:=\Pp_\lambda(D^\pi=?)\le\alpha$.
Under this fixed instance, the copies are independent and have
the same distribution as $\pi$. Thus, for every integer $m\ge1$,
\[
 \Pp_\lambda(L\ge m)=r_\lambda^{m-1},\qquad
 \Pp_\lambda(L=\infty)=\lim_{m\to\infty}r_\lambda^m=0.
\]
Each copy terminates almost surely, so the repeated policy also
terminates almost surely and never abstains. Its error probability is
\begin{align}
 \Pp_\lambda(D^{\mathrm{rep}}\ne a^\star(\lambda))
 &=\sum_{m=1}^{\infty}r_\lambda^{m-1}
       \Pp_\lambda(D^\pi\in[K],D^\pi\ne a^\star(\lambda))\notag\\
 &=\frac{\Pp_\lambda(D^\pi\in[K],D^\pi\ne a^\star(\lambda))}
          {1-r_\lambda}
 \le\frac{\delta}{1-\alpha}.
 \label{freq:restart-error}
\end{align}
If $\E_\lambda[\tau^\pi]=\infty$, then
$\tau^{\mathrm{rep}}\ge\tau^{(1)}$ implies
$\E_\lambda[\tau^{\mathrm{rep}}]=\infty$.
Otherwise, $\{L\ge m\}$ is determined by the first $m-1$ copies and
is independent of $\tau^{(m)}$. Tonelli's theorem gives
\begin{align}
 \E_\lambda[\tau^{\mathrm{rep}}]
 &=\sum_{m=1}^{\infty}
       \E_\lambda[\tau^{(m)}\ind\{L\ge m\}]\notag\\
 &=\sum_{m=1}^{\infty}r_\lambda^{m-1}\E_\lambda[\tau^\pi]
 =\frac{\E_\lambda[\tau^\pi]}{1-r_\lambda}.
 \label{freq:restart-cost}
\end{align}
This identity also holds when the expectations are infinite

Because \eqref{freq:restart-error} holds at every $\lambda$, the repeated
policy is frequentist $(\delta/(1-\alpha),0)$-feasible. Evaluating
\eqref{freq:restart-cost} at $\mu$ gives
\begin{align*}
 \E_\mu[\tau^\pi]
 &=(1-r_\mu)\E_\mu[\tau^{\mathrm{rep}}]\\
 &\ge(1-r_\mu)\,
       \mathcal T_{\mathrm F}^\star\!\left(
         \mu;\frac{\delta}{1-\alpha},0\right)\\
 &\ge(1-\alpha)\,
       \mathcal T_{\mathrm F}^\star\!\left(
         \mu;\frac{\delta}{1-\alpha},0\right).
\end{align*}
Taking the infimum over $\pi$ proves the lower inequality in
\eqref{freq:value-identity}.

\paragraph{Step 2: Construct Abstaining Algorithm from Non-Abstaining Algorithm - Abstain Before Sampling.}
Conversely, let $\pi_0$ be any frequentist
$(\delta/(1-\alpha),0)$-feasible policy. Draw a Bernoulli variable $B$
with $\Pp(B=1)=\alpha$, independently of the observations and of $\pi_0$.
Set $(\tau,D)=(0,?)$ if $B=1$, and otherwise run $\pi_0$.
For every unique-best $\lambda$,
\begin{align*}
 \Pp_\lambda(D=?)&=\alpha,\\
 \Pp_\lambda(D\in[K],D\ne a^\star(\lambda))
 &=(1-\alpha)\Pp_\lambda(D^{\pi_0}\ne a^\star(\lambda))
 \le\delta,\\
 \E_\lambda[\tau]&=(1-\alpha)\E_\lambda[\tau^{\pi_0}].
\end{align*}
The policy terminates almost surely. Taking the infimum over $\pi_0$
gives the reverse inequality in \eqref{freq:value-identity}.
\end{proof}

\subsection{The Finite Lower Bound and the Fixed-Abstention Limit} \label{sec:finite_lb}

\begingroup
\renewcommand{\proofname}{Proof of Theorem~\ref{thm:frequentist}}
\begin{proof}
\paragraph{Step 1: General $(\delta,\alpha)$-feasible policy}
Set $\varepsilon:=\delta/(1-\alpha)\in(0,1/2)$.
For every frequentist $(\varepsilon,0)$-feasible policy $\pi_0$,
\citet[Theorem~1]{GK16} applying to unit-variance Gaussian rewards gives
\begin{equation}\label{freq:classical-lower}
 \E_\mu[\tau^{\pi_0}]
 \ge H(\mu)\operatorname{kl}(1-\varepsilon,\varepsilon),
\end{equation}
where
$\operatorname{kl}(p,q):=p\log(p/q)+(1-p)\log((1-p)/(1-q))$.
For a frequentist $(\delta,\alpha)$-feasible policy $\pi$, apply
\eqref{freq:classical-lower} to the repeated policy constructed from $\pi$ in
Lemma~\ref{freq:reduction}. With
$r_\mu:=\Pp_\mu(D^\pi=?)\le\alpha$, \eqref{freq:restart-cost} gives
\begin{align*}
 \E_\mu[\tau^\pi]
 &=(1-r_\mu)\E_\mu[\tau^{\mathrm{rep}}]\\
 &\ge(1-\alpha)H(\mu)
       \operatorname{kl}(1-\varepsilon,\varepsilon).
\end{align*}
Expanding the binary relative entropy gives
\[
 (1-\alpha)\operatorname{kl}(1-\varepsilon,\varepsilon)
 =(1-\alpha)(1-2\varepsilon)
       \log\frac{1-\varepsilon}{\varepsilon}
 =(1-\alpha-2\delta)\log\frac{1-\alpha-\delta}{\delta}.
\]
This proves \eqref{freq:finite-lower}.

\paragraph{Step 2: Fixed Abstention Case.}
Fix $\alpha\in[0,1)$ and any family
$(\pi_{\delta,\alpha})_{0<\delta<(1-\alpha)/2}$ as in the theorem.
For each $\delta$, \eqref{freq:finite-lower} gives
\[
 \frac{\E_\mu[\tau^{\pi_{\delta,\alpha}}]}{\log(1/\delta)}
 \ge H(\mu)(1-\alpha-2\delta)
 \left[1+\frac{\log(1-\alpha-\delta)}{\log(1/\delta)}\right].
\]
As $\delta\downarrow0$, the factor $1-\alpha-2\delta$ tends to
$1-\alpha$, and the bracket tends to one because
$\log(1-\alpha-\delta)\to\log(1-\alpha)\in\R$.
Taking the lower limit proves \eqref{freq:fixed-limit}.
\end{proof}
\endgroup

We next show that the coefficient in \eqref{freq:fixed-limit} is sharp (attainable).

\begin{proposition}[Sharpness at Fixed Abstention]
\label{freq:sharp-fixed}
For every unique-best instance $\mu$ and every fixed $\alpha\in[0,1)$,
\begin{equation}\label{freq:optimal-fixed-limit}
 \lim_{\delta\downarrow0}
 \frac{\mathcal T_{\mathrm F}^\star(\mu;\delta,\alpha)}
      {\log(1/\delta)}
 =(1-\alpha)H(\mu).
\end{equation}
\end{proposition}

\begin{proof}
\paragraph{Step 1: The Non-Abstaining Optimal-Value.}
We first prove the non-abstaining FC-BAI result:
\begin{equation}\label{freq:classical-limit}
 \lim_{\varepsilon\downarrow0}
 \frac{\mathcal T_{\mathrm F}^\star(\mu;\varepsilon,0)}
      {\log(1/\varepsilon)}=H(\mu).
\end{equation}
Indeed, the lower bound follows from
\eqref{freq:classical-lower} and
\[
 \lim_{\varepsilon\downarrow0}
 \frac{\operatorname{kl}(1-\varepsilon,\varepsilon)}
      {\log(1/\varepsilon)}=1.
\]
For the upper bound, \citet[Proposition~12 and Theorem~14]{GK16}
provide, for every fixed $\gamma>1$, a family of frequentist
$(\varepsilon,0)$-feasible policies $\pi_{\varepsilon,\gamma}$ such that
\[
 \limsup_{\varepsilon\downarrow0}
 \frac{\E_\mu[\tau^{\pi_{\varepsilon,\gamma}}]}
      {\log(1/\varepsilon)}\le\gamma H(\mu).
\]
Taking $\gamma\downarrow1$ proves the upper bound in
\eqref{freq:classical-limit}.

\paragraph{Step 2: Transfer to Fixed Abstention.}
For fixed $\alpha\in[0,1)$, we have
$\varepsilon=\delta/(1-\alpha)\downarrow0$ and
\[
 \frac{\log(1/\varepsilon)}{\log(1/\delta)}
 =1+\frac{\log(1-\alpha)}{\log(1/\delta)}\longrightarrow1.
\]
Using \eqref{freq:value-identity}, write
\[
 \frac{\mathcal T_{\mathrm F}^\star(\mu;\delta,\alpha)}
      {\log(1/\delta)}
 =(1-\alpha)
   \frac{\mathcal T_{\mathrm F}^\star(\mu;\varepsilon,0)}
        {\log(1/\varepsilon)}
   \frac{\log(1/\varepsilon)}{\log(1/\delta)}.
\]
The middle factor tends to $H(\mu)$ by
\eqref{freq:classical-limit}, proving \eqref{freq:optimal-fixed-limit}.
\end{proof}

\subsection{Vanishing Abstention and Prior-Averaged Cost}

\begin{corollary}[Consequences of Uniform Frequentist Guarantees]
\label{freq:consequences}
For every fixed unique-best instance $\mu$, along every path with
$\alpha,\delta\to0$,
\begin{equation}\label{freq:vanishing-limit}
 \lim_{\substack{\alpha\to0\\\delta\to0}}
 \frac{\mathcal T_{\mathrm F}^\star(\mu;\delta,\alpha)}
      {H(\mu)\log(1/\delta)}=1.
\end{equation}
Under the Gaussian prior \eqref{eq:prior}, every frequentist
$(\delta,\alpha)$-feasible policy satisfies
\begin{equation}\label{freq:infinite-bayes-cost}
 \E_\nu[\tau^\pi]=\infty
 \qquad\text{if $\alpha\ge0$, $\delta>0$, and $\alpha+2\delta<1$.}
\end{equation}
\end{corollary}

\begin{proof}
Along any path in \eqref{freq:vanishing-limit}, eventually
$\delta<(1-\alpha)/2$. Set $\varepsilon=\delta/(1-\alpha)$.
Then $\varepsilon\to0$, $1-\alpha\to1$, and
\[
 \frac{\log(1/\varepsilon)}{\log(1/\delta)}
 =1+\frac{\log(1-\alpha)}{\log(1/\delta)}\longrightarrow1.
\]
Substituting these limits and \eqref{freq:classical-limit} into
\eqref{freq:value-identity} proves \eqref{freq:vanishing-limit}.

For the second assertion, Lemma~\ref{geom:gap} gives
$f_\Gamma(x)\to\kap>0$ as $x\downarrow0$. Choose $u>0$ such that
$f_\Gamma(x)\ge\kap/2$ for $0<x<u$.
By \eqref{geom:gap-sandwich},
\[
 \E_\nu[H]
 \ge8\E_\nu[\Gamma^{-2}]
 \ge8\int_0^u\frac{f_\Gamma(x)}{x^2}\,\dd x
 \ge4\kap\int_0^u\frac{\dd x}{x^2}=\infty.
\]
When $\alpha+2\delta<1$, the coefficient
$(1-\alpha-2\delta)\log((1-\alpha-\delta)/\delta)$ is strictly
positive. Integrating \eqref{freq:finite-lower} against $P_\nu$,
which assigns probability one to unique-best instances, therefore yields
\[
 \E_\nu[\tau^\pi]
 \ge(1-\alpha-2\delta)\log\frac{1-\alpha-\delta}{\delta}
      \E_\nu[H]=\infty.
\]
\end{proof}

\section{Computation of the Allocation and the Prior Quantile}
\label{comp:section}
\label{sec:computation}

The algorithm requires calculating an allocation $w^\star(\mu)$ for a instance vector $\mu$; and a prior quantile $h_a$.
We describe these two calculations below.

\subsection{A Scalar Equation for the Optimal Allocation}
\label{comp:root}

The allocation characterization of \citet[Theorem~5]{GK16}
specializes to a scalar equation for Gaussian rewards.
Fix an instance vector $\mu$ with a unique best arm, and define
\[
 b:=\argmax_{i\in[K]}\mu_i,\qquad
 \ell:=(\ell_i)_{i\ne b},\ \ell_i:=\mu_b-\mu_i>0,
 \qquad \ell_{\min}:=\min_{i\ne b}\ell_i.
\]

\begin{proposition}[Optimal Allocation]
\label{comp:scalar}
There is a unique $z_*\in(0,\ell_{\min}^2)$ satisfying
\begin{equation}\label{comp:root-bracket}
 \sum_{i\ne b}\left(\frac{z_*}{\ell_i^2-z_*}\right)^2=1,
 \qquad
 \frac{\ell_{\min}^2}{1+\sqrt{K-1}}
 \le z_*\le\frac{\ell_{\min}^2}{2}.
\end{equation}
The unique optimal allocation $w^\star(\mu)$ and characteristic
time are given by
\begin{equation}\label{comp:allocation}
 \begin{aligned}
 w_b^\star&=\left(1+\sum_{i\ne b}
                   \frac{z_*}{\ell_i^2-z_*}\right)^{-1},
 &w_i^\star&=\frac{z_*}{\ell_i^2-z_*}w_b^\star\quad(i\ne b),\\
 H(\mu)&=\frac{2}{z_*w_b^\star}.
 \end{aligned}
\end{equation}
\end{proposition}
\begin{proof}
Recall that
\[
 I(\mu,w)=\min_{i\ne b}
   \frac{w_bw_i}{w_b+w_i}\frac{\ell_i^2}{2},
 \qquad H(\mu)^{-1}=\max_{w\in\Delta_K}I(\mu,w).
\]
By Lemma~\ref{geom:hardness}, a maximizing allocation exists and
all its weights are positive. We first optimize the remaining
weights for a fixed $w_b$, then optimize over $w_b$.

\paragraph{Step 1: Fix the Weight of the Best Arm.}
Fix $w_b=t\in(0,1)$. We characterize when an allocation with this
weight can achieve $I(\mu,w)\ge c$ for a given $c>0$.
For each $i\ne b$,
\begin{align*}
 \frac{tw_i\ell_i^2}{2(t+w_i)}\ge c
 &\quad\Longleftrightarrow\quad
       w_i(t\ell_i^2-2c)\ge2ct\\
 &\quad\Longleftrightarrow\quad
       t\ell_i^2-2c>0,\qquad
       w_i\ge\frac{2ct}{t\ell_i^2-2c}.
\end{align*}
The strict positivity of $t\ell_i^2-2c$ is necessary because
$2ct>0$ and $w_i\ge0$.
Since $\sum_{i\ne b}w_i=1-t$, such an allocation exists if and only if
\begin{equation}\label{comp:feasible-information}
 0<c<\frac{t\ell_{\min}^2}{2},\qquad
 \sum_{i\ne b}\frac{2ct}{t\ell_i^2-2c}\le1-t.
\end{equation}
Necessity follows by summing the preceding lower bounds on $w_i$
and using $\sum_{i\ne b}w_i=1-t$.
For sufficiency, suppose \eqref{comp:feasible-information} holds.
Choose any challenger $j\ne b$ and set
\[
 w_i:=\frac{2ct}{t\ell_i^2-2c}
       \quad(i\notin\{b,j\}),\qquad
 w_j:=1-t-\sum_{i\notin\{b,j\}}
                  \frac{2ct}{t\ell_i^2-2c}.
\]
By \eqref{comp:feasible-information},
\[
 w_j\ge\frac{2ct}{t\ell_j^2-2c},
 \qquad \sum_{i\ne b}w_i=1-t.
\]
Thus the weights form a feasible allocation and satisfy
every comparison inequality, giving $I(\mu,w)\ge c$.

As $c$ increases from zero to $t\ell_{\min}^2/2$, the sum in
\eqref{comp:feasible-information} is continuous and strictly
increasing from zero to infinity. Therefore the largest feasible
$c$ is the unique solution of
\[
 \sum_{i\ne b}\frac{2ct}{t\ell_i^2-2c}=1-t.
\]
At this value, every challenger weight must equal its lower bound:
\[
 w_i=\frac{2ct}{t\ell_i^2-2c}\quad(i\ne b).
\]
Indeed, their lower bounds already sum to $1-t$, so increasing
any one weight would violate the sum constraint.

\paragraph{Step 2: Express the Optimization Using One Scalar.}
For the optimal $c$ at a fixed $t$, set $z:=2c/t$.
Then $0<z<\ell_{\min}^2$, and the preceding equalities become
\[
 w_i=t\frac{z}{\ell_i^2-z},\qquad
 t\left(1+\sum_{i\ne b}\frac{z}{\ell_i^2-z}\right)=1.
\]
The normalization equation determines the best-arm weight $t$
as a function of $z$:
 $t=\left(1+\sum_{i\ne b}\frac{z}{\ell_i^2-z}\right)^{-1}$.
We can therefore parameterize these allocations by $z$ instead
of $t$. Denote the resulting allocation by $w(z)$, whose
coordinates are
\begin{equation}\label{comp:scalar-family}
 w_b(z):=\left(1+\sum_{i\ne b}\frac{z}{\ell_i^2-z}\right)^{-1},
 \qquad
 w_i(z):=\frac{z}{\ell_i^2-z}w_b(z)\quad(i\ne b).
\end{equation}
We can verify that the formulas in
\eqref{comp:scalar-family} define a feasible allocation: For every $z\in(0,\ell_{\min}^2)$, these weights are
positive and sum to one. For each challenger, substitution gives
\[
 \frac{w_b(z)w_i(z)}{w_b(z)+w_i(z)}\frac{\ell_i^2}{2}
 =\frac{w_b(z)z}{\ell_i^2}\frac{\ell_i^2}{2}
 =\frac{zw_b(z)}2.
\]
All comparisons therefore have the same information $I$, which we denote by
\[
 c(z):=I(\mu,w(z))
 =\frac{zw_b(z)}2=\frac{z}{2\left(1+\sum_{i\ne b}z/(\ell_i^2-z)\right)}.
\]
Step~1 shows that any allocation with positive information can
be improved or matched by one of the form \eqref{comp:scalar-family}, while every such
allocation is feasible. Consequently,
\[
 \max_{w\in\Delta_K}I(\mu,w)
 =\sup_{0<z<\ell_{\min}^2}c(z).
\]

\paragraph{Step 3: Identify the Unique Maximizer.}
Using
$\frac{\dd}{\dd z}\frac{z}{\ell_i^2-z}
 =\frac{\ell_i^2}{(\ell_i^2-z)^2}$,
differentiation gives
\begin{align*}
 c'(z)
 &=\frac{1+\sum_{i\ne b}z/(\ell_i^2-z)
             -z\sum_{i\ne b}\ell_i^2/(\ell_i^2-z)^2}
        {2\left(1+\sum_{i\ne b}z/(\ell_i^2-z)\right)^2}\\
 &=\frac{1-\sum_{i\ne b}[z/(\ell_i^2-z)]^2}
        {2\left(1+\sum_{i\ne b}z/(\ell_i^2-z)\right)^2}.
\end{align*}
Each $z/(\ell_i^2-z)$ is positive and strictly increasing, and
\[
 \lim_{z\downarrow0}\sum_{i\ne b}
       \left(\frac{z}{\ell_i^2-z}\right)^2=0,
 \qquad
 \lim_{z\uparrow\ell_{\min}^2}\sum_{i\ne b}
       \left(\frac{z}{\ell_i^2-z}\right)^2=\infty.
\]
Continuity and strict increase give a unique $z_*$ where this sum
is one. Therefore $c'(z)>0$ for $z<z_*$ and $c'(z)<0$ for
$z>z_*$, so $z_*$ uniquely maximizes $c$.
The unique weights from Step~1 are thus $w(z_*)$, and
\[
 H(\mu)=\frac1{c(z_*)}=\frac{2}{z_*w_b(z_*)},
\]
which proves \eqref{comp:allocation}.

\paragraph{Step 4: Bound the Root.}
Since $\ell_i\ge\ell_{\min}$ for all challengers, with equality
for at least one of them, for every $0<z<\ell_{\min}^2$,
\[
 \left(\frac{z}{\ell_{\min}^2-z}\right)^2
 \le\sum_{i\ne b}\left(\frac{z}{\ell_i^2-z}\right)^2
 \le(K-1)\left(\frac{z}{\ell_{\min}^2-z}\right)^2.
\]
At $z=z_*$ the middle expression equals one. Taking positive
square roots gives
\[
 \frac1{\sqrt{K-1}}
 \le\frac{z_*}{\ell_{\min}^2-z_*}\le1.
\]
The left inequality yields
$\ell_{\min}^2\le(1+\sqrt{K-1})z_*$, and the right yields
$2z_*\le\ell_{\min}^2$. These are the bounds in
\eqref{comp:root-bracket}.
\end{proof}

Bisection on the interval in \eqref{comp:root-bracket} computes
$z_*$.
Substitution into \eqref{comp:allocation} gives
$w^\star(\mu)$ and $H(\mu)$.
For \textup{\textsc{Screen-and-Confirm}}, apply this calculation to
the instance $\mu^\ell$ in \eqref{alg:gap-instance}.
\subsection{Computation of the Prior Quantile}
\label{comp:quantile}

For a given $a\in(0,1)$, the quantile is the unique solution of
\[
 P_\nu(H>h_a)
 =\int_{\R^K}\ind\{H(\mu)>h_a\}
       \prod_{i=1}^K f_i(\mu_i)\,\dd\mu=a.
\]
It depends only on the prior and $a$, and can therefore be computed
before collecting any bandit observations.

\paragraph{Monte Carlo Computation.}
Draw $N$ independent mean vectors
$\mu^{(1)},\ldots,\mu^{(N)}\sim P_\nu$.
For each draw, compute
\[
 b^{(m)}:=\argmax_i\mu_i^{(m)},\qquad
 \ell_i^{(m)}:=\mu_{b^{(m)}}^{(m)}-\mu_i^{(m)}
       \quad(i\ne b^{(m)}),\qquad
 H^{(m)}:=H(\mu^{(m)}),
\]
using Proposition~\ref{comp:scalar}. The best arm is unique and
these gaps are positive almost surely. Sort the resulting values
as $H_{(1)}\le\cdots\le H_{(N)}$, and use the empirical quantile
\begin{equation}\label{comp:empirical-quantile}
 \widehat h_{a,N}:=H_{(\lceil N(1-a)\rceil)}.
\end{equation}

\begin{proposition}[Consistency of the Monte Carlo Quantile]
\label{comp:mc-consistency}
For every fixed $a\in(0,1)$,
\[
 \widehat h_{a,N}\longrightarrow h_a
 \qquad\text{almost surely as }N\to\infty,
\]
where almost-sure convergence is with respect to the independent
prior draws.
\end{proposition}
\begin{proof}
By Lemma~\ref{geom:regularity}, for any $0<u<h_a<v$,
\[
 P_\nu(H\le u)<1-a<P_\nu(H\le v).
\]
The strong law of large numbers gives, at $t=u,v$,
$N^{-1}\sum_{m=1}^N\ind\{H^{(m)}\le t\}\to P_\nu(H\le t)$
almost surely. Thus $u<\widehat h_{a,N}\le v$ eventually.
Taking countable sequences $u\uparrow h_a$ and $v\downarrow h_a$
proves the claim.
\end{proof}

\paragraph{Small-Budget Approximation.}
For small $a$, Corollary~\ref{geom:quantile}
provides a simple approximation in this regime:
\begin{equation}\label{comp:quantile-small-a}
 \lim_{a\downarrow0}\frac{h_a}{8\kappa_\nu^2/a^2}=1,
 \qquad \lim_{a\downarrow0}h_a=\infty.
\end{equation}
Thus $8\kappa_\nu^2/a^2$ approximates the quantile with vanishing
relative error as $a\downarrow0$. The constant $\kappa_\nu$ can be
computed by the one-dimensional integrals in \eqref{eq:kappa},
using numerical quadrature.

In the algorithm, $a=(1-4\eta)(\alpha+\delta)$, so
$\alpha+\delta\to0$ implies $a\to0$ and
\eqref{comp:quantile-small-a} applies. In the separated shrinking
regime, $\eta\to0$ and $a/\alpha\to1$, hence
\[
 \lim_{\substack{\alpha,\delta\downarrow0\\\alpha/\delta\to\infty}}
       \alpha^2h_a=8\kappa_\nu^2.
\]
For fixed $\alpha\in(0,1)$ and $\delta\downarrow0$, instead
$a\to\alpha$, and continuity of the quantile map
(Lemma~\ref{geom:regularity}) gives $h_a\to h_\alpha$.

\section{Related Work: Erasure Decoding}\label{app:erasure}

Our abstention formulation has a natural analogue in erasure decoding in information theory. A decoder may either output a message or declare an erasure, distinguishing an incorrect decision from an inconclusive outcome. Classical analyses characterize the tradeoff between total error probability
(including erasures) and undetected error probability, beginning with \citet{forney1968exponential} and with subsequent refinements by
\citet{merhav2008error,somekhbaruch2011exact,hayashi2015asymmetric}.
Abstention in BAI plays a role analogous to erasure.
Unlike fixed-blocklength decoding, however, the learner also
adaptively chooses which arms to sample and when to stop,
so the objective includes minimizing the sampling cost.

\end{document}